\documentclass[10pt,twocolumn,letterpaper]{article}

\usepackage{iccv}
\usepackage{times}
\usepackage{epsfig}
\usepackage{graphicx}
\usepackage{amsmath}
\usepackage{amssymb}
\usepackage{multirow}
\usepackage{rotating}
\usepackage{booktabs}
\usepackage{caption}
\usepackage{subcaption}

\usepackage[accsupp]{axessibility}  

\usepackage[pagebackref=true,breaklinks=true,bookmarks=false,colorlinks=true]{hyperref}

\iccvfinalcopy 

\def\iccvPaperID{9103} 

\usepackage[capitalize]{cleveref}
\crefname{section}{Sec.}{Secs.}
\Crefname{section}{Section}{Sections}
\Crefname{table}{Table}{Tables}
\crefname{table}{Tab.}{Tabs.}

\usepackage{graphicx}
\usepackage[table]{xcolor}

\definecolor{LightCyan}{rgb}{0.87,0.92,0.96}
\definecolor{forestgreen}{rgb}{0.133, 0.545, 0.133}
\definecolor{LightGreen}{RGB}{245, 255, 245}
\definecolor{m_red}{RGB}{255, 126, 121}
\definecolor{m_yellow}{RGB}{255, 225, 0}
\definecolor{m_green}{RGB}{190, 225, 102}
\definecolor{m_violet}{RGB}{215, 131, 255}
\definecolor{m_darkred}{RGB}{204, 0, 0}
\definecolor{m_blue}{RGB}{0, 102, 204}
\definecolor{m_gray}{RGB}{128, 128, 128}
\definecolor{light_blue}{RGB}{0, 153, 153}
\definecolor{light_purple}{RGB}{153, 153, 255}
\definecolor{m_pink}{RGB}{255, 0, 127}
\definecolor{m_sky}{RGB}{0, 112, 192}
\definecolor{m_gray}{RGB}{165, 165, 165}
\definecolor{m_mlp}{RGB}{192, 0, 0}
\definecolor{m_comm}{RGB}{255, 192, 0}
\definecolor{correct}{RGB}{173, 173, 173}
\definecolor{incorrect}{RGB}{192, 0, 0}

\definecolor{fidnet}{RGB}{165, 165, 165}
\definecolor{cenet}{RGB}{192, 0, 0}
\definecolor{rangeformer}{RGB}{0, 112, 192}
\definecolor{groundtruth}{RGB}{0, 0, 0}

\newcommand{\PQ}{PQ}
\newcommand{\PQda}{PQ\textsuperscript{$\dagger$}}

\newcommand{\PQth}{PQ\textsuperscript{Th}}
\newcommand{\RQth}{RQ\textsuperscript{Th}}
\newcommand{\SQth}{SQ\textsuperscript{Th}}
\newcommand{\PQst}{PQ\textsuperscript{St}}
\newcommand{\RQst}{RQ\textsuperscript{St}}
\newcommand{\SQst}{SQ\textsuperscript{St}}

\definecolor{Royal}{rgb}{0.,0.4392,0.7529}

\usepackage{marvosym}

\usepackage{algorithm}
\usepackage{listings}

\begin{document}

\title{Rethinking Range View Representation for LiDAR Segmentation}

\author{Lingdong Kong$^{1,2}$ \quad Youquan Liu$^{1,3}$ \quad Runnan Chen$^{1,4}$ \quad Yuexin Ma$^5$ \quad Xinge Zhu$^6$\\
Yikang Li$^1$ \quad Yuenan Hou$^{1,\textrm{\Letter}}$ \quad Yu Qiao$^1$ \quad Ziwei Liu$^{7,\textrm{\Letter}}$\\
{\small
$^1$Shanghai AI Laboratory\quad
$^2$National University of Singapore\quad
$^3$Hochschule Bremerhaven\quad
$^4$The University of Hong Kong}\\
{\small
$^5$ShanghaiTech University\quad
$^6$The Chinese University of Hong Kong\quad
$^7$S-Lab, Nanyang Technological University}
\\
{\tt\small \{konglingdong,liuyouquan,chenrunnan,houyuenan\}@pjlab.org.cn \quad {\tt\small ziwei.liu@ntu.edu.sg}
}}

\maketitle

\begin{abstract}
   LiDAR segmentation is crucial for autonomous driving perception. Recent trends favor point- or voxel-based methods as they often yield better performance than the traditional range view representation. In this work, we unveil several key factors in building powerful range view models. We observe that the ``many-to-one" mapping, semantic incoherence, and shape deformation are possible impediments against effective learning from range view projections. We present \textbf{RangeFormer} -- a full-cycle framework comprising novel designs across network architecture, data augmentation, and post-processing -- that better handles the learning and processing of LiDAR point clouds from the range view. We further introduce a \textbf{S}calable \textbf{T}raining from \textbf{R}ange view \textbf{(STR)} strategy that trains on arbitrary low-resolution 2D range images, while still maintaining satisfactory 3D segmentation accuracy. We show that, for the first time, a range view method is able to surpass the point, voxel, and multi-view fusion counterparts in the competing LiDAR semantic and panoptic segmentation benchmarks, i.e., SemanticKITTI, nuScenes, and ScribbleKITTI.
\end{abstract}

\section{Introduction}
\label{sec:introduction}

LiDAR point clouds have unique characteristics. As the direct reflections of real-world scenes, they are often diverse and unordered and thus bring extra difficulties in learning \cite{guo2020deep,li2020deep}. Inevitably, a good representation is needed for efficient and effective LiDAR point cloud processing \cite{2022Analyzing}.

Although there exist various LiDAR representations as shown in ~\cref{table:compare}, the prevailing approaches are mainly based on point view \cite{hu2019randla,thomas2019kpconv}, voxel view \cite{2019Minkowski,tang2020searching,zhu2021cylindrical,4D-DS-Net}, and multi-view fusion \cite{2020AMVNet,xu2021rpvnet,qiu2022GFNet}. These methods, however, require computationally intensive neighborhood search \cite{qi2017pointnet++}, 3D convolution operations \cite{2019PVCNN}, or multi-branch networks \cite{2021MPF,2021Tornado-net}, which are often inefficient during both training and inference stages. The projection-based representations, such as range view \cite{wu2018squeezeseg,milioto2019rangenet++} and bird's eye view \cite{zhang2020polarnet,zhou2021panoptic}, are more tractable options. The 3D-to-2D rasterizations and mature 2D operators open doors for fast and scalable in-vehicle LiDAR perception \cite{milioto2019rangenet++,xu2020squeezesegv3,2022Analyzing}. Unfortunately, the segmentation accuracy of current projection-based methods \cite{zhao2021fidnet,cheng2022cenet,zhang2020polarnet} is still far behind the trend \cite{2022_2DPASS,xu2021rpvnet,2022GASN}. 

\begin{figure}[t]
    \begin{center}
    \includegraphics[width=1.0\linewidth]{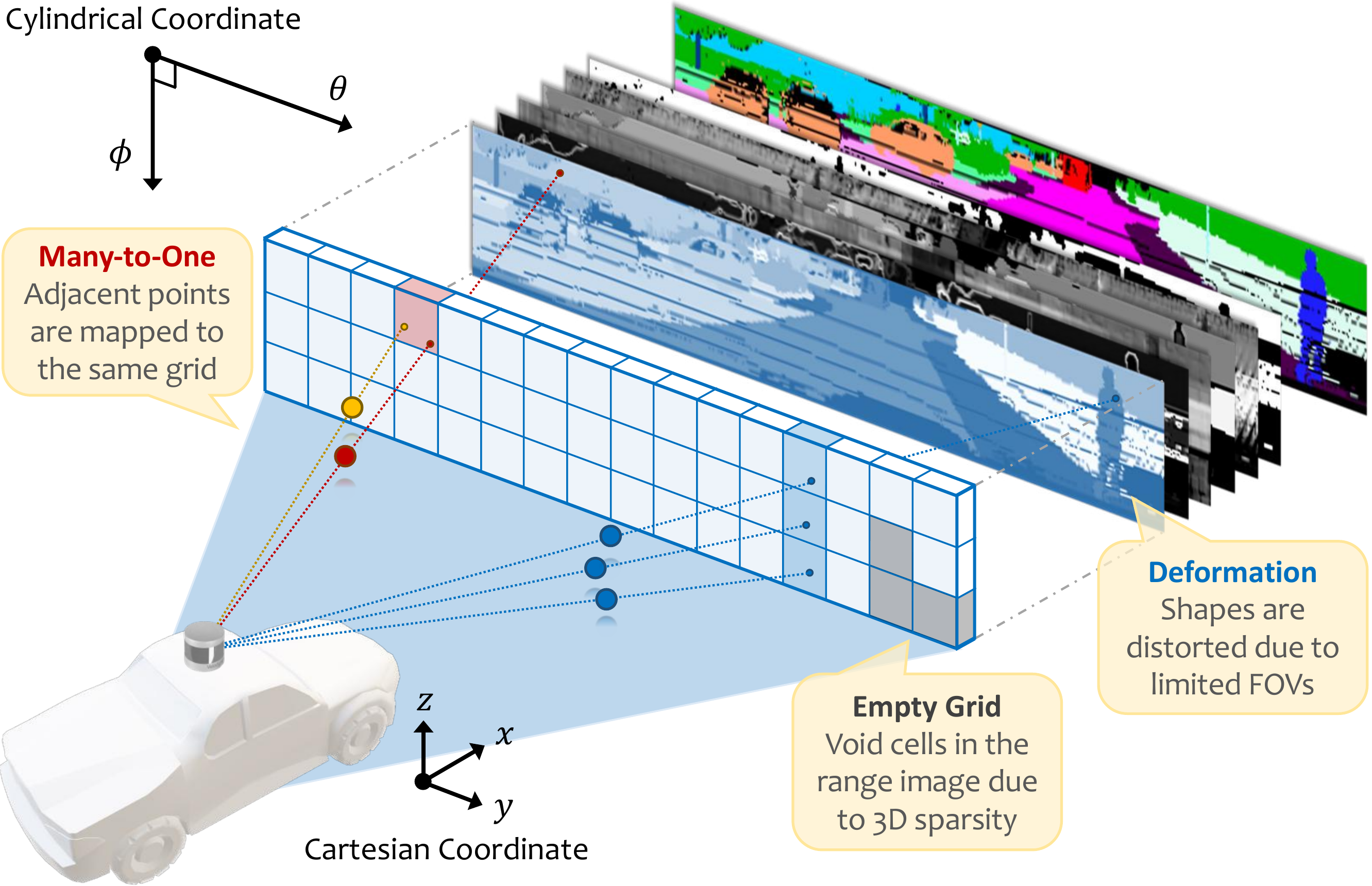}
    \end{center}
    \vspace{-0.45cm}
    \caption{Three detrimental factors observed in the LiDAR range view representation: 1) the ``many-to-one" problem; 2) ``holes" or empty grids; and 3) shape distortions.}
    \label{figure:teaser}
    \vspace{0.16cm}
\end{figure}

\begin{table}[t]
\caption{Comparisons for different LiDAR representations.}
\vspace{-0.2cm}
\centering\scalebox{0.662}{
\begin{tabular}{c|ccc}
\toprule
\textbf{View} & \textbf{Formation} & \textbf{Complexity} & \textbf{Representative}
\\\midrule
Raw Points & Bag-of-Points & $\mathcal{O}(N\cdot d)$ & RandLA-Net, KPConv
\\
\cellcolor{LightCyan}Range View & \cellcolor{LightCyan}Range Image & \cellcolor{LightCyan}$\mathcal{O}(\frac{H\cdot W}{r^2}\cdot d)$ & \cellcolor{LightCyan}SqueezeSeg, RangeNet++
\\
Bird's Eye View & Polar Image & $\mathcal{O}(\frac{H\cdot W}{r^2}\cdot d)$ & PolarNet
\\
Voxel (Dense) & Voxel Grid & $\mathcal{O}(\frac{H\cdot W\cdot L}{r^3}\cdot d)$ & PVCNN
\\
Voxel (Sparse) & Sparse Grid & $\mathcal{O}(N\cdot d)$ & MinkowskiNet, SPVNAS
\\
Voxel (Cylinder) & Sparse Grid & $\mathcal{O}(N\cdot d)$ & Cylinder3D
\\
Multi-View & Multiple & $\mathcal{O}((N+\frac{H\cdot W}{r^2})\cdot d)$ & AMVNet, RPVNet
\\\bottomrule
\end{tabular}}
\label{table:compare}
\end{table}

The challenge of learning from projected LiDAR scans comes from the potential detrimental factors of the LiDAR data representation \cite{milioto2019rangenet++}. As shown in \cref{figure:teaser}, the range view projection\footnote{We show a frustum of the LiDAR scan for simplicity; the complete range view projection is a cylindrical panorama around the ego-vehicle.} often suffers from several difficulties, including 1) the ``many-to-one" conflict of adjacent points, caused by limited horizontal angular resolutions; 2) the ``holes" in the range images due to 3D sparsity and sensor disruptions; and 3) potential shape deformations during the rasterization process. While these problems are ubiquitous in range view learning, previous works hardly consider tackling them. Stemming from the image segmentation community \cite{zhang2021k-net}, prior arts widely adopt the fully-convolutional networks (FCNs) \cite{long2015fully,ChenPKMY18} for range view LiDAR segmentation \cite{milioto2019rangenet++,zhao2021fidnet,cheng2022cenet,kong2021conda}. The limited receptive fields of FCNs cannot directly model long-term dependencies and are thus less effective in handling the mentioned impediments.

In this work, we seek an alternative in lieu of the current range view LiDAR segmentation models. Inspired by the success of Vision Transformer (ViT) and its follow-ups \cite{2021_ViT,2021_PVT,2021segformer,2021_Swin,2021_Segmenter}, we design a new framework dubbed \textit{RangeFormer} to better handle the learning and processing of LiDAR point clouds from the range view. We formulate the segmentation of range view grids as a seq2seq problem and adopt the standard self-attention modules \cite{2017transformer} to capture the rich contextual information in a ``global" manner, which is often omitted in FCNs \cite{milioto2019rangenet++,aksoy2019salsanet,cheng2022cenet}. The hierarchical features extracted with such global awareness are then fed into multi-layer perceptions (MLPs) for decoding. In this way, every point in the range image is able to establish interactions with other points -- no matter whether close or far and valid or empty -- and further lead to more effective representation learning from the LiDAR range view.

It is worth noting that such architectures, albeit straightforward, still suffer several difficulties. The first issue is related to data diversity. The prevailing LiDAR segmentation datasets \cite{nuScenes,Panoptic-nuScenes,SemanticKITTI,WaymoOpen} contain tens of thousands of LiDAR scans for training. These scans, however, are less diverse in the sense that they are collected in a sequential way. This hinders the training of Transformer-based architectures as they often rely on sufficient samples and strong data augmentations \cite{2021_ViT}. To better handle this, We design an augmentation combo that is tailored for range view. Inspired by recent 3D augmentation techniques \cite{zhou2021panoptic,kong2022lasermix,nekrasov2021mix3d}, we manipulate the range view grids with row mixing, view shifting, copy-paste, and grid fill. As we will show in the following sections, these lightweight operations can significantly boost the performance of SoTA range view methods.

The second issue comes from data post-processing. Prior works adopt CRF \cite{wu2018squeezeseg} or k-NN \cite{milioto2019rangenet++} to smooth/infer the range view predictions. However, it is often hard to find a good balance between the under- and over-smoothing of the 3D labels in unsupervised manners \cite{kochanov2020kprnet}. In contrast, we design a supervised post-processing approach that first sub-samples the whole LiDAR point cloud into equal-interval ``sub-clouds" and then infer their semantics, which holistically reduces the uncertainty of aliasing range view grids.

To further reduce the overhead in range view learning, we propose \textit{STR} -- a scalable range view training paradigm. \textit{STR} first ``divides" the whole LiDAR scan into multiple groups along the azimuth direction and then ``conquers" each of them. This transforms range images of high horizontal resolutions into a stack of low-resolution ones while can better maintain the best-possible granularity to ease the ``many-to-one" conflict. Empirically, We find \textit{STR} helpful in reducing the complexity during training, without sacrificing much convergence rate and segmentation accuracy.

The advantages of \textit{RangeFormer} and \textit{STR} are demonstrated from aspects of LiDAR segmentation accuracy and efficiency on prevailing benchmarks. Concretely, we achieve $73.3\%$ mIoU and $64.2\%$ $\text{PQ}$ on SemanticKITTI \cite{SemanticKITTI}, surpassing prior range view methods \cite{zhao2021fidnet,cheng2022cenet} by significant margins and also better than SoTA fusion-based methods \cite{2022_2DPASS,pvkd2022,2022GASN}. We also establish superiority on the nuScenes \cite{Panoptic-nuScenes} (sparser point clouds) and ScribbleKITTI \cite{2022ScribbleKITTI} (weak supervisions) datasets, which validates our scalability. While being more effective, our approaches run $2\times$ to $5\times$ faster than recent voxel \cite{zhu2021cylindrical,tang2020searching} and fusion \cite{xu2021rpvnet,2022_2DPASS} methods and can operate at sensor frame rate.

\section{Related Work}
\label{sec:relatedwork}

\noindent\textbf{LiDAR Representation}. The LiDAR sensor is designed to capture high-fidelity 3D structural information which can be represented by various forms, \textit{i.e.}, raw point \cite{qi2017pointnet,qi2017pointnet++,thomas2019kpconv}, range view \cite{hu2018squeeze,wu2019squeezesegv2,xu2020squeezesegv3,aksoy2019salsanet}, bird's eye view (BEV) \cite{zhang2020polarnet}, voxel \cite{2019PVCNN,2019Minkowski,zhu2021cylindrical,2022GASN,chen2023clip2scene}, and multi-view fusion \cite{2020AMVNet,xu2021rpvnet,2022_2DPASS}, as summarized in \cref{table:compare}. The point and sparse voxel methods are prevailing but suffer $\mathcal{O}(N\cdot d)$ complexity, where $N$ is the number of points and often in the order of $10^5$ \cite{2022Analyzing}. BEV offers an efficient representation but only yields sub-par performance \cite{PolarStream}. As for fusion-based methods, they often comprise multiple networks that are too heavy to yield reasonable training overhead and inference latency \cite{qiu2022GFNet,2022GASN,2022MSSNet}. Among all representations, range view is the one that directly reflects the LiDAR sampling process~\cite{FCOS-LiDAR,2021_RangeDet,triess2020scan}. We thus focus on this modality to further embrace its compactness and rich semantic/structural cues.

\noindent\textbf{Architecture}. Previous range view methods are built upon mature FCN structures~\cite{long2015fully,wu2018squeezeseg,wu2019squeezesegv2,xu2020squeezesegv3,alonso20203d-mininet}. RangeNet++~\cite{milioto2019rangenet++} proposed an encoder-decoder FCN based on DarkNet~\cite{2018_redmon_yolov3}. SalsaNext~\cite{cortinhal2020salsanext} uses dilated convolutions to further expand the receptive fields. Lite-HDSeg~\cite{razani2021lite} proposed to adopt harmonic convolution to reduce the computation overhead. EfficientLPS~\cite{sirohi2021efficientlps} proposed a proximity convolution module to leverage neighborhood points in the range image. FIDNet~\cite{zhao2021fidnet} and CENet~\cite{cheng2022cenet} switch the encoders to ResNet and replace the decoder with simple interpolations. In contrast to using FCNs, we build \textit{RangeFormer} upon self-attentions and demonstrate potential and advantages for long-range dependency modeling in range view learning.

\begin{figure*}[t]
    \begin{center}
    \includegraphics[width=1.0\textwidth]{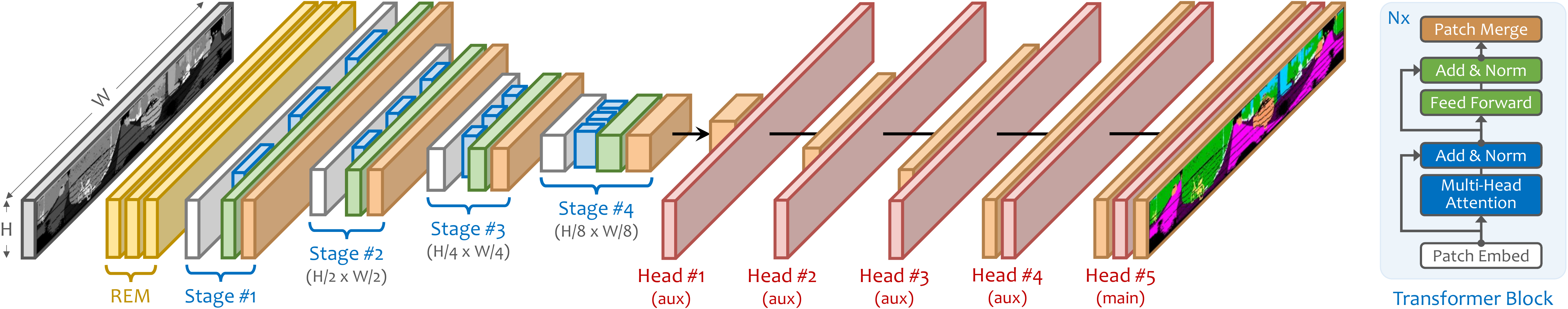}
    \end{center}
    \vspace{-0.4cm}
    \caption{\textbf{Architecture overview}. The rasterized LiDAR point cloud of spatial size $H\times W$ is fed into four consecutive stages where each comprising several standard Transformer blocks as shown in the right subfigure. The multi-scale features extracted from these different stages are then fed into the MLP heads for decoding. The final predictions in 2D will be projected back to 3D in a reverse manner of \cref{eq:rv}.}
    \label{figure:network}
\end{figure*}

\noindent\textbf{Augmentation}. Most 3D data augmentation techniques are object-centric \cite{2022_PointCutMix,2020_PointMixUp,ren2022modelnet-c,2021_RSMix} and thus not generalizable to scenes. Panoptic-PolarNet \cite{zhou2021panoptic} over-samples rare instance points during training. Mix3D \cite{nekrasov2021mix3d} proposed an out-of-context mixing by supplementing points from one scene to another. MaskRange \cite{2022MaskRange} designs a weighted paste drop augmentation to alleviate overfitting and improve class balance. LaserMix \cite{kong2022lasermix} proposed to mix labeled and unlabeled LiDAR scans along the inclination axis for effective semi-supervised learning. In this work, we present a novel and lightweight augmentation combo tailored for range view learning that combines mixing, shifting, union, and copy-paste operations directly on the rasterized grids, while still maintaining the structural consistency of the scenes. 

\noindent\textbf{Post-Processing}. Albeit being an indispensable module of range view LiDAR segmentation, prior works hardly consider improving the post-processing process \cite{2022Analyzing}. Most works follow the CRF \cite{wu2018squeezeseg} or k-NN \cite{milioto2019rangenet++} to smooth or infer the semantics for conflict points. Recently, Zhao~\textit{et al.} proposed another unsupervised method named NLA for nearest label assignment \cite{zhao2021fidnet}. We tackle this in a supervised way by creating ``sub-clouds" from the full point cloud and inferring labels for each subset, which directly reduces the information loss and helps alleviate the ``many-to-one" problem.

\section{Technical Approach}
\label{sec:approach}

In this section, we first revisit the details of range view rasterization (\cref{sec:preliminaries}). To better tackle the impediments in range view learning, we introduce \textit{RangeFormer} (\cref{sec:rangeformer}) and \textit{STR} (\cref{sec:str}) which emphasize the effectiveness and efficiency, respectively, for scalable LiDAR segmentation.  

\subsection{Preliminaries}
\label{sec:preliminaries}
Mounted on the roof of the ego-vehicle (as illustrated in \cref{figure:teaser}), the rotating LiDAR sensor emits isotropic laser beams with predefined angles and perceives the positions and reflection intensity of surroundings via time measurements in the scan cycle. Specifically, each LiDAR scan captures and returns $N$ points in a single scan cycle, where each point $p_n$ in the scan is represented by the Cartesian coordinates $(p^x_n,p^y_n,p^z_n)$, intensity $p^i_n$, and existence $p^e_n$. 

\noindent\textbf{Rasterization}.
For a given LiDAR point cloud, we rasterize points within this scan into a 2D cylindrical projection $\mathcal{R}(u,v)$ (\textit{a.k.a.}, range image) of size $H\times W$, where $H$ and $W$ are the height and width, respectively. The rasterization process for each point $p_n$ can be formulated as follows:
\begin{equation}
\label{eq:rv}
\begin{pmatrix}
\mathit{u}_n  \\
\mathit{v}_n
\end{pmatrix}
=
\begin{pmatrix}
\frac{1}{2}~[1-\arctan(p^y_n,p^x_n)\pi^{-1}]~W  \\
~[1-(\arcsin(p^z_n,{(p^d_n)}^{-1})+\phi^{\text{down}})\xi^{-1}]~H~
\end{pmatrix},
\end{equation} 
where $(u_n,v_n)$ denotes the grid coordinate of point $p_n$ in range image $\mathcal{R}(u,v)$; $p^d_n=\sqrt{(p^x_n)^2+(p^y_n)^2+(p^z_n)^2}$ is the depth between the point and LiDAR sensor (ego-vehicle); $\xi=|\phi^{\text{up}}|+|\phi^{\text{down}}|$ denotes the vertical field-of-views (FOVs) of the sensor and $\phi^{\text{up}}$ and $\phi^{\text{down}}$ are the inclination angles at the upward and downward directions, respectively. Note that $H$ is often predefined by the beam number of the LiDAR sensor, while $W$ can be set based on requirements.

\noindent\textbf{Formation}. The final range image $\mathcal{R}(u,v)\in\mathbb{R}^{(6,H,W)}$ is composed of six rasterized feature embeddings, \textit{i.e.}, coordinates $(p^x,p^y,p^z)$, depth $p^d$, intensity $p^i$, and existence $p^e$ (indicates whether or not a grid is occupied by valid point). The range semantic label $y(u,v)\in\mathbb{R}^{(H,W)}$ -- which is rasterized from the per-point label in 3D -- shares the same rasterization index and resolution with $\mathcal{R}(u,v)$.
The 3D segmentation problem is now turned into a 2D one and the grid predictions in the range image can then be projected back to point-level in a reverse manner of \cref{eq:rv}.

\subsection{RangeFormer: A Full-Cycle Framework}
\label{sec:rangeformer}

As discussed in previous sections, there exist potential detrimental factors in the range view representation (\cref{figure:teaser}). The one-to-one correspondences from \cref{eq:rv} are often untenable since $H\times W$ is much less than $N$. Typically, prior arts \cite{milioto2019rangenet++,2021MPF,cheng2022cenet} adopt $(H,W)=(64,512)$ to rasterize LiDAR scans of around $120$k points each \cite{SemanticKITTI}, resulting in over $70\%$ information loss\footnote{Note: \# of 2D grids $/$ \# of 3D points $=$ $64\times512$ $/$ $120000\approx 27.3\%$.}. The restricted horizontal angular resolutions and an intensive number of empty grids in range image tend to bring extra difficulties during model training, such as shape deformation, semantic incoherence,~\textit{etc}.

\noindent\textbf{Architecture}. To pursue larger receptive fields and longer dependency modeling, we design a self-attention-based network comprising standard Transformer blocks and MLP heads as shown in \cref{figure:network}. Given a batch of rasterized range image $\mathcal{R}(u,v)$, the range embedding module (REM) which consists of three MLP layers first maps each point in the grid to a higher-dim embedding $\mathcal{F}_0\in\mathbb{R}^{(128,H,W)}$. This is analogous to PointNet \cite{qi2017pointnet}. Next, we divide $\mathcal{F}_0$ into overlapping patches of size $3$ by $3$ and feed them into the Transformer blocks. Similar to PVT \cite{2021_PVT}, we design a pyramid structure to facilitate multi-scale feature fusions, yielding $\{\mathcal{F}_1, \mathcal{F}_2, \mathcal{F}_3, \mathcal{F}_4\}$ for four stages, respectively, with downsampling factors $1$, $2$, $4$, and $8$. Each stage consists of customized numbers of Transformer blocks and each block includes two modules. 1) \textit{Multi-head self-attention} \cite{2017transformer}, serves as the main computing bottleneck and can be formulated as:
\vspace{-0.4cm}
\begin{equation}
\label{eq:attention}
O=\text{Mul}(Q,K,V) = \text{Concat}(\text{head}_1,...,\text{head}_h)W^{O},
\end{equation}
where $\text{head}_i=\text{Attention}(QW_i^Q, KW_i^K, VW_i^V)$ denotes the self-attention operation with $\text{Attention}=\sigma(\frac{QK}{\sqrt{d^{\text{head}}}})V$; $\sigma$ denotes softmax and $d^{\text{head}}$ is the dimension of each head; $W^Q$, $W^K$, $W^V$, and $W^O$ are the weight matrices of query $Q$, key $K$, value $V$, and output $O$. As suggested in \cite{2021_PVT}, the sequence lengths of $K$ and $V$ are further reduced by a factor $R$ to save the computation overhead. 2) \textit{Feed-forward network (FFN)}, which consists of MLPs and activation as:
\begin{equation}
\label{eq:ffn}
\mathcal{F} = \text{FFN}(O) = \text{Linear}(\text{GELU}(\text{Linear}(O))) \oplus O,
\end{equation}
where $\oplus$ denotes the residual connection~\cite{he2016deep}. Different from ViT \cite{geiger2013vision}, we discard the explicit position embedding and rather incorporate it directly within the feature embeddings. As introduced in \cite{2021segformer}, this can be achieved by adding a single $3$ by $3$ convolution with zero paddings into FFN.

\noindent\textbf{Semantic Head}. To avoid heavy computations in decoding, we adopt simple MLPs as the segmentation heads. After retrieving all features from the four stages, we first unify their dimensions. This is achieved in two steps: 1) \textit{Channel unification}, where each $\mathcal{F}_i$ with embedding size $d^{\mathcal{F}_i}, i=1, 2, 3, 4$, is unified via one MLP layer. 2) \textit{Spatial unification}, where $\mathcal{F}_i$ from the last three stages are resized to the range embedding size $H\times W$ by simple bi-linear interpolation. The decoding process for stage $i$ is thus:
\begin{equation}
\label{eq:decode}
\mathcal{H}_i=\text{Bi-Interpolate}(\text{Linear}(\mathcal{F}_i)).
\end{equation}
As proved in \cite{zhao2021fidnet}, the bi-linear interpolation of range view grids is equivalent to the distance interpolation (with four neighbors) in PointNet++~\cite{qi2017pointnet++}. Here the former operation serves as the better option since it is totally parameter-free.
Finally, we concatenate four $\mathcal{H}_i$ together and feed it into another two MLP layers, where the channel dimension is gradually mapped to $d^{\text{cls}}$, \textit{i.e.} the class number, to form the class probability distribution. Additionally, we add an extra MLP layer for each $\mathcal{H}_i$ as the auxiliary head. The predictions from the main head and four auxiliary heads are supervised separately during training. As for inference, we only keep the main head and discard the auxiliary ones.

\noindent\textbf{Panoptic Head}. Similar to Panoptic-PolarNet \cite{zhou2021panoptic}, we add a panoptic head on top of \textit{RangeFormer} to estimate the instance centers and offsets, dubbed \textit{Panoptic-RangeFormer}. Since we tackle this problem in a bottom-up manner, the semantic predictions of the \textit{things} classes are utilized as the foreground mask to form instance groups in 3D. Next, we conduct 2D class-agnostic instance grouping by predicting the center heatmap \cite{2020_Panoptic-DeepLab} and offsets for each point on the $XY$-plane. Based on \cite{zhou2021panoptic}, the predictions from the above two aspects can then be fused via majority voting. As we will show in the experiments, the advantages of \textit{RangeFormer} in semantic learning further yield much better panoptic segmentation performance.

\noindent\textbf{RangeAug}. Data augmentation often helps the model learn more general representations and thus increases both accuracy and robustness. Prior arts in LiDAR segmentation conduct a series of augmentations at point-level~\cite{zhu2021cylindrical}, \textit{i.e.}, global rotation, jittering, flipping, and random dropping, which we refer to as ``common" augmentations. To better embrace the rich semantic and structural cues of the range view representation, we propose an augmentation combo comprising the following four operations.

1) \textit{RangeMix}, which mixes two scans along the inclination $\phi=\arctan(\frac{p^z}{\sqrt{(p^x)^2+(p^y)^2}})$ and azimuth $\theta$ directions. This can be interpreted as switching certain rows of two range images. After calculating $\phi$ and $\theta$ for the current scan and the randomly sampled scan, we then split points into $k_\text{mix}$ equal spanning inclination ranges, \textit{i.e.}, different mixing strategies. The corresponding points in the same inclination range from the two scans are then switched. In our experiments, we design mixing strategies from a combination, and $k_\text{mix}$ is randomly sampled from a list $[2, 3, 4, 5, 6]$. 

2) \textit{RangeUnion}, which fills in the empty grids of one scan with grids from another scan. Due to the sparsity in 3D and potential sensor disruptions, a huge number of grids are empty even after rasterization. We thus use the existence embedding $p^e$ to search and fill in these void grids and this further enriches the actual capacity of the range image. Given a number of $N_\text{union}=\sum_n p^e_n$ empty range view grids, we randomly select $k_\text{union}N_\text{union}$ candidate grids for point filling, where $k_\text{union}$ is set as $50\%$.

3) \textit{RangePaste}, which copies tail classes from one scan to another scan at correspondent positions in the range image. This boosts the learning of rare classes and also maintains the objects' spatial layout in the projection. The ground-truth semantic labels of a randomly sampled scan are used to create pasting masks. The classes to be pasted are those in the ``tail" distribution, which forms a semantic class list (\texttt{sem classes}). After indexing the rare classes' points, we paste them into the current scan while maintaining the corresponding positions in the range image.

4) \textit{RangeShift}, which slides the scan along the azimuth direction $\theta=\arctan(p^y/p^x)$ to change the global position embedding. This corresponds to shifting the range view grids along the row direction with $k_\text{shift}$ rows. In our experiments, $k_\text{shift}$ is randomly sampled from a range of $\frac{W}{4}$ to $\frac{3W}{4}$. These four augmentations are tailored for range view and can operate on-the-fly during the data loading process, without adding extra overhead during training. As we will show in the next section, they play a vital role in boosting the performance of range view segmentation models.

\noindent\textbf{RangePost}. The widely-used k-NN~\cite{milioto2019rangenet++} votes and assigns labels for points near the boundary in an unsupervised way, which cannot handle the ``many-to-one" conflict concretely. Differently, we tackle this in a supervised manner. We first sub-sample the whole point cloud into equal-interval ``sub-clouds". Since adjacent points have a high likelihood of belonging to the same class, these ``sub-clouds" are sharing very similar semantics. Next, we stack and feed these subsets to the network. After obtaining the predictions, we then stitch them back to their original positions. For each scan, this will automatically assign labels for points that are merged during rasterization in just a single forward pass, which directly reduces the information loss caused by ``many-to-one" mappings. Finally, prior post-processing techniques \cite{milioto2019rangenet++,zhao2021fidnet} can then be applied to these new predictions to further enhance the re-rasterization process.

\begin{figure}[t]
    \begin{center}
    \includegraphics[width=1.0\linewidth]{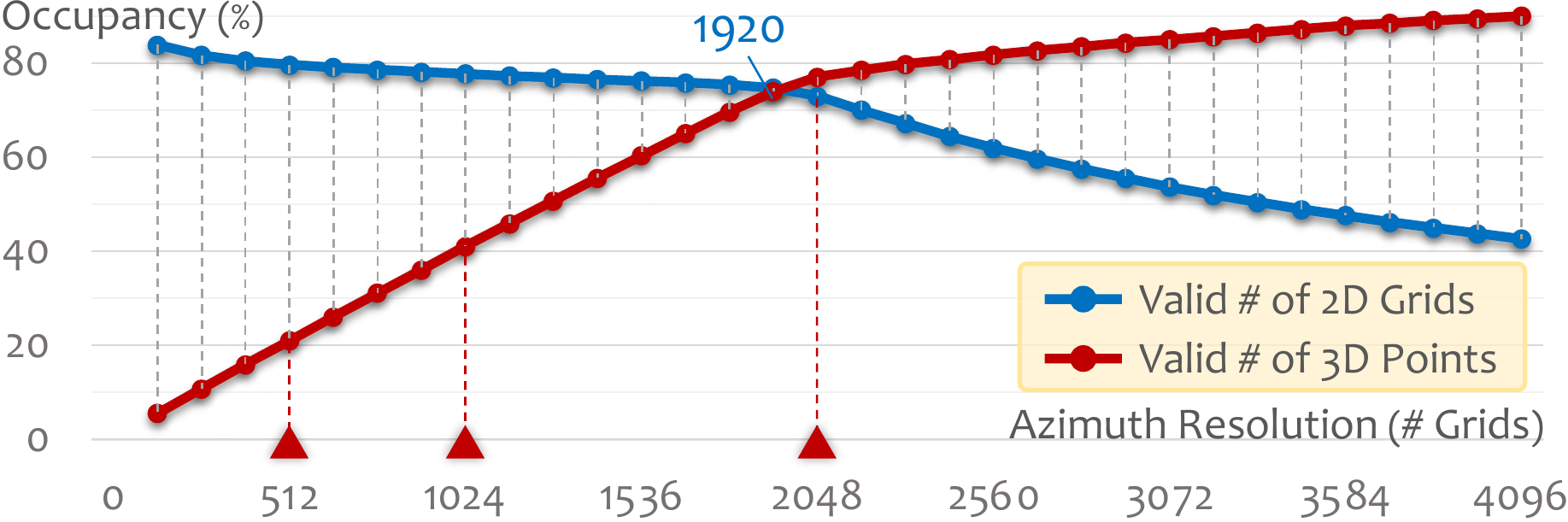}
    \end{center}
    \vspace{-0.4cm}
    \caption{The \textbf{occupancy trade-off} between 2D grids \& 3D points in the LiDAR range view representation. Statistics calculated on the SemanticKITTI~\cite{SemanticKITTI} dataset.}
    \label{fig:occupancy}
\end{figure}

\subsection{STR: Scalable Training from Range View}
\label{sec:str}
To pursue better training efficiency, prior works adopt low horizontal angular resolutions, \textit{i.e.}, small values of $W$ in \cref{eq:rv}, for range image rasterization \cite{milioto2019rangenet++,2021MPF}. This inevitably intensifies the ``many-to-one" conflict, causes more severe shape distortions, and leads to sub-par performance.

\noindent\textbf{2D \& 3D Occupancy}. Instead of directly assigning small $W$ for $\mathcal{R}(u,v)$, we first lookup for the best possible options. We find an ``occupancy trade-off" between the number of points in the LiDAR scan and the desired capacity of the range image. As shown in \cref{fig:occupancy}, the conventional choices, \textit{i.e.}, $512$, $1024$, and $2048$, are not optimal. The crossover of two lines indicates that the range image of width $1920$ tends to be the most \textit{informative} representation. However, this configuration inevitably consumes much more memory than the conventionally used $512$ or $1024$ resolutions and further increases the training and inference overhead. 

\begin{figure}[t]
    \begin{center}
    \includegraphics[width=1.0\linewidth]{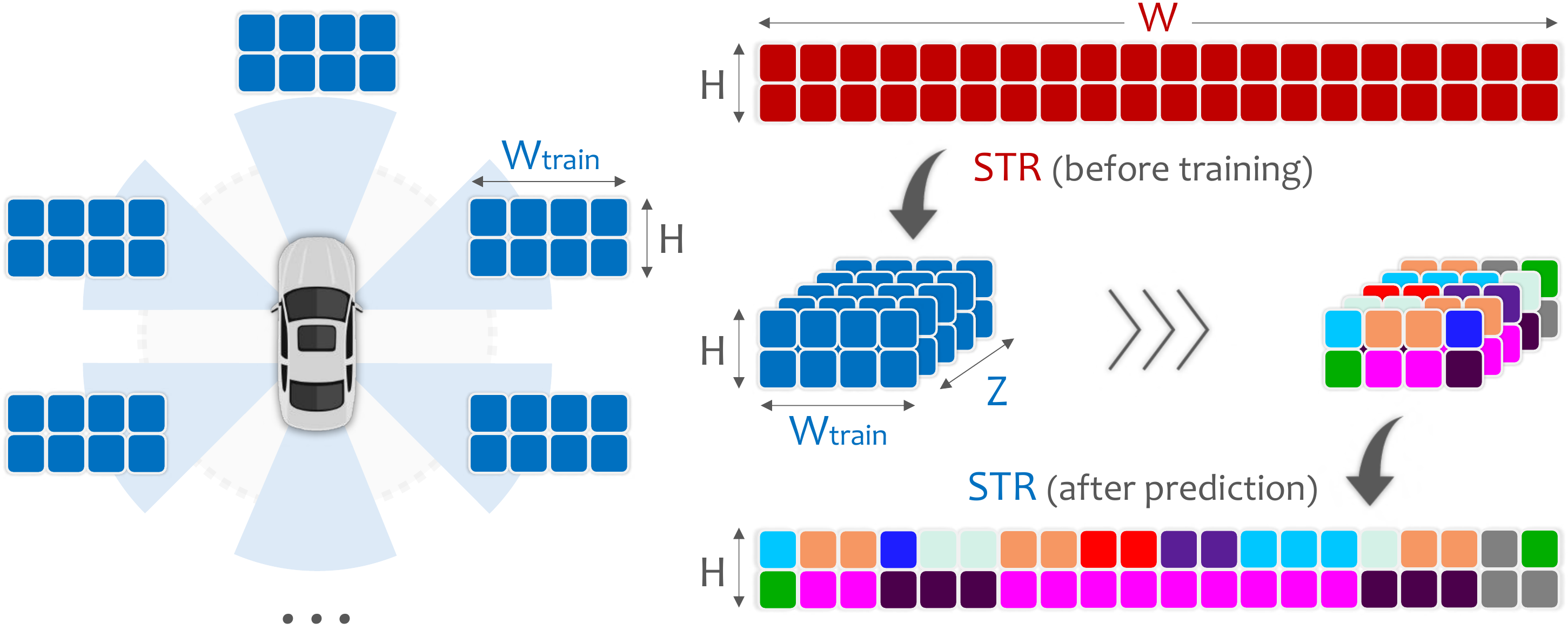}
    \end{center}
    \vspace{-0.37cm}
    \caption{Illustration of the proposed \textbf{STR paradigm}. We split LiDAR points into multiple ``views" (left) and rasterized them into range images with high horizontal angular resolutions (right). After training, the predictions are concatenated sequentially to form the complete LiDAR scan.}
    \label{figure:scalable}
\end{figure}

\noindent\textbf{Multi-View Partition}. To maintain the relatively high resolution of $W$ while pursuing efficiency at the same time, we propose a ``divide-and-conquer" learning paradigm. Specifically, we first partition points in the LiDAR scan into multiple groups based on the unique azimuth angle of each point, \textit{i.e.}, $\theta_i=\arctan(p^y_i/p^x_i)$. This will constitute $Z$ non-overlapping ``views" of the complete $360^{\circ}$ range view panorama as shown in \cref{figure:scalable}, where $Z$ is a hyperparameter and determines the total number of groups to be split. Next, points from each group will be rasterized separately with a high horizontal resolution to mitigate ``many-to-one" and deformation issues. In this way, the actual horizontal training resolution of the range image is eased by $Z$ times, \textit{i.e.}, $W_\text{train}=\frac{W}{Z}$, while the granularity ($\#$ of grids) of the range view projection in each ``view" is perfectly maintained.

\noindent\textbf{Training \& Inference}. During training, for each LiDAR scan, we randomly select only one of the $Z$ point groups for rasterization. That is to say, the model will be trained with a batch of randomly sampled ``views" at each step. During inference, we rasterize all groups for a given scan and stack the range images along the batch dimension. All ``views" can now be inferred in a single pass and the predictions are then wrapped back to form the complete scan. Despite being an empirical design, we find this \textit{STR} paradigm highly scalable during training. The convergence rate of training from multiple ``views" tends to be consistent with the conventional training paradigm, \textit{i.e.}, \textit{STR} can achieve competitive results using the \textit{same} number of iterations, while the memory consumption has now been reduced to only $\frac{1}{Z}$, which liberates the use of \textit{small-memory} GPUs for training.

\section{Experimental Analysis}
\label{sec:experiments}

\subsection{Settings}
\label{sec:settings}

\noindent\textbf{Benchmarks}. 
We conduct experiments on three standard LiDAR segmentation datasets. \textbf{\textit{SemanticKITTI}}  \cite{SemanticKITTI} provides $22$ sequences with $19$ semantic classes, captured by a 64-beam LiDAR sensor. Sequences $00$ to $10$ (\textit{exc.} $08$), $08$, and $11$ to $21$ are used for training, validation, and testing, respectively. \textbf{\textit{nuScenes}} \cite{Panoptic-nuScenes} consists of $1000$ driving scenes collected from Boston and Singapore, which are sparser due to the use of a 32-beam sensor. $16$ classes are adopted after merging similar and infrequent classes. \textbf{\textit{ScribbleKITTI}} \cite{2022ScribbleKITTI} shares the exact same data configurations with \cite{SemanticKITTI} but is weakly annotated with line scribbles, which corresponds to around $8.06\%$ semantic labels available during training. 

\begin{table*}[t]
\caption{Comparisons among state-of-the-art LiDAR \textbf{range view} semantic segmentation approaches with different spatial resolutions ($512$, $1024$, and $2048$) on the \textit{test} set of SemanticKITTI~\cite{SemanticKITTI}. All IoU scores are given in percentage (\%). For each resolution block: \textbf{bold} - best in column; \underline{underline} - second best in column. Symbol $^{\dagger}$: $W_{\text{train}}=384$.}
\vspace{-0.2cm}
\centering\scalebox{0.637}{
\begin{tabular}{c|r|c|ccccccccccccccccccc}
\toprule
\# & \textbf{Method~\small{(year)}} & \rotatebox{0}{$\text{mIoU}$} & \rotatebox{0}{car} & \rotatebox{0}{bicy} & \rotatebox{0}{moto} & \rotatebox{0}{truc} & \rotatebox{0}{o.veh} & \rotatebox{0}{ped} & \rotatebox{0}{b.list} & \rotatebox{0}{m.list} & \rotatebox{0}{road} & \rotatebox{0}{park} & \rotatebox{0}{walk} & \rotatebox{0}{o.gro} & \rotatebox{0}{build} & \rotatebox{0}{fenc} & \rotatebox{0}{veg} & \rotatebox{0}{trun} & \rotatebox{0}{terr} & \rotatebox{0}{pole} & \rotatebox{0}{sign}
\\\midrule
\multirow{5}{*}{\rotatebox{90}{$64\times512$}} & RangeNet++~\cite{milioto2019rangenet++}~\small{['19]} & $41.9$ & $87.4$ & $26.2$ & $26.5$ & $18.6$ & $15.6$ & $31.8$ & $33.6$ & $4.0$ & $\mathbf{91.4}$ & $57.0$ & $74.0$ & $26.4$ & $81.9$ & $52.3$ & $77.6$ & $48.4$ & $63.6$ & $36.0$ & $50.0$
\\
& MPF~\cite{2021MPF}~\small{['21]} & $48.9$ & $91.1$ & $22.0$ & $19.7$ & $18.8$ & $16.5$ & $30.0$ & $36.2$ & $4.2$ & $91.1$ & $61.9$ & $74.1$ & $29.4$ & $86.7$ & $56.2$ & \underline{$82.3$} & $51.6$ & \underline{$68.9$} & $38.6$ & $49.8$
\\
& FIDNet~\cite{zhao2021fidnet}~\small{['21]} & $51.3$ & $90.4$ & $28.6$ & $30.9$ & $34.3$ & $27.0$ & $43.9$ & $48.9$ & $16.8$ & $90.1$ & $58.7$ & $71.4$ & $19.9$ & $84.2$ & $51.2$ & $78.2$ & $51.9$ & $64.5$ & $32.7$ & $50.3$
\\
& CENet~\cite{cheng2022cenet}~\small{['22]} & \underline{$60.7$} & \underline{$92.1$} & \underline{$45.4$} & \underline{$42.9$} & \underline{$43.9$} & \underline{$46.8$} & \underline{$56.4$} & \underline{$63.8$} & \underline{$29.7$} & \underline{$91.3$} & \underline{$66.0$} & \underline{$75.3$} & \underline{$31.1$} & \underline{$88.9$} & \underline{$60.4$} & $81.9$ & \underline{$60.5$} & $67.6$ & \underline{$49.5$} & \underline{$59.1$}
\\
& \cellcolor{LightCyan}\textbf{RangeFormer} & \cellcolor{LightCyan}$\mathbf{70.0}$ & \cellcolor{LightCyan}$\mathbf{94.7}$ & \cellcolor{LightCyan}$\mathbf{60.5}$ & \cellcolor{LightCyan}$\mathbf{70.2}$ & \cellcolor{LightCyan}$\mathbf{58.4}$ & \cellcolor{LightCyan}$\mathbf{64.6}$ & \cellcolor{LightCyan}$\mathbf{72.8}$ & \cellcolor{LightCyan}$\mathbf{73.0}$ & \cellcolor{LightCyan}$\mathbf{55.4}$ & \cellcolor{LightCyan}$90.8$ & \cellcolor{LightCyan}$\mathbf{70.4}$ & \cellcolor{LightCyan}$\mathbf{75.4}$ & \cellcolor{LightCyan}$\mathbf{39.9}$ & \cellcolor{LightCyan}$\mathbf{90.7}$ & \cellcolor{LightCyan}$\mathbf{66.6}$ & \cellcolor{LightCyan}$\mathbf{84.6}$ & \cellcolor{LightCyan}$\mathbf{68.6}$ & \cellcolor{LightCyan}$\mathbf{70.5}$ & \cellcolor{LightCyan}$\mathbf{59.4}$ & \cellcolor{LightCyan}$\mathbf{63.6}$
\\\midrule
\multirow{5}{*}{\rotatebox{90}{$64\times1024$}} & RangeNet++~\cite{milioto2019rangenet++}~\small{['19]} & $48.0$ & $90.3$ & $20.6$ & $27.1$ & $25.2$ & $17.6$ & $29.6$ & $34.2$ & $7.1$ & $90.4$ & $52.3$ & $72.7$ & $22.8$ & $83.9$ & $53.3$ & $77.7$ & $52.5$ & $63.7$ & $43.8$ & $47.2$
\\
& MPF~\cite{2021MPF}~\small{['21]} & $53.6$ & $92.7$ & $28.2$ & $30.5$ & $26.9$ & $25.2$ & $42.5$ & $45.5$ & $9.5$ & $90.5$ & $64.7$ & \underline{$74.3$} & \underline{$32.0$} & $88.3$ & $59.0$ & \underline{$83.4$} & $56.6$ & \underline{$69.8$} & $46.0$ & $54.9$
\\
& FIDNet~\cite{zhao2021fidnet}~\small{['21]} & $56.0$ & $92.4$ & $44.0$ & $41.5$ & $33.2$ & $30.8$ & $57.9$ & $52.6$ & $18.0$ & \underline{$91.0$} & $61.2$ & $73.8$ & $12.6$ & $88.2$ & $57.9$ & $80.8$ & $59.5$ & $65.1$ & $45.3$ & $58.4$
\\
& CENet~\cite{cheng2022cenet}~\small{['22]} & \underline{$62.3$} & \underline{$93.0$} & \underline{$50.5$} & \underline{$47.6$} & \underline{$41.7$} & \underline{$43.4$} & \underline{$64.5$} & \underline{$65.2$} & \underline{$32.5$} & $90.5$ & \underline{$65.5$} & $74.1$ & $29.2$ & \underline{$90.9$} & \underline{$65.4$} & $81.6$ & \underline{$65.4$} & $65.6$ & \underline{$55.9$} & \underline{$61.0$}
\\
& \cellcolor{LightCyan}\textbf{RangeFormer} & \cellcolor{LightCyan}$\mathbf{72.1}$ & \cellcolor{LightCyan}$\mathbf{95.7}$ & \cellcolor{LightCyan}$\mathbf{66.2}$ & \cellcolor{LightCyan}$\mathbf{72.9}$ & \cellcolor{LightCyan}$\mathbf{59.8}$ & \cellcolor{LightCyan}$\mathbf{66.5}$ & \cellcolor{LightCyan}$\mathbf{75.8}$ & \cellcolor{LightCyan}$\mathbf{74.5}$ & \cellcolor{LightCyan}$\mathbf{56.5}$ & \cellcolor{LightCyan}$\mathbf{91.8}$ & \cellcolor{LightCyan}$\mathbf{71.9}$ & \cellcolor{LightCyan}$\mathbf{77.4}$ & \cellcolor{LightCyan}$\mathbf{41.6}$ & \cellcolor{LightCyan}$\mathbf{91.6}$ & \cellcolor{LightCyan}$\mathbf{68.9}$ & \cellcolor{LightCyan}$\mathbf{85.8}$ & \cellcolor{LightCyan}$\mathbf{71.5}$ & \cellcolor{LightCyan}$\mathbf{71.6}$ & \cellcolor{LightCyan}$\mathbf{64.2}$ & \cellcolor{LightCyan}$\mathbf{65.8}$
\\\midrule
\multirow{14}{*}{\rotatebox{90}{$64\times2048$}}
& SqSeg~\cite{wu2018squeezeseg}~\small{['18]} & $30.8$ & $68.3$ & $18.1$ & $5.1$ & $4.1$ & $4.8$ & $16.5$ & $17.3$ & $1.2$ & $84.9$ & $28.4$ & $54.7$ & $4.6$ & $61.5$ & $29.2$ & $59.6$ & $25.5$ & $54.7$ & $11.2$ & $36.3$
\\
& SqSegV2~\cite{wu2019squeezesegv2}~\small{['19]} & $39.6$ & $82.7$ & $21.0$ & $22.6$ & $14.5$ & $15.9$ & $20.2$ & $24.3$ & $2.9$ & $88.5$ & $42.4$ & $65.5$ & $18.7$ & $73.8$ & $41.0$ & $68.5$ & $36.9$ & $58.9$ & $12.9$ & $41.0$
\\
& RangeNet++~\cite{milioto2019rangenet++}~\small{['19]} & $52.2$ & $91.4$ & $25.7$ & $34.4$ & $25.7$ & $23.0$ & $38.3$ & $38.8$ & $4.8$ & $91.8$ & $65.0$ & $75.2$ & $27.8$ & $87.4$ & $58.6$ & $80.5$ & $55.1$ & $64.6$ & $47.9$ & $55.9$
\\
& SqSegV3~\cite{xu2020squeezesegv3}~\small{['20]} & $55.9$ & $92.5$ & $38.7$ & $36.5$ & $29.6$ & $33.0$ & $45.6$ & $46.2$ & $20.1$ & $91.7$ & $63.4$ & $74.8$ & $26.4$ & $89.0$ & $59.4$ & $82.0$ & $58.7$ & $65.4$ & $49.6$ & $58.9$
\\
& 3D-MiniNet~\cite{alonso20203d-mininet}~\small{['20]} & $55.8$ & $90.5$ & $42.3$ & $42.1$ & $28.5$ & $29.4$ & $47.8$ & $44.1$ & $14.5$ & $91.6$ & $64.2$ & $74.5$ & $25.4$ & $89.4$ & $60.8$ & $82.8$ & $60.8$ & $66.7$ & $48.0$ & $56.6$
\\
& SalsaNext~\cite{cortinhal2020salsanext}~\small{['20]} & $59.5$ & $91.9$ & $48.3$ & $38.6$ & $38.9$ & $31.9$ & $60.2$ & $59.0$ & $19.4$ & $91.7$ & $63.7$ & $75.8$ & $29.1$ & $90.2$ & $64.2$ & $81.8$ & $63.6$ & $66.5$ & $54.3$ & $62.1$
\\
& KPRNet~\cite{kochanov2020kprnet}~\small{['21]} & $63.1$ & \underline{$95.5$} & $54.1$ & $47.9$ & $23.6$ & $42.6$ & $65.9$ & $65.0$ & $16.5$ & $\mathbf{93.2}$ & $\mathbf{73.9}$ & $\mathbf{80.6}$ & $30.2$ & $91.7$ & $68.4$ & \underline{$85.7$} & $69.8$ & \underline{$71.2$} & $58.7$ & $64.1$
\\
& LiteHDSeg~\cite{razani2021lite}~\small{['21]} & $63.8$ & $92.3$ & $40.0$ & $55.4$ & $37.7$ & $39.6$ & $59.2$ & \underline{$71.6$} & \underline{$54.3$} & $93.0$ & $68.2$ & $78.3$ & $29.3$ & $91.5$ & $65.0$ & $78.2$ & $65.8$ & $65.1$ & $59.5$ & $\mathbf{67.7}$
\\
& MPF~\cite{2021MPF}~\small{['21]} & $55.5$ & $93.4$ & $30.2$ & $38.3$ & $26.1$ & $28.5$ & $48.1$ & $46.1$ & $18.1$ & $90.6$ & $62.3$ & $74.5$ & $30.6$ & $88.5$ & $59.7$ & $83.5$ & $59.7$ & $69.2$ & $49.7$ & $58.1$
\\
& FIDNet~\cite{zhao2021fidnet}~\small{['21]} & $59.5$ & $93.9$ & $54.7$ & $48.9$ & $27.6$ & $23.9$ & $62.3$ & $59.8$ & $23.7$ & $90.6$ & $59.1$ & $75.8$ & $26.7$ & $88.9$ & $60.5$ & $84.5$ & $64.4$ & $69.0$ & $53.3$ & $62.8$
\\
& RangeViT~\cite{2023RangeViT}~\small{['23]} & $64.0$ & $95.4$ & $55.8$ & $43.5$ & $29.8$ & $42.1$ & $63.9$ & $58.2$ & $38.1$ & \underline{$93.1$} & $70.2$ & \underline{$80.0$} & \underline{$32.5$} & \underline{$92.0$} & \underline{$69.0$} & $85.3$ & \underline{$70.6$} & \underline{$71.2$} & $60.8$ & $64.7$
\\
& CENet~\cite{cheng2022cenet}~\small{['22]} & $64.7$ & $91.9$ & \underline{$58.6$} & $50.3$ & $40.6$ & $42.3$ & \underline{$68.9$} & $65.9$ & $43.5$ & $90.3$ & $60.9$ & $75.1$ & $31.5$ & $91.0$ & $66.2$ & $84.5$ & $69.7$ & $70.0$ & \underline{$61.5$} & \underline{$67.6$}
\\
& MaskRange~\cite{2022MaskRange}~\small{['22]} & \underline{$66.1$} & $94.2$ & $56.0$ & \underline{$55.7$} & \underline{$59.2$} & \underline{$52.4$} & $67.6$ & $64.8$ & $31.8$ & $91.7$ & $70.7$ & $77.1$ & $29.5$ & $90.6$ & $65.2$ & $84.6$ & $68.5$ & $69.2$ & $60.2$ & $66.6$ 
\\
& \cellcolor{LightCyan}\textbf{RangeFormer} & \cellcolor{LightCyan}$\mathbf{73.3}$ & \cellcolor{LightCyan}$\mathbf{96.7}$ & \cellcolor{LightCyan}$\mathbf{69.4}$ & \cellcolor{LightCyan}$\mathbf{73.7}$ & \cellcolor{LightCyan}$\mathbf{59.9}$ & \cellcolor{LightCyan}$\mathbf{66.2}$ & \cellcolor{LightCyan}$\mathbf{78.1}$ & \cellcolor{LightCyan}$\mathbf{75.9}$ & \cellcolor{LightCyan}$\mathbf{58.1}$ & \cellcolor{LightCyan}$92.4$ & \cellcolor{LightCyan}\underline{$73.0$} & \cellcolor{LightCyan}$78.8$ & \cellcolor{LightCyan}$\mathbf{42.4}$ & \cellcolor{LightCyan}$\mathbf{92.3}$ & \cellcolor{LightCyan}$\mathbf{70.1}$ & \cellcolor{LightCyan}$\mathbf{86.6}$ & \cellcolor{LightCyan}$\mathbf{73.3}$ & \cellcolor{LightCyan}$\mathbf{72.8}$ & \cellcolor{LightCyan}$\mathbf{66.4}$ & \cellcolor{LightCyan}$66.6$
\\\midrule
\multirow{3}{*}{\rotatebox{90}{STR$^{\dagger}$}}
& \cellcolor{red!4}\textbf{FIDNet \textit{w/} STR} & \cellcolor{red!4}$60.1$ & \cellcolor{red!4}\underline{$93.6$} & \cellcolor{red!4}$48.8$ & \cellcolor{red!4}$44.4$ & \cellcolor{red!4}\underline{$45.0$} & \cellcolor{red!4}$38.4$ & \cellcolor{red!4}$58.1$ & \cellcolor{red!4}$65.5$ & \cellcolor{red!4}$7.0$ & \cellcolor{red!4}$\mathbf{92.2}$ & \cellcolor{red!4}$68.3$ & \cellcolor{red!4}$76.2$ & \cellcolor{red!4}$27.4$ & \cellcolor{red!4}$88.1$ & \cellcolor{red!4}$61.3$ & \cellcolor{red!4}$82.8$ & \cellcolor{red!4}$61.0$ & \cellcolor{red!4}$69.5$ & \cellcolor{red!4}$55.6$ & \cellcolor{red!4}$58.4$
\\
& \cellcolor{red!4}\textbf{CENet \textit{w/} STR} & \cellcolor{red!4}\underline{$65.8$} & \cellcolor{red!4}\underline{$93.6$} & \cellcolor{red!4}\underline{$60.2$} & \cellcolor{red!4}\underline{$60.0$} & \cellcolor{red!4}$43.5$ & \cellcolor{red!4}\underline{$47.4$} & \cellcolor{red!4}\underline{$69.4$} & \cellcolor{red!4}\underline{$67.6$} & \cellcolor{red!4}\underline{$19.7$} & \cellcolor{red!4}$92.0$ & \cellcolor{red!4}\underline{$70.2$} & \cellcolor{red!4}\underline{$77.6$} & \cellcolor{red!4}$\mathbf{43.6}$ & \cellcolor{red!4}\underline{$90.2$} & \cellcolor{red!4}\underline{$66.9$} & \cellcolor{red!4}\underline{$84.7$} & \cellcolor{red!4}\underline{$66.2$} & \cellcolor{red!4}\underline{$71.3$} & \cellcolor{red!4}\underline{$60.5$} & \cellcolor{red!4}$\mathbf{65.4}$
\\
& \cellcolor{red!4}\textbf{RangeFormer \textit{w/} STR} & \cellcolor{red!4}$\mathbf{72.2}$ & \cellcolor{red!4}$\mathbf{96.4}$ & \cellcolor{red!4}$\mathbf{67.1}$ & \cellcolor{red!4}$\mathbf{72.2}$ & \cellcolor{red!4}$\mathbf{58.8}$ & \cellcolor{red!4}$\mathbf{67.4}$ & \cellcolor{red!4}$\mathbf{74.9}$ & \cellcolor{red!4}$\mathbf{74.7}$ & \cellcolor{red!4}$\mathbf{57.5}$ & \cellcolor{red!4}\underline{$92.1$} & \cellcolor{red!4}$\mathbf{72.5}$ & \cellcolor{red!4}$\mathbf{78.2}$ & \cellcolor{red!4}\underline{$42.4$} & \cellcolor{red!4}$\mathbf{91.8}$ & \cellcolor{red!4}$\mathbf{69.7}$ & \cellcolor{red!4}$\mathbf{85.8}$ & \cellcolor{red!4}$\mathbf{70.4}$ & \cellcolor{red!4}$\mathbf{72.3}$ & \cellcolor{red!4}$\mathbf{62.8}$ & \cellcolor{red!4}\underline{$65.0$}
\\\bottomrule
\end{tabular}}
\vspace{0.1cm}
\label{table:semantickitti}
\end{table*}
\begin{table*}
\begin{minipage}[b]{.413\linewidth}
    \centering
    \includegraphics[width=0.987\linewidth]{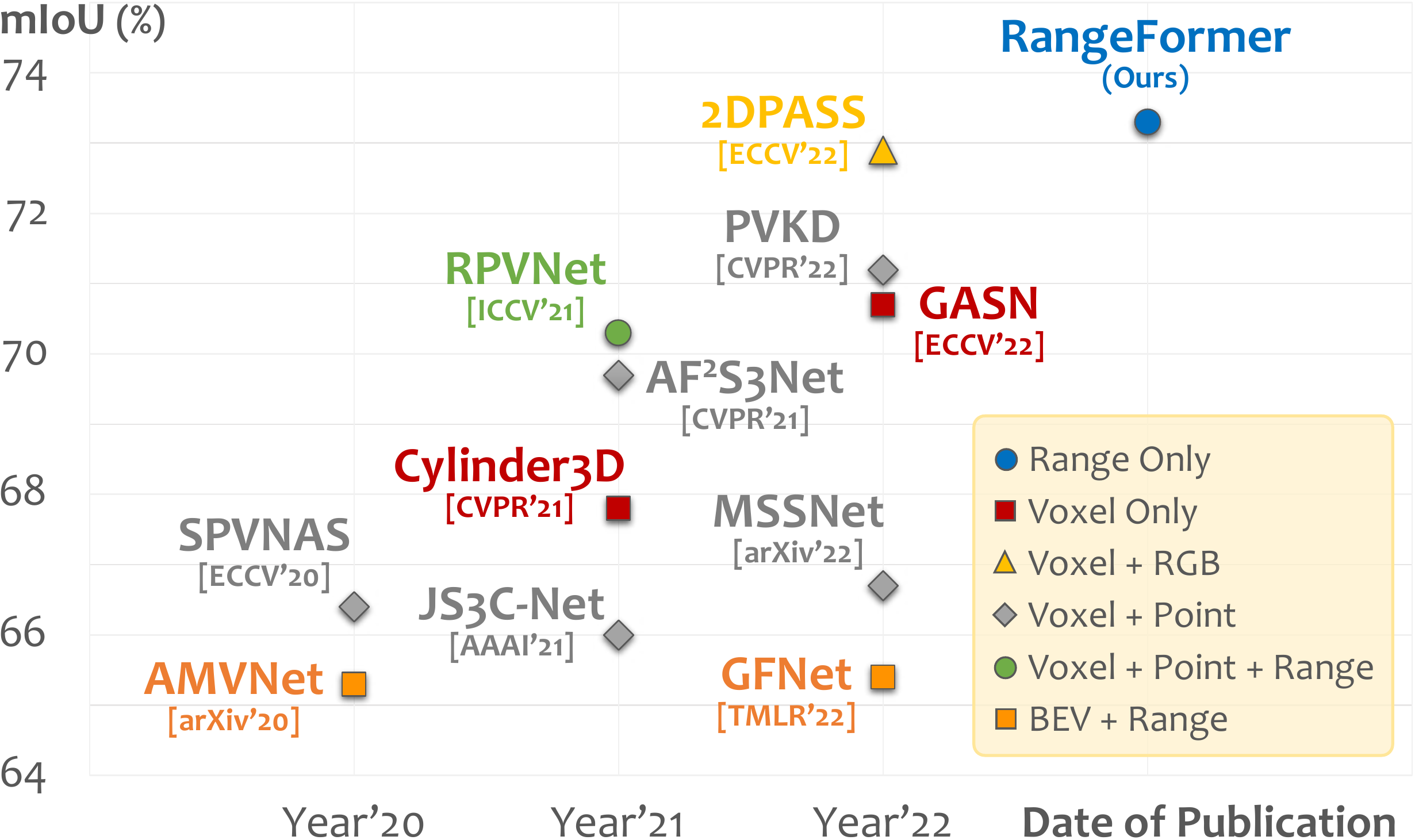}
    \caption{State-of-the-art LiDAR semantic segmentation approaches on the \textit{test} set of SemanticKITTI~\cite{SemanticKITTI}.}
    \label{figure:semantickitti}
\end{minipage}~~~
\begin{minipage}[b]{.573\linewidth}
    \centering
    \caption{Comparisons among state-of-the-art LiDAR \textbf{panoptic segmentation} methods on the \textit{test} set of SemanticKITTI~\cite{SemanticKITTI}. All scores are given in percentage (\%). For each metric: \textbf{bold} - best in column; \underline{underline} - second best in column. RN denotes RangeNet++~\cite{milioto2019rangenet++}. PP denotes PointPillars~\cite{lang2019pointpillars}. Symbol $^{\dagger}$: $W_{\text{train}}=384$.}
    \vspace{-0.2cm}
    \centering\scalebox{0.637}{
    \begin{tabular}{r|cccc|ccc|ccc|c}
    \toprule
    \textbf{Method} & $\text{PQ}$ & $\text{PQ}^\dagger$ & $\text{RQ}$ & $\text{SQ}$ & $\text{PQ}^{\text{Th}}$ & $\text{RQ}^{\text{Th}}$ & $\text{SQ}^{\text{Th}}$ & $\text{PQ}^{\text{St}}$ & $\text{RQ}^{\text{St}}$ & $\text{SQ}^{\text{St}}$ & $\text{mIoU}$
    \\\midrule
    RN + PP & $37.1$ & $45.9$ & $47.0$ & $75.9$ & $20.2$ & $25.2$ & $75.2$ & $49.3$ & $62.8$ & $76.5$ & $52.4$
    \\
    KPConv + PP & $44.5$ & $52.5$ & $54.4$ & $80.0$ & $32.7$ & $38.7$ & $81.5$ & $53.1$ & $65.9$ & $79.0$ & $58.8$ 
    \\
    Panoster~\cite{gasperini2020panoster} & $52.7$ & $59.9$ & $64.1$ & $80.7$ & $49.4$ & $58.5$ & $83.3$ & $55.1$ & $68.2$ & $78.8$ & $59.9$
    \\
    MaskRange~\cite{2022MaskRange} & $53.1$ & $59.2$ & $64.6$ & $81.2$ & $44.9$ & $53.0$ & $83.5$ & $59.1$ & $73.1$ & $79.5$ & $61.8$
    \\
    P-PolarNet~\cite{zhou2021panoptic} & $54.1$ & $60.7$ & $65.0$ & $81.4$ & $53.3$ & $60.6$ & $87.2$ & $54.8$ & $68.1$ & $77.2$ & $59.5$
    \\
    DS-Net~\cite{DS-Net} & $55.9$ & $62.5$ & $66.7$ & $82.3$ & $55.1$ & $62.8$ & \underline{$87.2$} & $56.5$ & $69.5$ & $78.7$ & $61.6$
    \\
    EfficientLPS~\cite{sirohi2021efficientlps} & $57.4$ & $63.2$ & $68.7$ & $83.0$ & $53.1$ & $60.5$ & $87.8$ & $60.5$ & $74.6$ & $79.5$ & $61.4$
    \\
    P-PHNet~\cite{Panoptic-PHNet} & $61.5$ & \underline{$67.9$} & $72.1$ & $\mathbf{84.8}$ & $\mathbf{63.8}$ & \underline{$70.4$} & $\mathbf{90.7}$ & $59.9$ & $73.3$ & $80.5$ & $66.0$
    \\
    \cellcolor{LightCyan}\textbf{P-RangeFormer} & \cellcolor{LightCyan}$\mathbf{64.2}$ & \cellcolor{LightCyan}$\mathbf{69.5}$ & \cellcolor{LightCyan}$\mathbf{75.9}$ & \cellcolor{LightCyan}\underline{$83.8$} & \cellcolor{LightCyan}\underline{$63.6$} & \cellcolor{LightCyan}$\mathbf{73.0}$ & \cellcolor{LightCyan}$86.8$ & \cellcolor{LightCyan}$\mathbf{64.6}$ & \cellcolor{LightCyan}$\mathbf{78.1}$ & \cellcolor{LightCyan}$\mathbf{81.7}$ & \cellcolor{LightCyan}$\mathbf{72.0}$
    \\\midrule
    \cellcolor{red!4}\textbf{\textit{w/} STR}$^{\dagger}$ & \cellcolor{red!4}\underline{$61.8$} & \cellcolor{red!4}$67.6$ & \cellcolor{red!4}\underline{$73.8$} & \cellcolor{red!4}$83.1$ & \cellcolor{red!4}$60.3$ & \cellcolor{red!4}$69.6$ & \cellcolor{red!4}$86.3$ & \cellcolor{red!4}\underline{$62.9$} & \cellcolor{red!4}\underline{$76.8$} & \cellcolor{red!4}\underline{$80.8$} & \cellcolor{red!4}\underline{$71.0$}
    \\\bottomrule
    \end{tabular}}
    \vspace{-0.cm}
    \label{table:semantickitti-panoptic}
\end{minipage}
\end{table*}

\noindent\textbf{Evaluation Metrics}. Following the standard practice, we report the Intersection-over-Union (IoU) for class $i$ and the average score (mIoU) over all classes, where $\text{IoU}_{i} = \frac{\text{TP}_{i}}{\text{TP}_{i} + \text{FP}_{i} + \text{FN}_{i}}$. $\text{TP}_{i}$, $\text{FP}_{i}$ and $\text{FN}_{i}$ are the true-positive, false-positive, and false-negative. For panoptic segmentation, the models are measured by the Panoptic Quality (PQ)~\cite{kirillov2019panoptic}
\begin{equation}
    \text{PQ} = \underbrace{\frac{\sum_{(i, j)\in \text{TP}}\text{IoU}(i,j)}{|\text{TP}|}}_\text{SQ} \times \underbrace{\frac{|\text{TP}|}{|\text{TP}| + \frac{1}{2}(|\text{FP}| + |\text{FN}|)}}_\text{RQ},
\end{equation}
which consists of Segmentation Quality (SQ) and Recognition Quality (RQ). We also report the separated scores for \textit{things} and \textit{stuff} classes, \textit{i.e.},
\PQth{}, \SQth{}, \RQth{}, and \PQst{}, \SQst{}, \RQst{}.
\PQda{} is defined by swapping the \PQ{} of each \textit{stuff} class to its IoU then averaging over all classes~\cite{porzi2019seamless}.

\noindent \textbf{Network Configurations}.
After range view rasterization, the input $\mathcal{R}(u,v)$ of size $6\times H\times W$ is first fed into REM for range view point embedding. It consists of three MLP layers that map the embedding dim of $\mathcal{R}(u,v)$ from $6$ to $64$, $128$, and $128$, respectively, with the batch norm and GELU activation.
The output of size $128\times H\times W$ from REM serves as the input of the Transformer blocks. Specifically, for each of the four stages, the patch embedding layer divides an input of size $H_\text{embed}, W_\text{embed}$ into $3\times 3$ patches with overlap stride equals to $1$ (for the first stage) and $2$ (for the last three stages). 
After the overlap patch embedding, the patches are processed with the standard multi-head attention operations as in \cite{2021_ViT,2021_PVT,2021segformer}. We keep the default setting of using the residual connection and layer normalization (Add \& Norm). The number of heads for each of the four stages is $[3, 4, 6, 3]$. The hierarchical features extracted from different stages are stored and used for decoding. Specifically, each of the four stages produces features of spatial size $[(H, W), (\frac{H}{2}, \frac{W}{2}), (\frac{H}{4}, \frac{W}{4}), (\frac{H}{8}, \frac{W}{8})]$, with the channel dimension of $[128, 128, 320, 512]$. As described in previous sections, we perform two unification steps to unify the channel and spatial sizes of different feature maps. We first map their channel dimensions to $256$, \textit{i.e.}, $[128, H, W]\rightarrow[256, H, W]$ for stage $1$, $[128, \frac{H}{2}, \frac{W}{2}]\rightarrow[256, \frac{H}{2}, \frac{W}{2}]$ for stage $2$, $[320, \frac{H}{4}, \frac{W}{4}]\rightarrow[256, \frac{H}{4}, \frac{W}{4}]$ for stage $3$, and $[512, \frac{H}{8}, \frac{W}{8}]\rightarrow[256, \frac{H}{8}, \frac{W}{8}]$ for stage $4$. We then interpolate four feature maps to the spatial size of $H\times W$. 
The probabilities of conducting the four augmentations in \textit{RangeAug} are set as $[0.9, 0.2, 0.9, 1.0]$. For \textit{RangePost}, we divide the whole scan into three ``sub-clouds" for the 2D-to-3D re-rasterization.

\noindent \textbf{Implementation Details}.
Following the conventional settings \cite{milioto2019rangenet++,cheng2022cenet}, we conduct experiments with $W_{\text{train}}=512, 1024, 2048$ on SemanticKITTI \cite{SemanticKITTI} and $W_{\text{train}}=1920$ on nuScenes \cite{Panoptic-nuScenes}. We use the AdamW optimizer~\cite{AdamW} and OneCycle scheduler \cite{OneCycle} with $lr=$ $1$e-$3$. For \textit{STR} training, we first partition points into $5$ and $2$ views and then rasterize them into range images of size $64\times 1920$ ($W_{\text{train}}=384$) and of size $32\times 960$ ($W_{\text{train}}=480$), for SemanticKITTI \cite{SemanticKITTI} and nuScenes \cite{Panoptic-nuScenes}, respectively. The models are pre-trained on Cityscapes \cite{Cityscapes} for $20$ epochs and then trained for $60$ epochs on SemanticKITTI \cite{SemanticKITTI} and ScribbleKITTI \cite{2022ScribbleKITTI} and for $100$ epochs on nuScenes \cite{Panoptic-nuScenes}, respectively, with a batch size of $32$. Similar to \cite{razani2021lite,cheng2022cenet}, we include the cross-entropy dice loss, Lovasz-Softmax loss \cite{berman2018lovasz}, and boundary loss \cite{razani2021lite} to supervise the model training. All models can be trained on \textit{single} NVIDIA A100/V100 GPUs for around $32$ hours.

\subsection{Comparative Study}
\label{sec:comparative}

\noindent\textbf{Semantic Segmentation}.
Firstly, we compare the proposed \textit{RangeFormer} with $13$ prior and SoTA range view LiDAR segmentation methods on SemanticKITTI \cite{SemanticKITTI} (see \cref{table:semantickitti}). In conventional $512$, $1024$, and $2048$ settings, we observe $9.3\%$, $9.8\%$, and $8.6\%$ mIoU improvements over the SoTA method CENet \cite{cheng2022cenet} and $7.2\%$ mIoU higher than MaskRange \cite{2022MaskRange}. Such superiority is general for almost all classes and especially overt for dynamic and small-scale ones like \textit{bicycle} and \textit{motorcycle}. In \cref{figure:semantickitti}, we further compare \textit{RangeFormer} with $11$ methods from other modalities. We can see that the current trend favors fusion-based methods which often combine the point and voxel views~\cite{pvkd2022,2021AF2S3Net}. Albeit using only range view, \textit{RangeFormer} achieves the best scores so far; it surpasses the best fusion-based method 2DPASS~\cite{2022_2DPASS} by $0.4\%$ mIoU and the best voxel-only method GASN~\cite{2022GASN} by $2.9\%$ mIoU. Similar observations also hold for nuScenes \cite{Panoptic-nuScenes} (see \cref{table:latency}).

\noindent\textbf{STR Paradigm}. As can be seen from the last three rows of \cref{table:semantickitti}, under the \textit{STR} paradigm ($W_{\text{train}}=384$), FIDNet~\cite{zhao2021fidnet} and CENet~\cite{cheng2022cenet} have achieved even better scores compared to their high-resolution ($W_{\text{train}}=2048$) versions. \textit{RangeFormer} achieves $72.2\%$ mIoU with \textit{STR}, which is better than most of the methods on the leaderboard (see \cref{figure:semantickitti}) while being $13.5\%$ faster than the high training resolution (\textit{i.e.}, $2048$) option (see \cref{table:latency}) and saves $80\%$ memory consumption. It is worth highlighting again that the convergence rate tends not to be affected. The \textit{same} number of training epochs are applied to both \textit{STR} and conventional training to ensure that the comparison is accurate.

\begin{figure*}
    \begin{subfigure}{0.68\columnwidth}
      \centering
      \includegraphics[width=\linewidth]{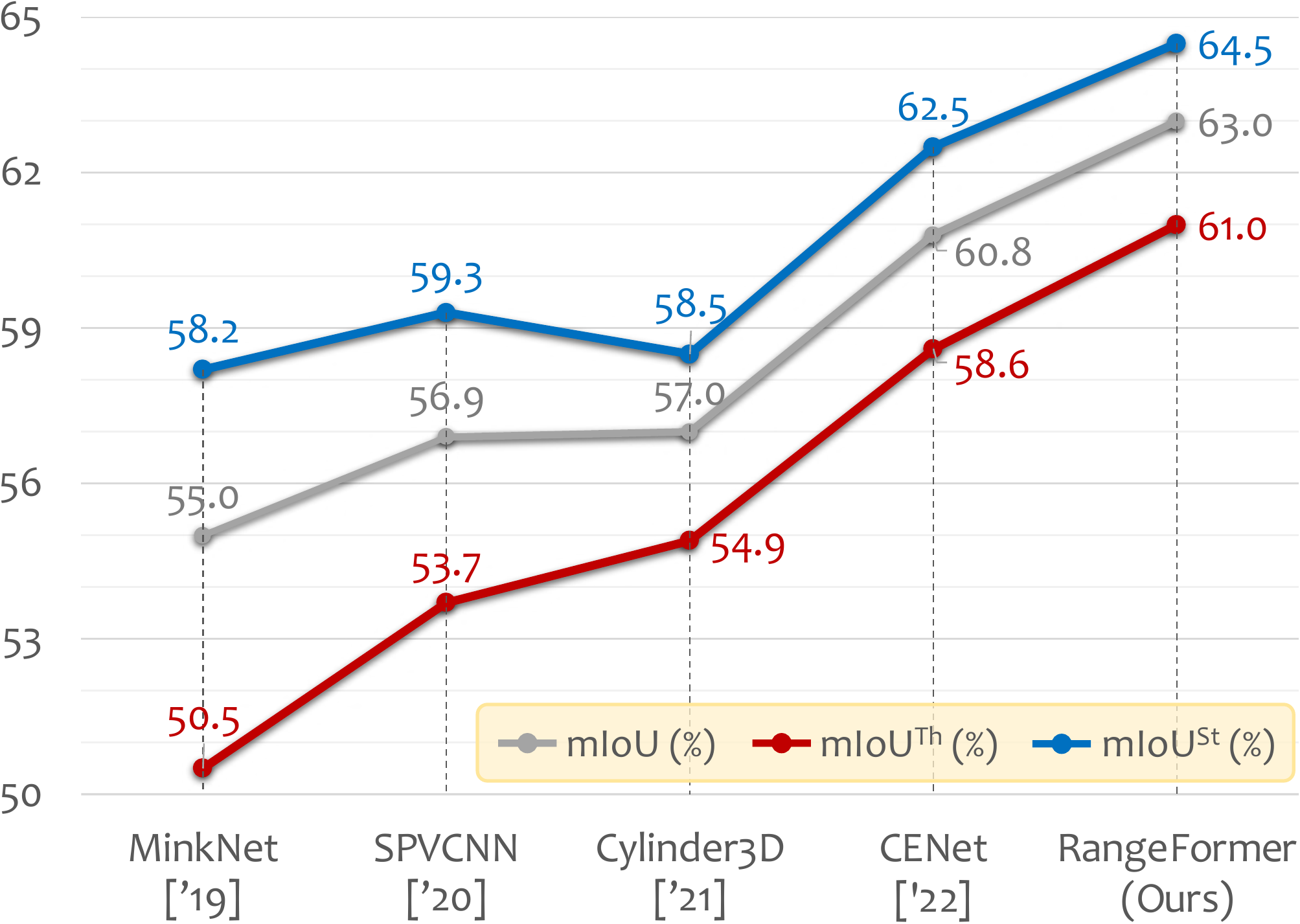}
      \caption{ScribbleKITTI Leaderboard}
      \label{figure:scribblekitti}
    \end{subfigure}~~
    \begin{subfigure}{0.68\columnwidth}
      \centering
      \includegraphics[width=\linewidth]{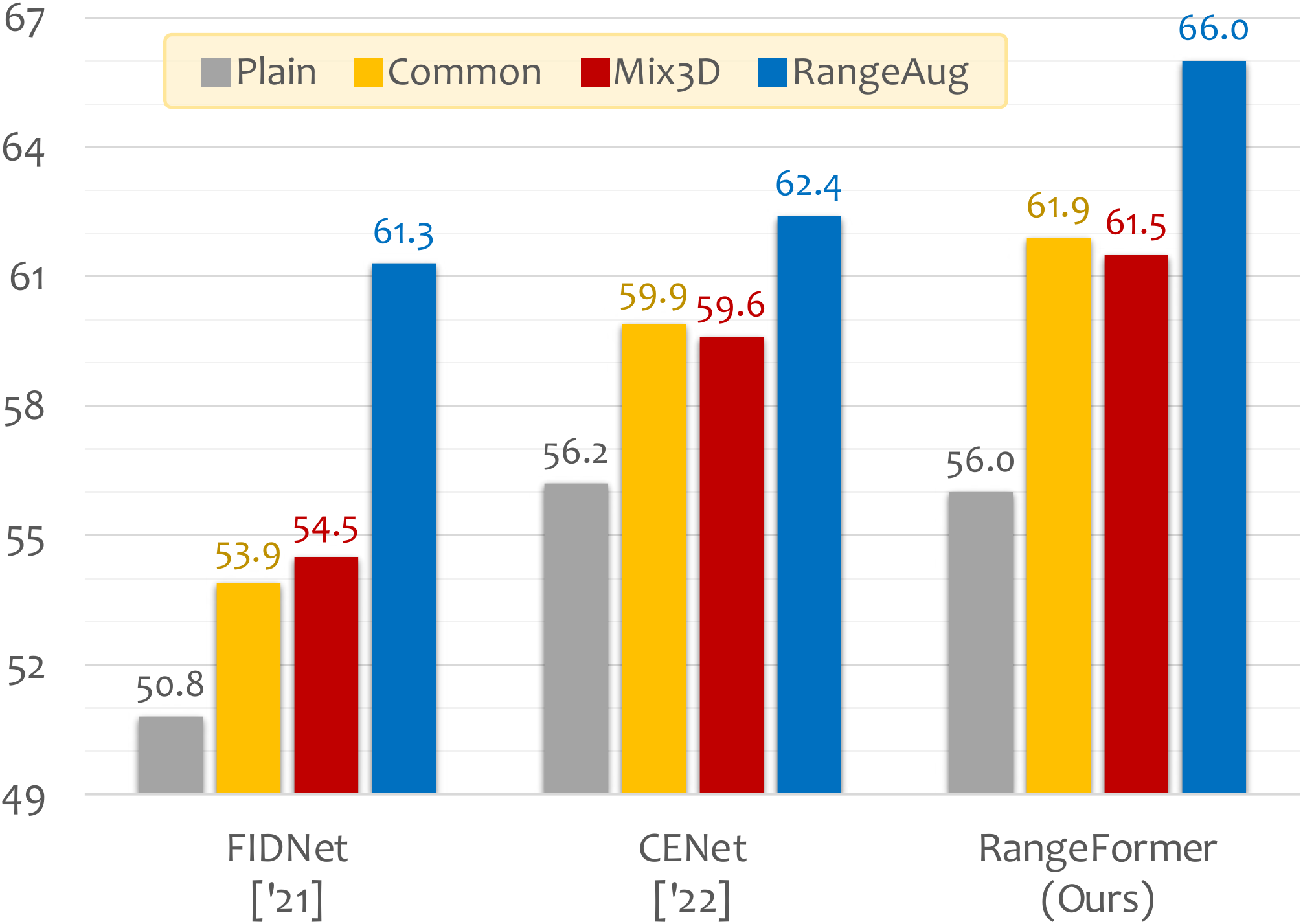}
      \caption{3D Augmentation}
      \label{figure:augmentation}
    \end{subfigure}~~
    \begin{subfigure}{0.68\columnwidth}
      \centering
      \includegraphics[width=\linewidth]{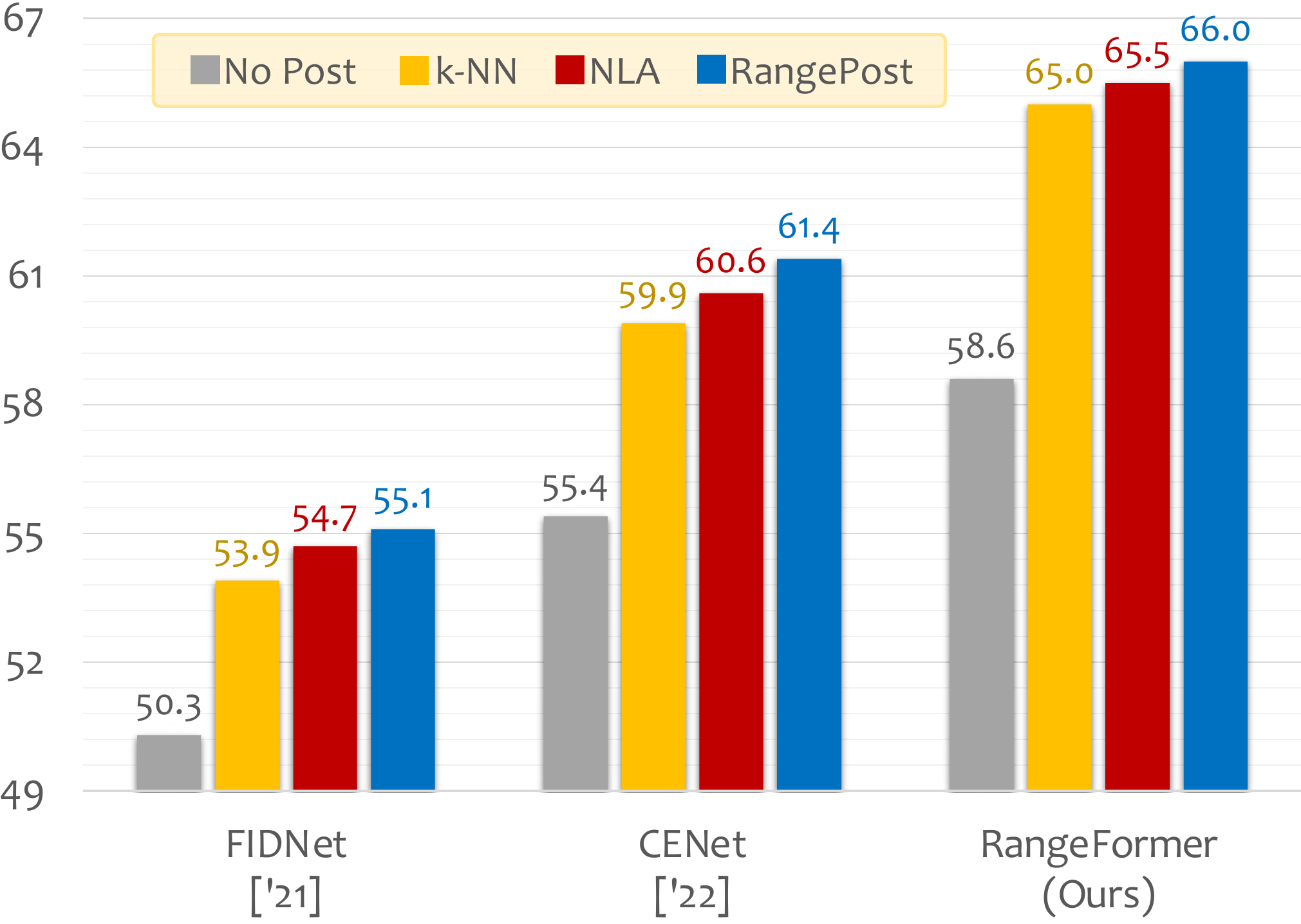}
      \caption{Post-Processing}
      \label{figure:post-processing}
    \end{subfigure}
\vspace{-0.5cm}
\caption{Comparative \& ablation study. (a) Weakly-supervised LiDAR semantic segmentation results on the \textit{val} set of ScribbleKITTI \cite{2022ScribbleKITTI} (the same as SemanticKITTI \cite{SemanticKITTI}). (b) Results of different 3D data augmentation approaches on the \textit{val} set of SemanticKITTI \cite{SemanticKITTI}. (c) Results of different post-processing methods on the \textit{val} set of SemanticKITTI \cite{SemanticKITTI}.}
\end{figure*}

\noindent\textbf{Panoptic Segmentation}. The advantages of \textit{RangeFormer} in semantic segmentation have further yielded better panoptic segmentation performance. From \cref{table:semantickitti-panoptic} we can see that \textit{Panoptic-RangeFormer} achieves better scores than the recent SoTA method Panoptic-PHNet~\cite{Panoptic-PHNet} in terms of $\text{PQ}$, $\text{PQ}^\dagger$, and $\text{RQ}$. Such superiority still holds under the \textit{STR} paradigm and is especially overt for the \textit{stuff} classes. The ability to unify both semantic and instance LiDAR segmentation further validates the scalability of our framework.

\noindent\textbf{Weakly-Supervised Segmentation}. Recently, \cite{2022ScribbleKITTI} adopts line scribbles to label LiDAR point clouds, which further saves the annotation budget. From \cref{figure:scribblekitti} we can observe that the range view methods are performing much better than the voxel-based methods~\cite{2019Minkowski,tang2020searching,zhu2021cylindrical} under weak supervisions. This is credited to the compact and semantic-abundant properties of the range view, which maintains better representations for learning. Without extra modules or procedures, \textit{RangeFormer} achieves $63.0\%$ mIoU and exhibits clear advantages for both the \textit{things} and \textit{stuff} classes.

\noindent\textbf{Accuracy \textit{vs.} Efficiency}. The trade-offs between segmentation accuracy and inference run-time are crucial for in-vehicle LiDAR segmentation. \cref{table:latency} summarizes the latency and mIoU scores of recent methods. We observe that the projection-based methods \cite{zhang2020polarnet,zhao2021fidnet,cheng2022cenet} tend to be much faster than the voxel- and fusion-based methods \cite{qiu2022GFNet,xu2021rpvnet,zhu2021cylindrical}, thanks to the dense and computation-friendly 2D representations.
Among all methods, \textit{RangeFormer} yields the best-possible trade-offs; it achieves much higher mIoU scores than prior range view methods~\cite{zhao2021fidnet,cheng2022cenet} while being $2\times$ to $5\times$ faster than the voxel and fusion counterparts \cite{2022_2DPASS,tang2020searching,xu2021rpvnet}. Furthermore, the range view methods also benefit from using pre-trained models on image datasets, \textit{e.g.} ImageNet \cite{deng2009imagenet} and Cityscapes \cite{Cityscapes}, as tested in \cref{tab:pre-train}.

\begin{figure*}[t]
    \begin{center}
    \includegraphics[width=1.0\linewidth]{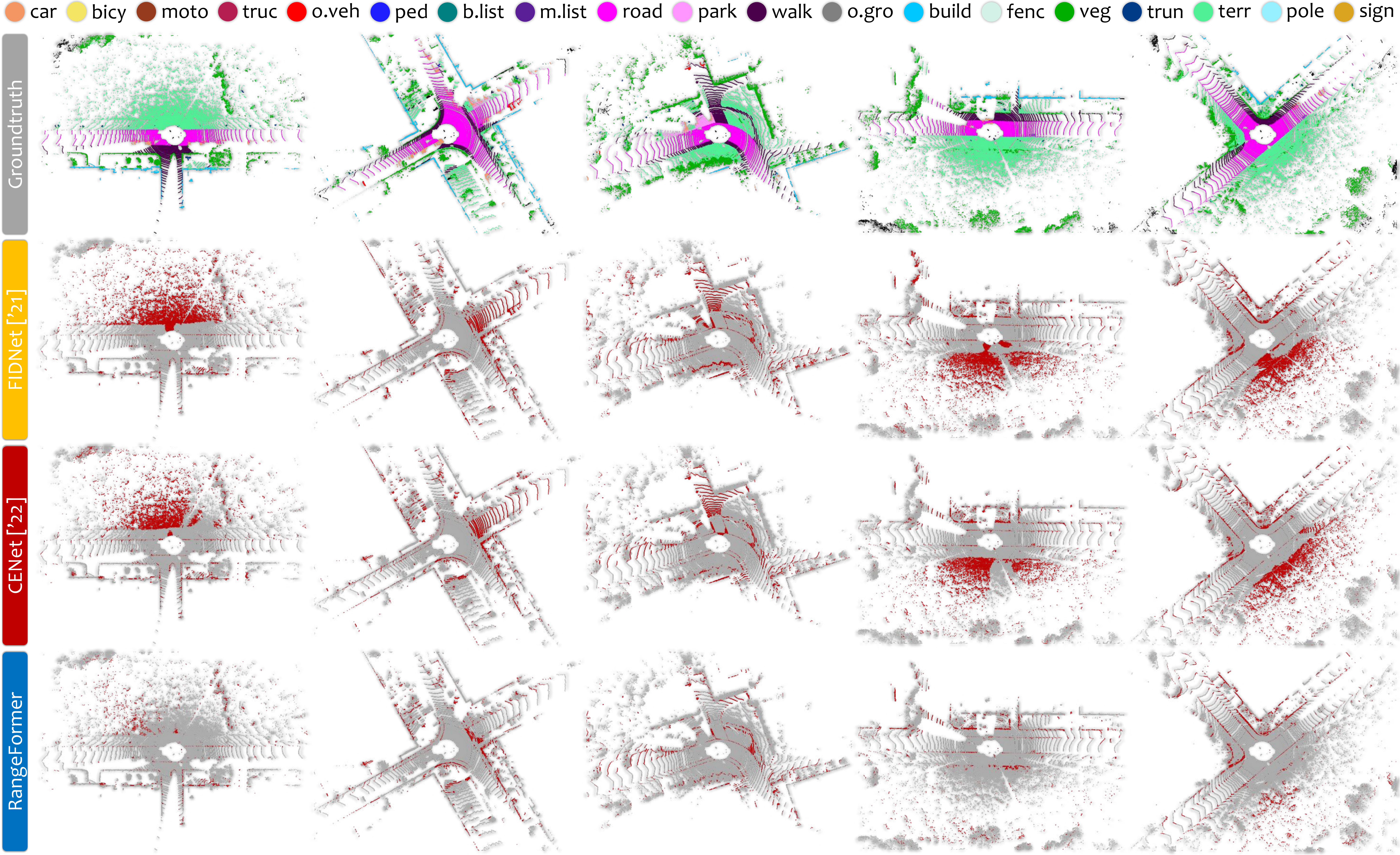}
    \end{center}
    \vspace{-0.44cm}
    \caption{Qualitative comparisons of state-of-the-art range view LiDAR segmentation methods~\cite{zhao2021fidnet,cheng2022cenet}. To highlight the differences, the \textbf{\textcolor{correct}{correct}} / \textbf{\textcolor{incorrect}{incorrect}} predictions are painted in \textbf{\textcolor{correct}{gray}} / \textbf{\textcolor{incorrect}{red}}, respectively. Each point cloud scene covers a region of size $50$m by $50$m, centered around the ego-vehicle. Best viewed in colors.}
    \label{figure:qualitative}
    \vspace{0.05cm}
\end{figure*}

\noindent\textbf{Qualitative Assessment}.
\cref{figure:qualitative} provides some visualization examples of SoTA range view LiDAR segmentation methods~\cite{zhao2021fidnet,cheng2022cenet} on sequence $08$ of SemanticKITTI \cite{SemanticKITTI}. As clearly shown from the error maps, prior arts find segmenting sparsely distributed regions difficult, \textit{e.g.}, \textit{terrain} and \textit{sidewalk}. In contrast, \textit{RangeFormer} -- which has the ability to model long-range dependencies and maintain large receptive fields -- is able to mitigate the errors holistically. We also find advantages in segmenting object shapes and boundaries. More visual comparisons are in Appendix.

\subsection{Ablation Study}
\label{sec:ablation}
Following~\cite{cheng2022cenet,xu2020squeezesegv3}, we probe each component in \textit{RangeFormer} with inputs of size $64\times 512$ on the \textit{val} set of SemanticKITTI~\cite{SemanticKITTI}. Since our contributions are generic, we also report results on SoTA range view methods~\cite{zhao2021fidnet,cheng2022cenet}.

\begin{table}[t]
\caption{The \textbf{trade-off} comparisons between \textbf{efficiency} (run-time) and \textbf{accuracy} (mIoU). Symbol $\clubsuit$: results on SemanticKITTI~\cite{SemanticKITTI} \textit{test} set. Symbol $\bigstar$~/~$\blacksquare$: results on nuScenes~\cite{Panoptic-nuScenes} \textit{val} / \textit{test} set. Latency is calculated on SemanticKITTI \cite{SemanticKITTI} and given in \textit{ms}. Symbol $^{\dagger}$: $W_{\text{train}}=384$ (SemanticKITTI) and $480$ (nuScenes), respectively.}
\vspace{-0.2cm}
\centering\scalebox{0.66}{
\begin{tabular}{r|cccccc}
\toprule
\textbf{Method~\small{(year)}} & Size & Latency & $\clubsuit$ & $\bigstar$ & $\blacksquare$ & Modality
\\\midrule
RangeNet++~\cite{milioto2019rangenet++}['19] & $50.4$M & 
$126$ & $52.2$ & $65.5$ & $-$ & Range Image
\\
KPConv~\cite{thomas2019kpconv}['19] & $18.3$M & $279$ & $58.8$ & $-$ & $-$ & Bag-of-Points
\\
MinkNet~\cite{2019Minkowski}['19] & $21.7$M & $294$ & $63.1$ & $-$ & $-$ & Sparse Voxel
\\
SalsaNext~\cite{cortinhal2020salsanext}['20] & $6.7$M & $71$ & $59.5$ & $72.2$ & $-$ & Range Image
\\
RandLA-Net~\cite{hu2019randla}['20] & $1.2$M & $880$ & $53.9$ & $-$ & $-$ & Bag-of-Points
\\
PolarNet~\cite{zhang2020polarnet}['20] & $13.6$M & $62$ & $57.2$ & $71.0$ & $69.4$ & Polar Image
\\
AMVNet~\cite{2020AMVNet}['20] & $-$ & $-$ & $65.3$ & $76.1$ & $77.3$ & Multiple
\\
SPVNAS~\cite{tang2020searching}['20] & $12.5$M & $259$ & $66.4$ & $-$ & $77.4$ & Sparse Voxel
\\
Cylinder3D~\cite{zhu2021cylindrical}['21] & $56.3$M & $170$ & $67.8$ & $76.1$ & $77.2$ & Sparse Voxel
\\
FIDNet~\cite{zhao2021fidnet}['21] & $6.1$M & \underline{$16$} & $58.6$ & $71.4$ & $72.8$ & Range Image
\\
AF2-S3Net~\cite{2021AF2S3Net}['21] & $-$ & $-$ & $69.7$ & $62.2$ & $-$ & Multiple
\\
RPVNet~\cite{xu2021rpvnet}['21] & $24.8$M & $111$ & $68.3$ & $77.6$ & $-$ & Multiple
\\
2DPASS~\cite{2022_2DPASS}['22] & $-$ & $62$ & \underline{$72.9$} & $-$ & \underline{$80.8$} & Multiple
\\
GFNet~\cite{qiu2022GFNet}['22] & $-$ & $100$ & $65.4$ & $76.8$ & $-$ & Multiple
\\
LidarMultiNet~\cite{lidarmultinet}['22] & $-$ & $-$ & $-$ & $\mathbf{82.0}$ & $\mathbf{81.4}$ & Multiple
\\
CENet~\cite{cheng2022cenet}['22] & $6.8$M & $\mathbf{14}$ & $64.7$ & $73.3$ & $74.7$ & Range Image
\\
RangeViT~\cite{2023RangeViT}['23] & $-$ & $-$ & $-$ & $-$ & $75.2$ & Range Image
\\
\cellcolor{LightCyan}\textbf{RangeFormer} & \cellcolor{LightCyan}$24.3$M & \cellcolor{LightCyan}$37$ & \cellcolor{LightCyan}$\mathbf{73.3}$ & \cellcolor{LightCyan}\underline{$78.1$} & \cellcolor{LightCyan}$80.1$ & \cellcolor{LightCyan}Range Image
\\\midrule
\cellcolor{red!4}\textbf{\textit{w/} STR}$^{\dagger}$ & \cellcolor{red!4}$24.3$M & \cellcolor{red!4}$32$ & \cellcolor{red!4}$72.2$ & \cellcolor{red!4}$77.1$ & \cellcolor{red!4}$78.7$ & \cellcolor{red!4}Range Image
\\\bottomrule
\end{tabular}}
\label{table:latency}
\vspace{0.1cm}
\end{table}

\begin{table}[t]
\caption{Effect of \textbf{pre-training} strategies on the \textit{val} sets of SemanticKITTI~\cite{SemanticKITTI} (left) and nuScenes~\cite{Panoptic-nuScenes} (right), with spatial sizes $64\times 2048$ and $32\times 1920$, respectively.}
\vspace{-0.2cm}
\centering\scalebox{0.679}{
\begin{tabular}{c|c|c|cc}
\toprule
\textbf{Method~\small{(year)}} & FIDNet~\cite{zhao2021fidnet}['21] & CENet~\cite{cheng2022cenet}['22] & \cellcolor{LightCyan}\textbf{RangeFormer}
\\\midrule
No Pre-Train & $60.4_{\textcolor{gray}{\mathbf{+0.0}}}$ / $71.4_{\textcolor{gray}{\mathbf{+0.0}}}$ & $63.4_{\textcolor{gray}{\mathbf{+0.0}}}$ / $73.3_{\textcolor{gray}{\mathbf{+0.0}}}$ & \cellcolor{LightCyan}$68.1_{\textcolor{gray}{\mathbf{+0.0}}}$ / $77.1_{\textcolor{gray}{\mathbf{+0.0}}}$
\\\midrule
ImageNet & $61.6_{\textcolor{Royal}{\mathbf{+1.2}}}$ / $72.1_{\textcolor{Royal}{\mathbf{+0.7}}}$ & $64.1_{\textcolor{Royal}{\mathbf{+0.7}}}$ / $73.9_{\textcolor{Royal}{\mathbf{+0.6}}}$ & \cellcolor{LightCyan}$68.9_{\textcolor{Royal}{\mathbf{+0.8}}}$ / $77.6_{\textcolor{Royal}{\mathbf{+0.5}}}$
\\
Cityscapes & $-$ / $-$ & $-$ / $-$ & \cellcolor{LightCyan}$69.6_{\textcolor{Royal}{\mathbf{+1.5}}}$ / $78.1_{\textcolor{Royal}{\mathbf{+1.0}}}$ 
\\\bottomrule
\end{tabular}}
\label{tab:pre-train}
\end{table}

\begin{figure}[t]
    \begin{center}
    \includegraphics[width=1.0\linewidth]{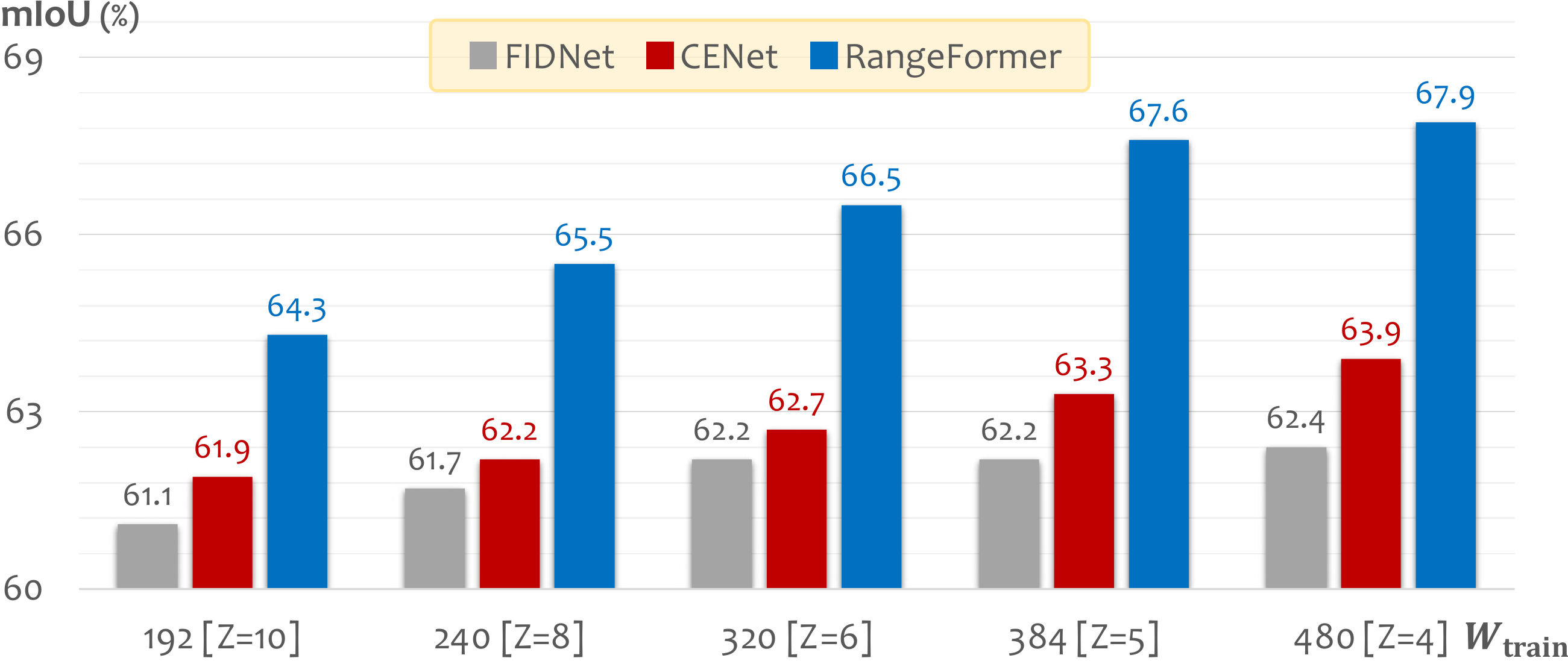}
    \end{center}
    \vspace{-0.4cm}
    \caption{Exploring best-possible ``view" partitions in STR.}
    \label{figure:str}
\end{figure}

\noindent\textbf{Augmentation}. As shown in \cref{figure:augmentation}, data augmentations help alleviate data scarcity and boost the segmentation performance by large margins. The attention-based models are known to be more dependent on data diversity~\cite{2021_ViT}. As a typical example, the ``plain" version of \textit{RangeFormer} yields a slightly lower score than CENet~\cite{cheng2022cenet}. On all three methods, \textit{RangeAug} helps to boost performance significantly and exhibits clear superiority over the common augmentations and the recent Mix3D~\cite{nekrasov2021mix3d}. It is worth mentioning that the extra overhead needed for \textit{RangeAug} is negligible on GPUs.

\noindent\textbf{Post-Processing}. \cref{figure:post-processing} attests again to the importance of post-processing in range view LiDAR segmentation. Without applying it, the ``many-to-one" problem will cause severe performance drops. Compared to the widely-adopted k-NN~\cite{milioto2019rangenet++} and the recent NLA~\cite{zhao2021fidnet}, \textit{RangePost} can better restore correct information since the aliasing among adjacent points has been reduced holistically. We also find the extra overhead negligible since the ``sub-clouds" are stacked along the batch dimension and can be processed in one forward pass. It is worth noting that such improvements happen after the training stage and are off-the-shelf and generic for various range view segmentation methods.

\noindent\textbf{Scalable Training}. To unveil the best-possible granularity in \textit{STR}, we split the point cloud into $4$, $5$, $6$, $8$, and $10$ views and show their results in \cref{figure:str}. We apply the \textit{same} training iteration for them hence their actual memory consumption becomes $\frac{1}{Z}$. We see that training on $4$ or $5$ views tends to yield better scores; while on more views the convergence rate will be affected, possibly by limited correlations in low-resolution range images. In summary, \textit{STR} opens up a new training paradigm for range view LiDAR segmentation which better balances the accuracy and efficiency.

\section{Conclusion}
\label{sec:conclusion}

In this work, in defense of the traditional range view representation, we proposed \textit{RangeFormer}, a novel framework that achieves superior performance than other modalities in both semantic and panoptic LiDAR segmentation. We also introduced \textit{STR}, a more scalable way of handling LiDAR point cloud learning and processing that yields better accuracy-efficiency trade-offs. Our approach has promoted more possibilities for accurate in-vehicle LiDAR perception. In the future, we seek more lightweight self-attention structures and computations to further increase efficiency.

\vspace{0.2cm}
\noindent\textbf{Acknowledgements}. This study is supported by the Ministry of Education, Singapore, under its MOE AcRF Tier 2 (MOE-T2EP20221-0012), NTU NAP, and under the RIE2020 Industry Alignment Fund – Industry Collaboration Projects (IAF-ICP) Funding Initiative, as well as cash and in-kind contribution from the industry partner(s).

\section*{Appendix}
In this appendix, we supplement more materials to support the main body of this paper. Specifically, this appendix is organized as follows.
\begin{itemize}
    \item \cref{sec:addition-implementation-details} elaborates on additional implementation details of the proposed methods and the experiments.
    \item \cref{sec:addition-quantitative-results} provides additional quantitative results, including the class-wise IoU scores for our comparative studies and ablation studies. 
    \item \cref{sec:addition-qualitative-results} attaches additional qualitative results, including more visual comparisons (figures) and demos (videos).
    \item \cref{sec:public-resources-used} acknowledges the public resources used during the course of this work.
\end{itemize}

\section{Additional Implementation Detail}
\label{sec:addition-implementation-details}
In this section, we provide additional technical details to help readers better understand our approach. Specifically, we first elaborate on the datasets and benchmarks used in our work. We then summarize the network configurations and provide more training and testing details.

\subsection{Benchmark}
\noindent\textbf{SemanticKITTI}. As an extension of the KITTI Vision
Odometry Benchmark, the SemanticKITTI~\cite{SemanticKITTI} dataset has been intensively used to evaluate and compare the model performance. It consists of $22$ sequences in total, collected from street scenes in Germany. The number of training, validation, and testing scans are $19130$, $4071$, and $20351$. The LiDAR point clouds are captured by the Velodyne HDL-64E sensor, resulting in around $120$k points per scan and a vertical angular resolution of $64$. Therefore, we set $H$ to $64$ during the 3D-to-2D rasterization. The conventional mapping of $19$ classes is adopted in this work.

\noindent\textbf{nuScenes}. As a multi-modal autonomous driving dataset, nuScenes~\cite{nuScenes} serves as the most comprehensive benchmark so far. It was developed by the team at Motional (formerly nuTonomy). The data was collected from Boston and Singapore. We use the \textit{lidarseg} set~\cite{Panoptic-nuScenes} in nuScenes for LiDAR segmentation. It contains $28130$ training scans and $6019$ validation scans. The Velodyne HDL32E sensor is used for data collection which yields sparser point clouds of around $40$k to $50$k points each. Therefore, we set $H$ to $32$ during the 3D-to-2D rasterization. We adopt the conventional $16$ classes from the official mapping in this work.

\noindent\textbf{ScribbleKITTI}. Since human labeling is often expensive and time-consuming, more and more recent works have started to seek weak annotations. ScribbleKITTI~\cite{2022ScribbleKITTI} re-labeled SemanticKITTI~\cite{SemanticKITTI} with line scribbles, resulting in a promising save of both time and effort. The final proportion of valid semantic labels over the point number is $8.06\%$. We adopt the same 3D-to-2D rasterization configuration as SemanticKITTI since these two sets share the same data format, \textit{i.e.}, $64$ beams, around $120$k points per LiDAR scan, and $16$ semantic classes. We follow the authors' original setting and report scores on sequence $08$ of SemanticKITTI.

\subsection{Model Configuration}
\noindent\textbf{Range Embedding Module (REM)}. After range view rasterization, the input $\mathcal{R}(u,v)$ of size $6\times H\times W$ is first fed into REM for range view point embedding. It consists of three MLP layers that map the embedding dim of $\mathcal{R}(u,v)$ from $6$ to $64$, $128$, and $128$, respectively, with the batch norm and GELU activation.

\noindent\textbf{Overlap Patch Embedding}. The output of size $128\times H\times W$ from REM serves as the input of the Transformer blocks. Specifically, for each of the four stages, the patch embedding layer divides an input of size $H_\text{embed}, W_\text{embed}$ into $3\times 3$ patches with overlap stride equals to $1$ (for the first stage) and $2$ (for the last three stages). 

\noindent\textbf{Multi-Head Attention \& Feed-Forward}. After the overlap patch embedding, the patches are processed with the standard multi-head attention operations as in \cite{2021_ViT,2021_PVT,2021segformer}. We keep the default setting of using the residual connection and layer normalization (Add \& Norm). The number of heads for each of the four stages is $[3, 4, 6, 3]$.

\noindent\textbf{Segmentation Head}. The hierarchical features extracted from different stages are stored and used for decoding. Specifically, each of the four stages produces features of spatial size $[(H, W), (\frac{H}{2}, \frac{W}{2}), (\frac{H}{4}, \frac{W}{4}), (\frac{H}{8}, \frac{W}{8})]$, with the channel dimension of $[128, 128, 320, 512]$. As described in the main body, we perform two unification steps to unify the channel and spatial sizes of different feature maps. We first map their channel dimensions to $256$, \textit{i.e.}, $[128, H, W]\rightarrow[256, H, W]$ for stage $1$, $[128, \frac{H}{2}, \frac{W}{2}]\rightarrow[256, \frac{H}{2}, \frac{W}{2}]$ for stage $2$, $[320, \frac{H}{4}, \frac{W}{4}]\rightarrow[256, \frac{H}{4}, \frac{W}{4}]$ for stage $3$, and $[512, \frac{H}{8}, \frac{W}{8}]\rightarrow[256, \frac{H}{8}, \frac{W}{8}]$ for stage $4$. We then interpolate four feature maps to the spatial size of $H\times W$. 

\subsection{Training \& Testing Configuration}
Our LiDAR segmentation model is implemented using PyTorch. The proposed data augmentations (\textit{RangeAug}), post-processing techniques (\textit{RangePost}), and \textit{STR} partition strategies are GPU-assisted and are within the data preparation process, which avoids adding extra overhead during model training. The configuration of the ``common" data augmentations, \textit{i.e.}, scaling, global rotation, jittering, flipping, and random dropping, is described as follows.

\begin{itemize}
    \item \textbf{Random Scaling:} A global transformation of point coordinates $(p^x, p^y, p^z)$, where for each point the coordinates are randomly scaled within a range of $-0.05\%$ to $0.05\%$.
    \item \textbf{Global Rotation:} A global transformation of point coordinates $(p^x, p^y)$ in the $XY$-plane, where the rotation angle is randomly selected within a range of $0$ degree to $360$ degree.
    \item \textbf{Random Jittering:} A global transformation of point coordinates $(p^x, p^y, p^z)$, where for each point the coordinates are randomly jittered within a range of $-0.3$m to $0.3$m.
    \item \textbf{Random Flipping:} A global transformation of point coordinates $(p^x, p^y)$ with three options, \textit{i.e.}, flipping along the $X$ axis only, flipping along the $Y$ axis only, and flipping along both the $X$ axis and $Y$ axis.
    \item \textbf{Random Dropping:} A global transformation that randomly removes a certain proportion $k_\text{drop}$ of points from the whole LiDAR point cloud before range view rasterization. In our experiments, $k_\text{drop}$ is set to $10\%$.
\end{itemize}

\begin{algorithm}[t]
\caption{CommonAug, NumPy-style}
\label{alg:common_aug}
\definecolor{codeblue}{rgb}{0.25,0.5,0.5}
\definecolor{codekw}{rgb}{0.85, 0.18, 0.50}
\lstset{
  backgroundcolor=\color{white},
  basicstyle=\fontsize{7.5pt}{7.5pt}\ttfamily\selectfont,
  columns=fullflexible,
  breaklines=true,
  captionpos=b,
  commentstyle=\fontsize{7.5pt}{7.5pt}\color{codeblue},
  keywordstyle=\fontsize{7.5pt}{7.5pt}\color{codekw},
}
\begin{lstlisting}[language=python]
# m: number of points in the lidar scan.
# c: number of channels for the lidar scan.
# scan: lidar scan, shape: [m, c].
# label: corresponding semantic label, shape: [m,].

# r_j: jittering rate, set as 0.3 in this work.
# r_s: scaling rate, set as 0.05 in this work.
# r_d: dropping rate, set as 0.1 in this work.


RAND = np.random.random()
NORM = np.random.normal()
UNIF = np.random.uniform()
RINT = np.random.randint()


def RandomScaling(scan, r_s):
    scale = UNIF(1, r_s)
    if RAND < 0.5:
        scale = 1 / scale
    scan[:, 0] *= scale
    scan[:, 1] *= scale
    
    return scan

def GlobalRotation(scan):
    rotate_rad = np.deg2rad(RAND() * 360)
    c, s = np.cos(rotate_rad), np.sin(rotate_rad)
    j = np.matrix([[c, s], [-s, c]])
    scan[:, :2] = np.dot(scan[:, :2], j)
    
    return scan

def RandomJittering(scan, r_j):
    jitter = np.clip(NORM(0, r_j, 3), -3*r_j, 3*r_j)
    scan += jitter
    
    return scan

def RandomFlipping(scan):
    flip_type = np.random.choice(4, 1)
    if flip_type == 1:
        scan[:, 0] = -scan[:, 0]
    elif flip_type == 2:
        scan[:, 1] = -scan[:, 1]
    elif flip_type == 3:
        scan[:, :2] = -scan[:, :2]
        
    return scan

def RandomDropping(scan, label):
    drop = int(len(scan) * r_d)
    drop = RINT(low=0, high=drop)
    to_drop = RINT(low=0, high=len(scan)-1, size=drop)
    to_drop = np.unique(to_drop)
    scan = np.delete(scan, to_drop, axis=0)
    label = np.delete(label, to_drop, axis=0)
    
    return scan, label

scan = RandomScaling(scan, r_s)
scan = GlobalRotation(scan)
scan = RandomJittering(scan, r_j)
scan = RandomFlipping(scan)
scan, label = RandomDropping(scan, label)

\end{lstlisting}
\end{algorithm}

\begin{algorithm}[t]
\caption{RangeAug, NumPy-style}
\label{alg:range_aug}
\definecolor{codeblue}{rgb}{0.25,0.5,0.5}
\definecolor{codekw}{rgb}{0.85, 0.18, 0.50}
\lstset{
  backgroundcolor=\color{white},
  basicstyle=\fontsize{7.5pt}{7.5pt}\ttfamily\selectfont,
  columns=fullflexible,
  breaklines=true,
  captionpos=b,
  commentstyle=\fontsize{7.5pt}{7.5pt}\color{codeblue},
  keywordstyle=\fontsize{7.5pt}{7.5pt}\color{codekw},
}
\begin{lstlisting}[language=python]
# m: number of points in the lidar scan.
# c: number of channels for the lidar scan.
# scan: lidar scan, shape: [m, c].
# label: corresponding semantic label, shape: [m,].

# h: height of the range view scan.
# w: width of the range view scan.
# cc: number of channels for the range view scan.
# rv: rasterized scan, shape: [cc, h, w].
# rvl: rasterized label, shape: [h, w].

# lidar_list: list of file names.
# sample: function to get a scan-label pair (idx).
# rasterize: range view rasterization function, refer to Eq. (1) in the main body.

# mix_strategies: select a strategy in RangeMix.
# sem_classes: semantic class list for RangePaste.


idx_a = np.random.randint(0, len(lidar_list))
idx_b = np.random.randint(0, len(lidar_list))


def RangeMix(xa, ya, xb, yb, mix_strategies):
    xa_, ya_ = xa.copy(), ya.copy()
    phi, theta = mix_strategies
    mix_h, mix_w = int(h / phi), int(w / theta)
    for i in range(1, mix_h):
        for j in range(1, mix_w):
            xa_[:, i-1:i, j-1:j] = xb[:, i-1:i, j-1:j]
            ya_[:, i-1:i, j-1:j] = yb[:, i-1:i, j-1:j]
    
    return xa_, y_a
    
def RangeUnion(xa, ya, xb, yb):
    xa_, ya_ = xa.copy(), ya.copy()
    mask = xa[-1, :, :]  # existence (0 or 1)
    void = mask == 0
    xa_[void], ya_[void] = xb[void], yb[void]
    
    return xa_, y_a

def RangePaste(xa, ya, xb, yb, sem_classes):
    xa_, ya_ = xa.copy(), ya.copy()
    for sem_class in sem_classes:
        pix = yb == sem_class
        xa_[pix], ya_[pix] = xb[pix], yb[pix]
    
    return xa_, y_a

def RangeShift(xa, ya):
    xa_, ya_ = xa.copy(), ya.copy()
    p = random.randint(int(0.25*w), int(0.75*w))
    xa_ = np.concatenate(xa[:, p:, :], xa[:, :p, :], axis=1)
    ya_ = np.concatenate(ya[p:, :], ya[:p, :], axis=1)
    
    return xa_, y_a


# Step 1: Sample two LiDAR scans 
scan_a, label_a = sample(idx_a)  # current sample
scan_b, label_b = sample(idx_b)  # another sample

# Step 2: 3D to 2D rasterization
rv_a, rvl_a = rasterize(scan_a, label_a)
rv_b, rvl_b = rasterize(scan_b, label_b)

# Step 3: RangeAug augmentation
rv_a, rvl_a = RangeMix(rv_a, rvl_a, rv_b, rvl_b, mix_strategies)
rv_a, rvl_a = RangePaste(rv_a, rvl_a, rv_b, rvl_b, sem_classes)
rv_a, rvl_a = RangeUnion(rv_a, rvl_a, rv_b, rvl_b)
rv_a, rvl_a = RangeShift(rv_a, rvl_a, rv_b, rvl_b)

\end{lstlisting}
\end{algorithm}

Additionally, the configuration of the proposed range view augmentation combo is described as follows.
\begin{itemize}
    \item \textbf{RangeMix:} After calculating all the inclination angles $\phi$ and azimuth angles $\theta$ for the current scan and the randomly sampled scan, we then split points into $k_\text{mix}$ equal spanning inclination ranges, \textit{i.e.}, different mixing strategies. The corresponding points in the same inclination range from the two scans are then switched. In our experiments, we design mixing strategies from a combination, and $k_\text{mix}$ is randomly sampled from a list $[2, 3, 4, 5, 6]$.
    \item \textbf{RangeUnion:} The existence $p^e$ in the point embedding is used to create a potential mask, which is then used as the indicator for supplementing the empty grids in the current range image with points (in the corresponding position) from a randomly sampled scan. Given a number of $N_\text{union}=\sum_n p^e_n$ empty range view grids, we randomly select $k_\text{union}N_\text{union}$ candidate grids for point filling, where $k_\text{union}$ is set as $50\%$.
    \item \textbf{RangePaste:} The groundtruth semantic labels of a randomly sampled scan are used to create pasting masks. The classes to be pasted are those in the ``tail" distribution, which forms a semantic class list (\texttt{sem classes}). After indexing the rare classes' points, we paste them into the current scan while maintaining the corresponding positions in the range image.
    \item \textbf{RangeShift:} A global transformation of points in the range view grids with respect to their azimuth angles $\theta$. This corresponds to shifting the range view grids along the row direction with $k_\text{shift}$ rows. In our experiments, $k_\text{shift}$ is randomly sampled from a range of $\frac{W}{4}$ to $\frac{3W}{4}$.
\end{itemize}

During \textit{\textbf{training}}, the probabilities of conducting the five common augmentations are set as $[1.0, 1.0, 1.0, 1.0, 0.9]$; while the probabilities of conducting our range view augmentations are set as $[0.9, 0.2, 0.9, 1.0]$. 

During \textit{\textbf{validation}}, all the data augmentations, \textit{i.e.}, both the common augmentation operations and the proposed range view augmentation operations, are set to false. We notice that some recent works use tricks on the validation set, such as test-time augmentation, model ensemble, \textit{etc}. It is worth mentioning that, we do not use any tricks to boost the validation performance so that the results are directly comparable with methods that follow the standard setting.

During \textit{\textbf{testing}}, we follow the conventional setting in CENet \cite{cheng2022cenet} and apply the test-time augmentation during the prediction stage. We use the code from the CENet authors for implementing this: it votes among multiple augmented inputs for generating the final predictions. Three common augmentations, \textit{i.e.}, global rotation, random jittering, and random flipping, are used to produce augmented inputs. The number of votes is set to $11$. We do not use model ensemble to boost the testing performance. Following the convention, we report the augmented results on the \textit{test} sets of the SemanticKITTI and nuScenes benchmarks. For ScribbleKITTI~\cite{2022ScribbleKITTI}, we reproduced FIDNet \cite{zhao2021fidnet}, CENet \cite{cheng2022cenet}, SPVCNN \cite{tang2020searching}, and Cylinder3D \cite{zhu2021cylindrical} and report their scores on the standard ScribbleKITTI \textit{val} set, without using test-time augmentation or model ensemble.

\subsection{STR: Scalable Training Strategy}
\begin{figure}[t]
    \begin{center}
    \includegraphics[width=1.0\linewidth]{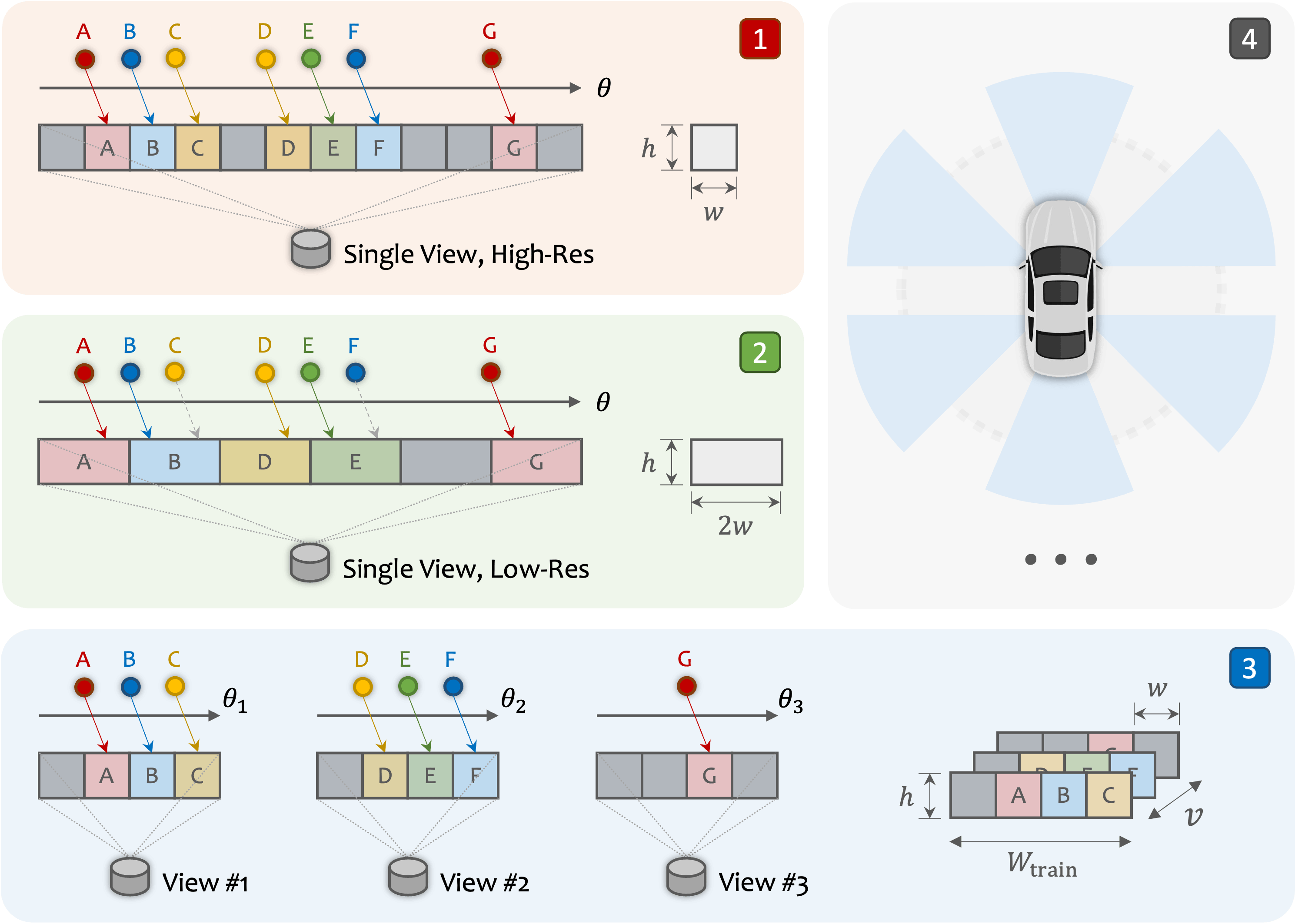}
    \end{center}
    \vspace{-0.4cm}
    \caption{Illustrative examples of the proposed \textbf{\textit{STR} strategy}. \textbf{1)}~Using a high-resolution range image during range view rasterization can better maintain semantic details, but tends to suffer from high computational cost. \textbf{2)} Using a low-resolution range image during range view rasterization saves computation budget, but tends to suffer from the ``many-to-one" issue. \textbf{3)} We propose to first split the whole LiDAR point cloud into multiple ``views" and then rasterize them into range images with high horizontal angular resolutions. Here we show a three-view example. \textbf{4)} In the actual implementation, we split the whole scene into several ``views" where each of them covers a specific region, centered around the ego-vehicle.}
    \label{figure-supp:scalable}
    \vspace{0.1cm}
\end{figure}

As stated in the main body, we propose a Scalable Training from Range view (STR) strategy to save computational costs during training. As shown in \cref{figure-supp:scalable}, STR allows us to train the range view models on arbitrary low-resolution 2D range images, while still maintaining satisfactory 3D segmentation accuracy. It offers a better trade-off between accuracy and efficiency, which are the two most important factors for in-vehicle LiDAR segmentation.

\subsection{Post-Processing Configuration}
As stated in the main body, we propose a novel RangePost technique to better handle the ``many-to-one" conflict in the range view rasterization. \cref{alg:range_post} shows the pseudo-code of the RangeAug operation. Specifically, we first sub-sample the whole LiDAR point cloud into equal-interval “subclouds”, which share similar semantics. Next, we stack and feed these subsets of the point cloud to the LiDAR segmentation model for inference. After obtaining the predictions, we then stitch them back to their original positions. As has been verified on several range view methods in our experiments, RangePost can better restore correct information since the aliasing among adjacent points has been reduced in a holistic manner.

\begin{algorithm}[t]
\caption{RangePost, NumPy/PyTorch-style}
\label{alg:range_post}
\definecolor{codeblue}{rgb}{0.25,0.5,0.5}
\definecolor{codekw}{rgb}{0.85, 0.18, 0.50}
\lstset{
  backgroundcolor=\color{white},
  basicstyle=\fontsize{7.5pt}{7.5pt}\ttfamily\selectfont,
  columns=fullflexible,
  breaklines=true,
  captionpos=b,
  commentstyle=\fontsize{7.5pt}{7.5pt}\color{codeblue},
  keywordstyle=\fontsize{7.5pt}{7.5pt}\color{codekw},
}
\begin{lstlisting}[language=python]
# m: number of points in the lidar scan.
# c: number of channels for the lidar scan.
# scan: lidar scan, shape: [m, c].

# h: height of the range view scan.
# w: width of the range view scan.
# cc: number of channels for the range view scan.
# rv: rasterized scan, shape: [cc, h, w].

# rasterize: range view rasterization function, refer to Eq. (1) in the main body.
# stack: function that joins a sequence of arrays along a new axis.
# unstack: function that split an array into a sequence of arrays.

# model: the LiDAR segmentation model.
# num_sub: number of sub-clouds to split for a scan.
# knn: the conventional clustering algorithm.
# file_path: path for saving the prediction file.


def SubCloud(scan, num_sub):
    subclouds = []
    for i in range(num_sub):
        scan_i = scan[i::num_sub, :]
        rv_i = rasterize(scan_i)
        subclouds.append(rv_i)
    
    return subclouds


# Step 1: Split whole cloud into sub-clouds
subclouds = SubCloud(scan, num_sub)

# Step 2: Stack sub-clouds along batch axis
rv_stack = stack(subclouds)

# Step 3: Model inference
with torch.no_grad():
    rv_preds = model(rv_stack)

# Step 4: Put sub-cloud back to whole cloud
rv_unstack = unstack(rv_preds)

# Step 5: Conventional post-processing & saving
pred = np.zeros(scan.size(), dtype=np.float32)
for j in range(len(rv_unstack)):
    pred_j = knn(rv_unstack[j])
    pred[j::num_sub] = pred_j

pred = pred.astype(np.uint32)
pred.tofile(file_path)

\end{lstlisting}
\end{algorithm}

\section{Additional Quantitative Result}
\label{sec:addition-quantitative-results}
In this section, we provide additional quantitative results of our comparative and ablation studies on the three tested LiDAR segmentation datasets.

\subsection{Comparative Study}
We conduct extensive experiments on three popular LiDAR segmentation benchmarks, \textit{i.e.}, SemanticKITTI~\cite{SemanticKITTI}, nuScenes~\cite{Panoptic-nuScenes}, and ScribbleKITTI~\cite{2022ScribbleKITTI}. \cref{table:semantickitti-class} shows the class-wise IoU scores of different LiDAR semantic segmentation methods on the \textit{test} set of SemanticKITTI~\cite{SemanticKITTI}. Among all competitors, we observe clear advantages of \textit{RangeFormer} and its \textit{STR} version over the raw point, bird's eye view, range view, and voxel methods. We also achieve better scores than the recent multi-view fusion-based methods~\cite{2022_2DPASS,pvkd2022,2022GASN,li2022SDSeg3D}, while using only the range view representation. \cref{table:scribblekitti-class} shows the class-wise IoU scores of different LiDAR semantic segmentation methods on the \textit{val} set of ScribbleKITTI~\cite{2022ScribbleKITTI} (the same as the \textit{val} set of SemanticKITTI~\cite{SemanticKITTI}). We can see that \textit{RangeFormer} yields much higher IoU scores than the SoTA voxel and range view methods on this weakly-annotated dataset. The superiority is especially overt for the dynamic classes, such as \textit{car}, \textit{bicycle}, \textit{motorcycle}, and \textit{person}. It is also worth noting that our approach has achieved better scores than some fully-supervised methods (\cref{table:semantickitti-class}) while using only $8.06\%$ semantic labels. \cref{table:nuscenes-class} and \cref{table:nuscenes-class-test} show the class-wise IoU scores of different LiDAR semantic segmentation methods on the \textit{val} set and \textit{test} set of nuScenes~\cite{Panoptic-nuScenes}, respectively. The results demonstrate again the advantages of both \textit{RangeFormer} and \textit{STR} for LiDAR semantic segmentation. We achieve new SoTA results on the three benchmarks which cover various cases, \textit{i.e.}, dense/sparse LiDAR point clouds, and full/weak supervision signals. Additionally, \cref{table:panoptic-class} shows the 
class-wise scores in terms of PQ, RQ, SQ, and IoU in the LiDAR panoptic segmentation benchmark of SemanticKITTI~\cite{SemanticKITTI}). For all four metrics, we observe advantages of both \textit{Panoptic-RangeFormer} and \textit{STR} compared to the recent SoTA LiDAR panoptic segmentation method~\cite{Panoptic-PHNet}.

\subsection{Ablation Study}
\cref{table:str-class} shows the class-wise IoU scores of FIDNet~\cite{zhao2021fidnet}, CENet~\cite{cheng2022cenet}, and \textit{RangeFormer}, under the \textit{STR} training strategies. We can see that the range view LiDAR segmentation methods are capable of training on very small resolution range images, \textit{e.g.}, $W=192$, $W=240$, and $W=320$. While saving a significant amount of memory consumption, the segmentation performance is relatively stable. For example, \textit{RangeFormer} can achieve $64.3\%$ mIoU with $W=192$, which is better than a wide range of prior LiDAR segmentation methods. The segmentation performance tends to be improved with higher horizontal resolutions. The flexibility of balancing accuracy and efficiency opens more possibilities and options for practitioners. 

\section{Additional Qualitative Result}
\label{sec:addition-qualitative-results}
In this section, we provide additional qualitative results of our approach to further demonstrate our superiority.

\subsection{Visual Comparison}
\cref{figure:qualitative_supp_01} and \cref{figure:qualitative_supp_02} include more visualization results of \textit{RangeFormer} and the SoTA range view LiDAR segmentation methods~\cite{zhao2021fidnet,cheng2022cenet}. Compared to the prior arts, we can see that \textit{RangeFormer} has yielded much better LiDAR segmentation performance. It holistically eliminates the erroneous predictions around the ego vehicle, especially for the complex regions where multiple classes are clustered together.

\subsection{Failure Case}
\begin{figure*}[t]
    \begin{center}
    \includegraphics[width=1.0\linewidth]{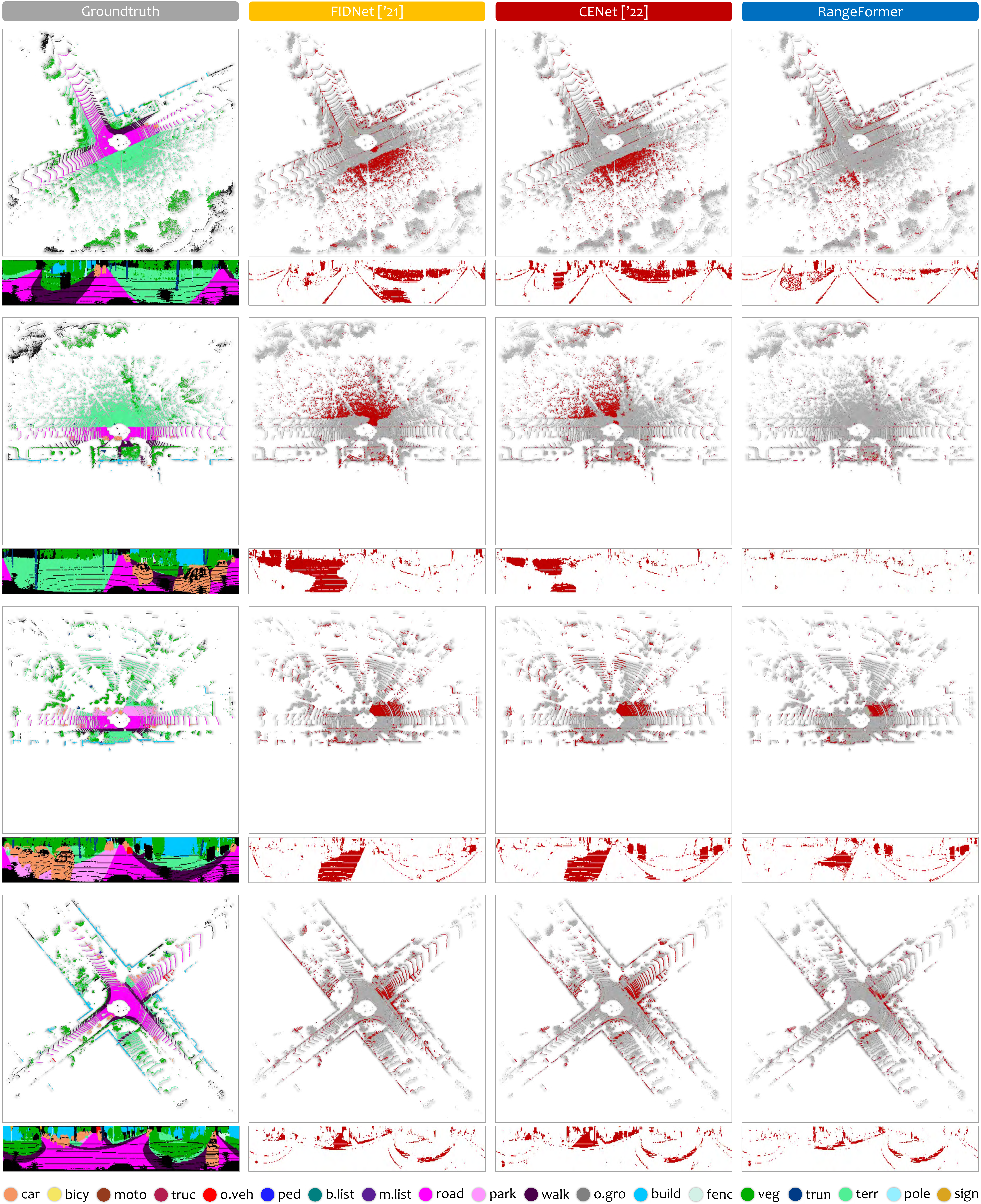}
    \end{center}
    \vspace{-0.5cm}
    \caption{\textbf{Additional qualitative comparisons (error maps)} with SoTA range view segmentation methods~\cite{zhao2021fidnet,cheng2022cenet}. To highlight the differences, the \textbf{\textcolor{correct}{correct}} / \textbf{\textcolor{incorrect}{incorrect}} predictions are painted in \textbf{\textcolor{correct}{gray}} / \textbf{\textcolor{incorrect}{red}}, respectively. Each scene is visualized from the LiDAR bird's eye view (top) and range view (bottom) and covers a region of size $50$m by $50$m, centered around the ego-vehicle. Best viewed in colors.}
    \label{figure:qualitative_supp_01}
\end{figure*}

\begin{figure*}[t]
    \begin{center}
    \includegraphics[width=1.0\linewidth]{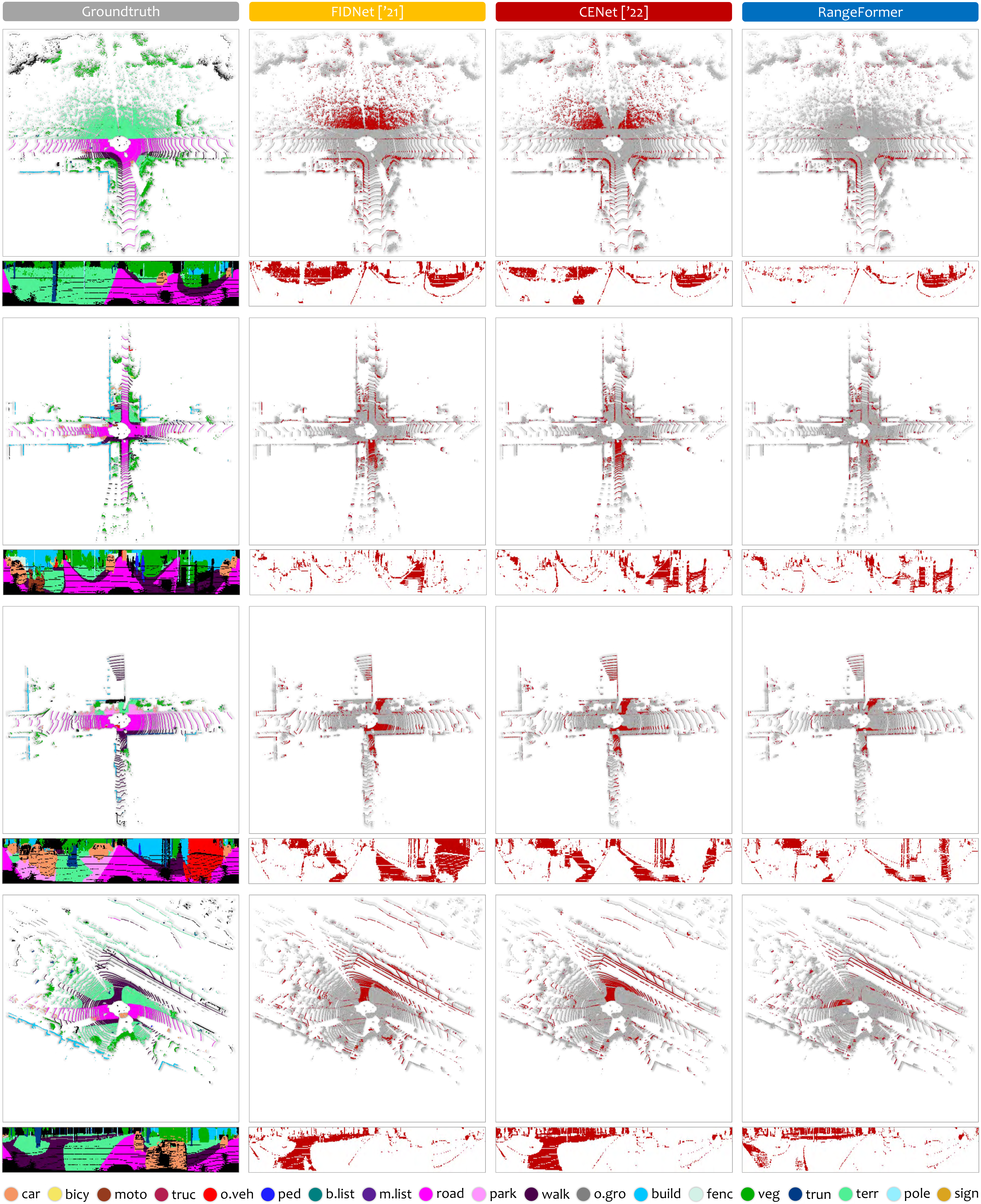}
    \end{center}
    \vspace{-0.5cm}
    \caption{\textbf{Additional qualitative comparisons (error maps)} with SoTA range view LiDAR segmentation methods~\cite{zhao2021fidnet,cheng2022cenet}. To highlight the differences, the \textbf{\textcolor{correct}{correct}} / \textbf{\textcolor{incorrect}{incorrect}} predictions are painted in \textbf{\textcolor{correct}{gray}} / \textbf{\textcolor{incorrect}{red}}, respectively. Each scene is visualized from the LiDAR bird's eye view (top) and range view (bottom) and covers a region of size $50$m by $50$m, centered around the ego-vehicle. Best viewed in colors.}
    \label{figure:qualitative_supp_02}
\end{figure*}

Although \textit{RangeFormer} improves the LiDAR segmentation performance by large margins, there are still some failure cases that tend to appear. We can see from the error maps in \cref{figure:qualitative_supp_01} and \cref{figure:qualitative_supp_02} that the erroneous predictions are likely to occur at the boundary of the objects and backgrounds (the first scene in \cref{figure:qualitative_supp_01}). There are also possible false predictions for rare classes (the second scene in \cref{figure:qualitative_supp_02}) and for long-distance regions (the fourth scene in \cref{figure:qualitative_supp_02}). A more sophisticated design with considerations of such cases will likely yield better LiDAR segmentation performance.

\subsection{Video Demo}
In addition to the figures, we have attached four video demos in the supplementary materials, \textit{i.e.}, \texttt{demo1.mp4}, \texttt{demo2.mp4}, \texttt{demo3.mp4}, and \texttt{demo4.mp4}. Each video demo consists of hundreds of frames that provide a more comprehensive evaluation of our proposed approach. These video demos will be publicly available on our website\footnote{\url{https://ldkong.com/RangeFormer}.}.

\section{Public Resources Used}
\label{sec:public-resources-used}

We acknowledge the use of the following public resources, during the course of this work:

\begin{itemize}
    \item SemanticKITTI\footnote{\url{http://semantic-kitti.org}.} \dotfill CC BY-NC-SA 4.0
    \item SemanticKITTI-API\footnote{\url{https://github.com/PRBonn/semantic-kitti-api}.} \dotfill MIT License
    \item nuScenes\footnote{\url{https://www.nuscenes.org/nuscenes}.} \dotfill CC BY-NC-SA 4.0
    \item nuScenes-devkit\footnote{\url{https://github.com/nutonomy/nuscenes-devkit}.} \dotfill Apache License 2.0
    \item ScribbleKITTI\footnote{\url{https://github.com/ouenal/scribblekitti}.} \dotfill Unknown
    \item RangeNet++\footnote{\url{https://github.com/PRBonn/lidar-bonnetal}.} \dotfill MIT License
    \item SqueezeSegV3\footnote{\url{https://github.com/chenfengxu714/SqueezeSegV3}.} \dotfill BSD 2-Clause License
    \item SalsaNext\footnote{\url{https://github.com/TiagoCortinhal/SalsaNext}.} \dotfill MIT License
    \item FIDNet\footnote{\url{https://github.com/placeforyiming/IROS21-FIDNet-SemanticKITTI}.} \dotfill Unknown
    \item CENet\footnote{\url{https://github.com/huixiancheng/CENet}.} \dotfill MIT License
    \item PVT\footnote{\url{https://github.com/whai362/PVT}.} \dotfill Apache License 2.0
    \item SegFormer\footnote{\url{https://github.com/NVlabs/SegFormer}.} \dotfill NVIDIA Source Code License
    \item Segmenter\footnote{\url{https://github.com/rstrudel/segmenter}.} \dotfill MIT License
    \item ViT-PyTorch\footnote{\url{https://github.com/lucidrains/vit-pytorch}.} \dotfill MIT License
    \item DS-Net\footnote{\url{https://github.com/hongfz16/DS-Net}.} \dotfill MIT License
    \item Panoptic-PolarNet\footnote{\url{https://github.com/edwardzhou130/Panoptic-PolarNet}.} \dotfill BSD 3-Clause License
\end{itemize}

\clearpage

\begin{table*}[t]
\caption{\textbf{The class-wise IoU scores} of different LiDAR semantic segmentation approaches (raw point, bird's eye view, range view, voxel, and multi-view fusion) on the \textbf{SemanticKITTI}~\cite{SemanticKITTI} leaderboard. All IoU scores are given in percentage (\%). For each class: \textbf{bold} - best in column; \underline{underline} - second best in column. Symbol $^{\dagger}$: $W_{\text{train}}=384$. Methods are arranged in \textit{ascending} order of mIoU.}
\vspace{-0.1cm}
\centering\scalebox{0.662}{
\begin{tabular}{r|c|ccccccccccccccccccc}
\toprule
\textbf{Method~\small{(year)}} & \rotatebox{0}{$\text{mIoU}$} & \rotatebox{90}{car} & \rotatebox{90}{bicycle} & \rotatebox{90}{motorcycle} & \rotatebox{90}{truck} & \rotatebox{90}{other-vehicle} & \rotatebox{90}{person} & \rotatebox{90}{bicyclist} & \rotatebox{90}{motorcyclist} & \rotatebox{90}{road} & \rotatebox{90}{parking} & \rotatebox{90}{sidewalk} & \rotatebox{90}{other-ground} & \rotatebox{90}{building} & \rotatebox{90}{fence} & \rotatebox{90}{vegetation} & \rotatebox{90}{trunk} & \rotatebox{90}{terrain} & \rotatebox{90}{pole} & \rotatebox{90}{traffic-sign}
\\\midrule
PointNet~\cite{qi2017pointnet}~\small{['17]} & $14.6$ & $46.3$ & $1.3$ & $0.3$ & $0.1$ & $0.8$ & $0.2$ & $0.2$ & $0.0$ & $61.6$ & $15.8$ & $35.7$ & $1.4$ & $41.4$ & $12.9$ & $31.0$ & $4.6$ & $17.6$ & $2.4$ & $3.7$
\\
PointNet++~\cite{qi2017pointnet++}~\small{['17]} & $20.1$ & $53.7$ & $1.9$ & $0.2$ & $0.9$ & $0.2$ & $0.9$ & $1.0$ & $0.0$ & $72.0$ & $18.7$ & $41.8$ & $5.6$ & $62.3$ & $16.9$ & $46.5$ & $13.8$ & $30.0$ & $6.0$ & $8.9$
\\
SqSeg~\cite{wu2018squeezeseg}~\small{['18]} & $30.8$ & $68.3$ & $18.1$ & $5.1$ & $4.1$ & $4.8$ & $16.5$ & $17.3$ & $1.2$ & $84.9$ & $28.4$ & $54.7$ & $4.6$ & $61.5$ & $29.2$ & $59.6$ & $25.5$ & $54.7$ & $11.2$ & $36.3$
\\
SqSegV2~\cite{wu2019squeezesegv2}~\small{['19]} & $39.6$ & $82.7$ & $21.0$ & $22.6$ & $14.5$ & $15.9$ & $20.2$ & $24.3$ & $2.9$ & $88.5$ & $42.4$ & $65.5$ & $18.7$ & $73.8$ & $41.0$ & $68.5$ & $36.9$ & $58.9$ & $12.9$ & $41.0$
\\
RandLA-Net~\cite{hu2019randla}~\small{['20]} & $50.3$ & $94.0$ & $19.8$ & $21.4$ & $42.7$ & $38.7$ & $47.5$ & $48.8$ & $4.6$ & $90.4$ & $56.9$ & $67.9$ & $15.5$ & $81.1$ & $49.7$ & $78.3$ & $60.3$ & $59.0$ & $44.2$ & $38.1$
\\
RangeNet++~\cite{milioto2019rangenet++}~\small{['19]} & $52.2$ & $91.4$ & $25.7$ & $34.4$ & $25.7$ & $23.0$ & $38.3$ & $38.8$ & $4.8$ & $91.8$ & $65.0$ & $75.2$ & $27.8$ & $87.4$ & $58.6$ & $80.5$ & $55.1$ & $64.6$ & $47.9$ & $55.9$
\\
PolarNet~\cite{zhang2020polarnet}~\small{['20]} & $54.3$ & $93.8$ & $40.3$ & $30.1$ & $22.9$ & $28.5$ & $43.2$ & $40.2$ & $5.6$ & $90.8$ & $61.7$ & $74.4$ & $21.7$ & $90.0$ & $61.3$ & $84.0$ & $65.5$ & $67.8$ & $51.8$ & $57.5$
\\
MPF~\cite{2021MPF}~\small{['21]} & $55.5$ & $93.4$ & $30.2$ & $38.3$ & $26.1$ & $28.5$ & $48.1$ & $46.1$ & $18.1$ & $90.6$ & $62.3$ & $74.5$ & $30.6$ & $88.5$ & $59.7$ & $83.5$ & $59.7$ & $69.2$ & $49.7$ & $58.1$
\\
3D-MiniNet~\cite{alonso20203d-mininet}~\small{['20]} & $55.8$ & $90.5$ & $42.3$ & $42.1$ & $28.5$ & $29.4$ & $47.8$ & $44.1$ & $14.5$ & $91.6$ & $64.2$ & $74.5$ & $25.4$ & $89.4$ & $60.8$ & $82.8$ & $60.8$ & $66.7$ & $48.0$ & $56.6$
\\
SqSegV3~\cite{xu2020squeezesegv3}~\small{['20]} & $55.9$ & $92.5$ & $38.7$ & $36.5$ & $29.6$ & $33.0$ & $45.6$ & $46.2$ & $20.1$ & $91.7$ & $63.4$ & $74.8$ & $26.4$ & $89.0$ & $59.4$ & $82.0$ & $58.7$ & $65.4$ & $49.6$ & $58.9$
\\
KPConv~\cite{thomas2019kpconv}~\small{['20]} & $58.8$ & $96.0$ & $32.0$ & $42.5$ & $33.4$ & $44.3$ & $61.5$ & $61.6$ & $11.8$ & $88.8$ & $61.3$ & $72.7$ & $31.6$ & $95.0$ & $64.2$ & $84.8$ & $69.2$ & $69.1$ & $56.4$ & $47.4$
\\
SalsaNext~\cite{cortinhal2020salsanext}~\small{['20]} & $59.5$ & $91.9$ & $48.3$ & $38.6$ & $38.9$ & $31.9$ & $60.2$ & $59.0$ & $19.4$ & $91.7$ & $63.7$ & $75.8$ & $29.1$ & $90.2$ & $64.2$ & $81.8$ & $63.6$ & $66.5$ & $54.3$ & $62.1$
\\
FIDNet~\cite{zhao2021fidnet}~\small{['21]} & $59.5$ & $93.9$ & $54.7$ & $48.9$ & $27.6$ & $23.9$ & $62.3$ & $59.8$ & $23.7$ & $90.6$ & $59.1$ & $75.8$ & $26.7$ & $88.9$ & $60.5$ & $84.5$ & $64.4$ & $69.0$ & $53.3$ & $62.8$
\\
FusionNet~\cite{zhang2020FusionNet}~\small{['20]} & $61.3$ & $95.3$ & $47.5$ & $37.7$ & $41.8$ & $34.5$ & $59.5$ & $56.8$ & $11.9$ & $91.8$ & $68.8$ & $77.1$ & $30.8$ & $92.5$ & $69.4$ & $84.5$ & $69.8$ & $68.5$ & $60.4$ & $66.5$
\\
PCSCNet~\cite{2023_PCSCNet}~\small{['22]} & $62.7$ & $95.7$ & $48.8$ & $46.2$ & $36.4$ & $40.6$ & $55.5$ & $68.4$ & $55.9$ & $89.1$ & $60.2$ & $72.4$ & $23.7$ & $89.3$ & $64.3$ & $84.2$ & $68.2$ & $68.1$ & $60.5$ & $63.9$
\\
KPRNet~\cite{kochanov2020kprnet}~\small{['21]} & $63.1$ & $95.5$ & $54.1$ & $47.9$ & $23.6$ & $42.6$ & $65.9$ & $65.0$ & $16.5$ & $93.2$ & $\mathbf{73.9}$ & $80.6$ & $30.2$ & $91.7$ & $68.4$ & $85.7$ & $69.8$ & $71.2$ & $58.7$ & $64.1$
\\
TornadoNet~\cite{2021Tornado-net}~\small{['21]} & $63.1$ & $94.2$ & $55.7$ & $48.1$ & $40.0$ & $38.2$ & $63.6$ & $60.1$ & $34.9$ & $89.7$ & $66.3$ & $74.5$ & $28.7$ & $91.3$ & $65.6$ & $85.6$ & $67.0$ & $71.5$ & $58.0$ & $65.9$
\\
LiteHDSeg~\cite{razani2021lite}~\small{['21]} & $63.8$ & $92.3$ & $40.0$ & $55.4$ & $37.7$ & $39.6$ & $59.2$ & $71.6$ & $54.3$ & $93.0$ & $68.2$ & $78.3$ & $29.3$ & $91.5$ & $65.0$ & $78.2$ & $65.8$ & $65.1$ & $59.5$ & $67.7$
\\
RangeViT~\cite{2023RangeViT}~\small{['23]} & $64.0$ & $95.4$ & $55.8$ & $43.5$ & $29.8$ & $42.1$ & $63.9$ & $58.2$ & $38.1$ & $93.1$ & $70.2$ & $80.0$ & $32.5$ & $92.0$ & $69.0$ & $85.3$ & $70.6$ & $71.2$ & $60.8$ & $64.7$
\\
CENet~\cite{cheng2022cenet}~\small{['22]} & $64.7$ & $91.9$ & $58.6$ & $50.3$ & $40.6$ & $42.3$ & $68.9$ & $65.9$ & $43.5$ & $90.3$ & $60.9$ & $75.1$ & $31.5$ & $91.0$ & $66.2$ & $84.5$ & $69.7$ & $70.0$ & $61.5$ & $67.6$
\\
SVASeg~\cite{2022_SVASeg}~\small{['22]} & $65.2$ & $96.7$ & $56.4$
 & $57.0$ & $49.1$ & $56.3$ & $70.6$ & $67.0$ & $15.4$ & $92.3$ & $65.9$ & $76.5$ & $23.6$ & $91.4$ & $66.1$ & $85.2$ & $72.9$ & $67.8$ & $63.9$ & $65.2$
\\
AMVNet~\cite{2020AMVNet}~\small{['20]} & $65.3$ & $96.2$ & $59.9$ & $54.2$ & $48.8$ & $45.7$ & $71.0$ & $65.7$ & $11.0$ & $90.1$ & $71.0$ & $75.8$ & $32.4$ & $92.4$ & $69.1$ & $85.6$ & $71.7$ & $69.6$ & $62.7$ & $67.2$
\\
GFNet~\cite{qiu2022GFNet}~\small{['22]} & $65.4$ & $96.0$ & $53.2$ & $48.3$ & $31.7$ & $47.3$ & $62.8$ & $57.3$ & $44.7$ & $\mathbf{93.6}$ & $72.5$ & $\mathbf{80.8}$ & $31.2$ & $\mathbf{94.0}$ & $\mathbf{73.9}$ & $85.2$ & $71.1$ & $69.3$ & $61.8$ & $68.0$
\\
JS3C-Net~\cite{yan2021sparse}~\small{['21]} & $66.0$ & $95.8$ & $59.3$ & $52.9$ & $54.3$ & $46.0$ & $69.5$ & $65.4$ & $39.9$ & $88.9$ & $61.9$ & $72.1$ & $31.9$ & $92.5$ & $70.8$ & $84.5$ & $69.8$ & $67.9$ & $60.7$ & $68.7$
\\
MaskRange~\cite{2022MaskRange}~\small{['22]} & $66.1$ & $94.2$ & $56.0$ & $55.7$ & $59.2$ & $52.4$ & $67.6$ & $64.8$ & $31.8$ & $91.7$ & $70.7$ & $77.1$ & $29.5$ & $90.6$ & $65.2$ & $84.6$ & $68.5$ & $69.2$ & $60.2$ & $66.6$ 
\\
SPVNAS~\cite{tang2020searching}~\small{['20]} & $66.4$ & $97.3$ & $51.5$ & $50.8$ & $59.8$ & $58.8$ & $65.7$ & $65.2$ & $43.7$ & $90.2$ & $67.6$ & $75.2$ & $16.9$ & $91.3$ & $65.9$ & $86.1$ & $73.4$ & $71.0$ & $64.2$ & $66.9$
\\
MSSNet~\cite{2022MSSNet}~\small{['21]} & $66.7$ & $96.8$ & $52.2$ & $48.5$ & $54.4$ & $56.3$ & $67.0$ & $70.9$ & $49.3$ & $90.1$ & $65.5$ & $74.9$ & $30.2$ & $90.5$ & $64.9$ & $84.9$ & $72.7$ & $69.2$ & $63.2$ & $65.1$
\\
Cylinder3D~\cite{zhu2021cylindrical}~\small{['21]} & $68.9$ & $97.1$ & $67.6$ & $63.8$ & $50.8$ & $58.5$ & $73.7$ & $69.2$ & $48.0$ & $92.2$ & $65.0$ & $77.0$ & $32.3$ & $90.7$ & $66.5$ & $85.6$ & $72.5$ & $69.8$ & $62.4$ & $66.2$
\\
AF2S3Net~\cite{2021AF2S3Net}~\small{['21]} & $69.7$ & $94.5$ & $65.4$ & $\mathbf{86.8}$ & $39.2$ & $41.1$ & $\mathbf{80.7}$ & $80.4$ & $\mathbf{74.3}$ & $91.3$ & $68.8$ & $72.5$ & $\mathbf{53.5}$ & $87.9$ & $63.2$ & $70.2$ & $68.5$ & $53.7$ & $61.5$ & \underline{$71.0$}
\\
RPVNet~\cite{xu2021rpvnet}~\small{['21]} & $70.3$  & $\mathbf{97.6}$ & \underline{$68.4$} & $68.7$ & $44.2$ & $61.1$ & $75.9$ & $74.4$ & $73.4$ & \underline{$93.4$} & $70.3$ & \underline{$80.7$} & $33.3$ & \underline{$93.5$} & $72.1$ & $86.5$ & $\mathbf{75.1}$ & $71.7$ & $64.8$ & $61.4$
\\
SDSeg3D~\cite{li2022SDSeg3D}~\small{['22]} & $70.4$ & \underline{$97.4$} & $58.7$ & $54.2$ & $54.9$ & $65.2$ & $70.2$ & $74.4$ & $52.2$ & $90.9$ & $69.4$ & $76.7$ & $41.9$ & $93.2$ & $71.1$ & $86.1$ & \underline{$74.3$} & $71.1$ & $65.4$ & $70.6$
\\
GASN~\cite{2022GASN}~\small{['22]} & $70.7$ & $96.9$ & $65.8$ & $58.0$ & $59.3$ & $61.0$ & \underline{$80.4$} & $\mathbf{82.7}$ & $46.3$ & $89.8$ & $66.2$ & $74.6$ & $30.1$ & $92.3$ & $69.6$ & $\mathbf{87.3}$ & $73.0$ & \underline{$72.5$} & \underline{$66.1$} & $\mathbf{71.6}$
\\
PVKD~\cite{pvkd2022}~\small{['22]} & $71.2$ & $97.0$ & $67.9$ & $69.3$ & $53.5$ & $60.2$ & $75.1$ & $73.5$ & $50.5$ & $91.8$ & $70.9$ & $77.5$ & $41.0$ & $92.4$ & $69.4$ & $86.5$ & $73.8$ & $71.9$ & $64.9$ & $65.8$
\\
\cellcolor{red!4}\textbf{STR$^{\dagger}$~(Ours)} & \cellcolor{red!4}$72.2$ & \cellcolor{red!4}$96.4$ & \cellcolor{red!4}$67.1$ & \cellcolor{red!4}$72.2$ & \cellcolor{red!4}$58.8$ & \cellcolor{red!4}$\mathbf{67.4}$ & \cellcolor{red!4}$74.9$ & \cellcolor{red!4}$74.7$ & \cellcolor{red!4}$57.5$ & \cellcolor{red!4}$92.1$ & \cellcolor{red!4}$72.5$ & \cellcolor{red!4}$78.2$ & \cellcolor{red!4}\underline{$42.4$} & \cellcolor{red!4}$91.8$ & \cellcolor{red!4}$69.7$ & \cellcolor{red!4}$85.8$ & \cellcolor{red!4}$70.4$ & \cellcolor{red!4}$72.3$ & \cellcolor{red!4}$62.8$ & \cellcolor{red!4}$65.0$
\\
2DPASS~\cite{2022_2DPASS}~\small{['22]} & \underline{$72.9$} & $97.0$ & $63.6$ & $63.4$ & $\mathbf{61.1}$ & $61.5$ & $77.9$ & \underline{$81.3$} & \underline{$74.1$} & $89.7$ & $67.4$ & $74.7$ & $40.0$ & \underline{$93.5$} & \underline{$72.9$} & $86.2$ & $73.9$ & $71.0$ & $65.0$ & $70.4$
\\
\cellcolor{LightCyan}\textbf{RangeFormer~(Ours)} & \cellcolor{LightCyan}$\mathbf{73.3}$ & \cellcolor{LightCyan}$96.7$ & \cellcolor{LightCyan}$\mathbf{69.4}$ & \cellcolor{LightCyan}\underline{$73.7$} & \cellcolor{LightCyan}\underline{$59.9$} & \cellcolor{LightCyan}\underline{$66.2$} & \cellcolor{LightCyan}$78.1$ & \cellcolor{LightCyan}$75.9$ & \cellcolor{LightCyan}$58.1$ & \cellcolor{LightCyan}$92.4$ & \cellcolor{LightCyan}\underline{$73.0$} & \cellcolor{LightCyan}$78.8$ & \cellcolor{LightCyan}\underline{$42.4$} & \cellcolor{LightCyan}$92.3$ & \cellcolor{LightCyan}$70.1$ & \cellcolor{LightCyan}\underline{$86.6$} & \cellcolor{LightCyan}$73.3$ & \cellcolor{LightCyan}$\mathbf{72.8}$ & \cellcolor{LightCyan}$\mathbf{66.4}$ & \cellcolor{LightCyan}$66.6$
\\\bottomrule
\end{tabular}}
\label{table:semantickitti-class}
\end{table*}

\begin{table*}[t]
\caption{\textbf{The class-wise IoU scores} of different LiDAR semantic segmentation approaches on the \textbf{ScribbleKITTI}~\cite{2022ScribbleKITTI} leaderboard. All IoU scores are given in percentage (\%). For each class: \textbf{bold} - best in column; \underline{underline} - second best in column. Methods are arranged in \textit{ascending} order of mIoU. Note that we have applied the proposed \textit{RangeAug} and \textit{RangePost} to FIDNet~\cite{zhao2021fidnet} and CENet~\cite{cheng2022cenet}, so the comparisons are only correlated to the model architecture.}
\vspace{-0.1cm}
\centering\scalebox{0.662}{
\begin{tabular}{r|c|ccccccccccccccccccc}
\toprule
\textbf{Method~\small{(year)}} & \rotatebox{0}{$\text{mIoU}$} & \rotatebox{90}{car} & \rotatebox{90}{bicycle} & \rotatebox{90}{motorcycle} & \rotatebox{90}{truck} & \rotatebox{90}{other-vehicle} & \rotatebox{90}{person} & \rotatebox{90}{bicyclist} & \rotatebox{90}{motorcyclist} & \rotatebox{90}{road} & \rotatebox{90}{parking} & \rotatebox{90}{sidewalk} & \rotatebox{90}{other-ground} & \rotatebox{90}{building} & \rotatebox{90}{fence} & \rotatebox{90}{vegetation} & \rotatebox{90}{trunk} & \rotatebox{90}{terrain} & \rotatebox{90}{pole} & \rotatebox{90}{traffic-sign}
\\\midrule
MinkNet~\cite{2019Minkowski}~\small{['19]} & $55.0$ & $88.1$ & $13.2$ & $55.1$ & $72.3$ & $36.9$ & $61.3$ & $77.1$ & $0.0$ & $83.4$ & $32.7$ & $71.0$ & $0.3$ & \underline{$90.0$} & \underline{$50.0$} & $84.1$ & \underline{$66.6$} & $65.8$ & $61.6$ & $35.2$
\\
FIDNet~\cite{zhao2021fidnet}~\small{['21]} & $56.4$ & \underline{$90.9$} & $37.1$ & $42.7$ & $72.8$ & $38.2$ & $56.4$ & $74.2$ & $0.0$ & $93.9$ & $34.1$ & $78.3$ & $6.6$ & $84.8$ & $47.2$ & $84.1$ & $60.5$ & $68.3$ & $\mathbf{63.0}$ & $38.7$
\\
SPVCNN~\cite{tang2020searching}~\small{['20]} & $56.9$ & $88.6$ & $25.7$ & $55.9$ & $67.4$ & $48.8$ & $65.0$ & \underline{$78.2$} & $0.0$ & $82.6$ & $30.4$ & $70.1$ & $0.3$ & $\mathbf{90.5}$ & $49.6$ & $84.4$ & $\mathbf{67.6}$ & $66.1$ & $61.6$ & $\mathbf{48.7}$
\\
Cylinder3D~\cite{zhu2021cylindrical}~\small{['21]} & $57.0$ & $88.5$ & \underline{$39.9$} & \underline{$58.0$} & $58.4$ & $48.1$ & \underline{$68.6$} & $77.0$ & $\mathbf{0.5}$ & $84.4$ & $30.4$ & $72.2$ & $2.5$ & $89.4$ & $48.4$ & $81.9$ & $64.6$ & $59.8$ & $61.2$ & $\mathbf{48.7}$
\\
CENet~\cite{cheng2022cenet}~\small{['22]} & \underline{$60.8$} & $87.6$ & $39.3$ & $55.8$ & $\mathbf{85.9}$ & $\mathbf{58.9}$ & $66.9$ & $74.0$ & $0.0$ & \underline{$94.5$} & $\mathbf{45.0}$ & $\mathbf{80.7}$ & $\mathbf{11.5}$ & $85.3$ & $49.7$ & \underline{$84.6$} & $58.4$ & \underline{$70.4$} & $62.6$ & $44.6$
\\
\cellcolor{LightCyan}\textbf{RangeFormer~(Ours)} & \cellcolor{LightCyan}$\mathbf{63.0}$ & \cellcolor{LightCyan}$\mathbf{92.6}$ & \cellcolor{LightCyan}$\mathbf{51.6}$ & \cellcolor{LightCyan}$\mathbf{65.7}$ & \cellcolor{LightCyan}\underline{$74.4$} & \cellcolor{LightCyan}\underline{$49.6$} & \cellcolor{LightCyan}$\mathbf{71.6}$ & \cellcolor{LightCyan}$\mathbf{82.1}$ & \cellcolor{LightCyan}$0.0$ & \cellcolor{LightCyan}$\mathbf{94.8}$ & \cellcolor{LightCyan}\underline{$44.4$} & \cellcolor{LightCyan}\underline{$80.6$} & \cellcolor{LightCyan}\underline{$11.4$} & \cellcolor{LightCyan}$85.6$ & \cellcolor{LightCyan}$\mathbf{56.9}$ & \cellcolor{LightCyan}$\mathbf{87.2}$ & \cellcolor{LightCyan}$64.1$ & \cellcolor{LightCyan}$\mathbf{77.0}$ & \cellcolor{LightCyan}\underline{$62.7$} & \cellcolor{LightCyan}\underline{$45.1$}
\\\bottomrule
\end{tabular}}
\label{table:scribblekitti-class}
\end{table*}

\clearpage

\begin{table*}[t]
\caption{\textbf{The class-wise IoU scores} of different LiDAR semantic segmentation approaches (raw point, bird's eye view, range view, voxel, and multi-view fusion) on the \textit{val} set of \textbf{nuScenes} \cite{Panoptic-nuScenes}. All IoU scores are given in percentage (\%). For each class: \textbf{bold} - best in column; \underline{underline} - second best in column. Symbol $^{\dagger}$: $W_{\text{train}}=480$. Methods are arranged in \textit{ascending} order of mIoU.}
\vspace{-0.1cm}
\centering\scalebox{0.662}{
\begin{tabular}{r|c|cccccccccccccccc}
\toprule
\textbf{Method~\small{(year)}} & \rotatebox{0}{mIoU} & \rotatebox{90}{barrier} & \rotatebox{90}{bicycle} & \rotatebox{90}{bus} & \rotatebox{90}{car} & \rotatebox{90}{construction-vehicle} & \rotatebox{90}{motorcycle} & \rotatebox{90}{pedestrian} & \rotatebox{90}{traffic-cone} & \rotatebox{90}{trailer} & \rotatebox{90}{truck} & \rotatebox{90}{driveable-surface} & \rotatebox{90}{other-ground} & \rotatebox{90}{sidewalk} & \rotatebox{90}{terrain} & \rotatebox{90}{manmade} & \rotatebox{90}{vegetation}
\\\midrule
AF2S3Net~\cite{2021AF2S3Net}~\small{['21]} & $62.2$ & $60.3$ & $12.6$ & $82.3$ & $80.0$ & $20.1$ & $62.0$ & $59.0$ & $49.0$ & $42.2$ & $67.4$ & $94.2$ & $68.0$ & $64.1$ & $68.6$ & $82.9$ & $82.4$
\\
RangeNet++~\cite{milioto2019rangenet++}~\small{['19]} & $65.5$ & $66.0$ & $21.3$ & $77.2$ & $80.9$ & $30.2$ & $66.8$ & $69.6$ & $52.1$ & $54.2$ & $72.3$ & $94.1$ & $66.6$ & $63.5$ & $70.1$ & $83.1$ & $79.8$
\\
PolarNet~\cite{zhang2020polarnet}~\small{['20]} & $71.0$ & $74.7$ & $28.2$ & $85.3$ & $90.9$ & $35.1$ & $77.5$ & $71.3$ & $58.8$ & $57.4$ & $76.1$ & $96.5$ & $71.1$ & $74.7$ & $74.0$ & $87.3$ & $85.7$
\\
PCSCNet~\cite{2023_PCSCNet}~\small{['22]} & $72.0$ & $73.3$ & $42.2$ & $87.8$ & $86.1$ & $44.9$ & $82.2$ & $76.1$ & $62.9$ & $49.3$ & $77.3$ & $95.2$ & $66.9$ & $69.5$ & $72.3$ & $83.7$ & $82.5$
\\
SalsaNext~\cite{cortinhal2020salsanext}~\small{['20]} & $72.2$ & $74.8$ & $34.1$ & $85.9$ & $88.4$ & $42.2$ & $72.4$ & $72.2$ & $63.1$ & $61.3$ & $76.5$ & $96.0$ & $70.8$ & $71.2$ & $71.5$ & $86.7$ & $84.4$
\\
SVASeg~\cite{2022_SVASeg}~\small{['22]} & $74.7$ & $73.1$ & $44.5$ & $88.4$ & $86.6$ & $48.2$ & $80.5$ & $77.7$ & $65.6$ & $57.5$ & $82.1$ & $96.5$ & $70.5$ & $74.7$ & $74.6$ & $87.3$ & $86.9$
\\
Cylinder3D~\cite{zhu2021cylindrical}~\small{['21]} & $76.1$ & $76.4$ & $40.3$ & $91.2$ & $\mathbf{93.8}$ & $51.3$ & $78.0$ & \underline{$78.9$} & $64.9$ & $62.1$ & \underline{$84.4$} & $96.8$ & $71.6$ & \underline{$76.4$} & $\mathbf{75.4}$ & $90.5$ & $87.4$
\\
AMVNet~\cite{2020AMVNet}~\small{['20]} & $76.1$ & $\mathbf{79.8}$ & $32.4$ & $82.2$ & $86.4$ & $\mathbf{62.5}$ & $81.9$ & $75.3$ & $\mathbf{72.3}$ & $\mathbf{83.5}$ & $65.1$ & $\mathbf{97.4}$ & $67.0$ & $\mathbf{78.8}$ & $74.6$ & $\mathbf{90.8}$ & \underline{$87.9$}
\\
\cellcolor{red!4}\textbf{STR$^{\dagger}$~(Ours)} & \cellcolor{red!4}$77.1$ & \cellcolor{red!4}$76.0$ & \cellcolor{red!4}\underline{$44.7$} & \cellcolor{red!4}$\mathbf{94.2}$ & \cellcolor{red!4}$92.2$ & \cellcolor{red!4}$54.2$ & \cellcolor{red!4}$82.1$ & \cellcolor{red!4}$76.7$ & \cellcolor{red!4}\underline{$69.3$} & \cellcolor{red!4}$61.8$ & \cellcolor{red!4}$83.4$ & \cellcolor{red!4}$96.7$ & \cellcolor{red!4}$\mathbf{75.7}$ & \cellcolor{red!4}$75.2$ & \cellcolor{red!4}$\mathbf{75.4}$ & \cellcolor{red!4}$88.8$ & \cellcolor{red!4}$87.3$
\\
RPVNet~\cite{xu2021rpvnet}~\small{['21]} & \underline{$77.6$} & \underline{$78.2$} & $43.4$ & $92.7$ & \underline{$93.2$} & $49.0$ & $\mathbf{85.7}$ & $\mathbf{80.5}$ & $66.0$ & \underline{$66.9$} & $84.0$ & \underline{$96.9$} & $73.5$ & $75.9$ & $70.6$ & \underline{$90.6$} & $\mathbf{88.9}$
\\
\cellcolor{LightCyan}\textbf{RangeFormer~(Ours)} & \cellcolor{LightCyan}$\mathbf{78.1}$ & \cellcolor{LightCyan}$78.0$ & \cellcolor{LightCyan}$\mathbf{45.2}$ & \cellcolor{LightCyan}\underline{$94.0$} & \cellcolor{LightCyan}$92.9$ & \cellcolor{LightCyan}\underline{$58.7$} & \cellcolor{LightCyan}\underline{$83.9$} & \cellcolor{LightCyan}$77.9$ & \cellcolor{LightCyan}$69.1$ & \cellcolor{LightCyan}$63.7$ & \cellcolor{LightCyan}$\mathbf{85.6}$ & \cellcolor{LightCyan}$96.7$ & \cellcolor{LightCyan}\underline{$74.5$} & \cellcolor{LightCyan}$75.1$ & \cellcolor{LightCyan}\underline{$75.3$} & \cellcolor{LightCyan}$89.1$ & \cellcolor{LightCyan}$87.5$
\\\bottomrule
\end{tabular}}
\label{table:nuscenes-class}
\end{table*}

\begin{table*}[t]
\caption{\textbf{The class-wise IoU scores} of different LiDAR semantic segmentation approaches (raw point, bird's eye view, range view, voxel, and multi-view fusion) on the \textit{test} set of \textbf{nuScenes}~\cite{Panoptic-nuScenes}. All IoU scores are given in percentage (\%). For each class: \textbf{bold} - best in column; \underline{underline} - second best in column. Symbol $^{\dagger}$: $W_{\text{train}}=480$. Methods are arranged in \textit{ascending} order of mIoU.}
\vspace{-0.1cm}
\centering\scalebox{0.662}{
\begin{tabular}{r|c|cccccccccccccccc}
\toprule
\textbf{Method~\small{(year)}} & \rotatebox{0}{mIoU} & \rotatebox{90}{barrier} & \rotatebox{90}{bicycle} & \rotatebox{90}{bus} & \rotatebox{90}{car} & \rotatebox{90}{construction-vehicle} & \rotatebox{90}{motorcycle} & \rotatebox{90}{pedestrian} & \rotatebox{90}{traffic-cone} & \rotatebox{90}{trailer} & \rotatebox{90}{truck} & \rotatebox{90}{driveable-surface} & \rotatebox{90}{other-ground} & \rotatebox{90}{sidewalk} & \rotatebox{90}{terrain} & \rotatebox{90}{manmade} & \rotatebox{90}{vegetation}
\\\midrule
PolarNet~\cite{zhang2020polarnet}~\small{['20]} & $ 69.4 $ & $  72.2 $ & $ 16.8 $ & $ 77.0 $ & $ 86.5 $ & $ 51.1 $ & $ 69.7 $ & $  64.8 $ & $ 54.1 $ & $ 69.7 $ & $ 63.5 $ & $ 96.6 $ & $ 67.1 $ & $ 77.7 $ & $ 72.1 $ & $ 87.1 $ & $ 84.5$
\\
JS3C-Net~\cite{yan2021sparse}~\small{['21]} & $ 73.6  $ & $ 80.1  $ & $ 26.2 $ & $ 87.8 $ & $ 84.5 $ & $ 55.2 $ & $ 72.6 $ & $ 71.3 $ & $ 66.3 $ & $ 76.8 $ & $ 71.2 $ & $ 96.8 $ & $ 64.5 $ & $ 76.9 $ & $ 74.1 $ & $ 87.5 $ & $ 86.1$
\\
PMF~\cite{zhuang2021perception}~\small{['21]} & $77.0 $ & $ 82.0$ & $ 40.0$ & $ 81.0$ & $ 88.0$ & $ 64.0 $ & $79.0$ & $ 80.0 $ & \underline{$ 76.0$} & $ 81.0$ & $ 67.0$ & $ 97.0$ & $ 68.0$ & $ 78.0$ & $ 74.0$ & $ 90.0$ & $ 88.0$
\\
Cylinder3D~\cite{zhu2021cylindrical}~\small{['21]}  & $ 77.2 $ & $ 82.8 $ & $ 29.8 $ & $ 84.3 $ & $ 89.4 $ & $ 63.0 $ & $ 79.3 $ & $ 77.2 $ & $ 73.4 $ & $ 84.6 $ & $ 69.1 $ & \underline{$ 97.7 $} & $ 70.2 $ & $ 80.3 $ & $ 75.5 $ & $ 90.4 $ & $ 87.6$  \\

AMVNet~\cite{2020AMVNet}~\small{['20]}  & $77.3 $ & $  80.6 $ & $ 32.0 $ & $ 81.7 $ & $ 88.9 $ & $ 67.1 $ & $ 84.3 $ & $ 76.1 $ & $ 73.5 $ & \underline{$84.9$} & $ 67.3 $ & $ 97.5 $ & $ 67.4 $ & $ 79.4 $ & $ 75.5 $ & $ 91.5 $ & $ 88.7$\\

SPVCNN~\cite{tang2020searching}~\small{['20]}  & $77.4$ & $ 80.0$ & $ 30.0 $ & $ 91.9 $ & $ 90.8 $ & $ 64.7 $ & $ 79.0 $ & $ 75.6 $ & $ 70.9 $ & $ 81.0 $ & $ 74.6 $ & $ 97.4 $ & $ 69.2 $ & $ 80.0 $ & $ 76.1 $ & $ 89.3 $ & $ 87.1$ \\

AF2S3Net~\cite{2021AF2S3Net}~\small{['21]}  & $78.3 $ & $ 78.9 $ & $ 52.2 $ & $ 89.9 $ & $ 84.2 $ & $\mathbf{77.4}$ & $ 74.3 $ & $ 77.3 $ & $ 72.0 $ & $ 83.9 $ & $ 73.8 $ & $ 97.1 $ & $ 66.5 $ & $ 77.5 $ & $ 74.0 $ & $ 87.7 $ & $ 86.8$ \\

\cellcolor{red!4}\textbf{STR$^{\dagger}$~(Ours)} & \cellcolor{red!4}$78.7$ & \cellcolor{red!4}$83.9$ & \cellcolor{red!4}$46.3$ & \cellcolor{red!4}$90.0$ & \cellcolor{red!4}$87.2$ & \cellcolor{red!4}$69.5$ & \cellcolor{red!4}$83.4$ & \cellcolor{red!4}$75.5$ & \cellcolor{red!4}$73.6$ & \cellcolor{red!4}$82.1$ & \cellcolor{red!4}$71.7$ & \cellcolor{red!4}$97.0$ & \cellcolor{red!4}$69.1$ & \cellcolor{red!4}$78.1$ & \cellcolor{red!4}$74.1$ & \cellcolor{red!4}$90.1$ & \cellcolor{red!4}$87.0$ 

\\
2D3DNet~\cite{genova2021learning}~\small{['21]}  & $ 80.0 $ & $ 83.0 $ & $\mathbf{59.4}$ & $ 88.0 $ & $85.1 $ & $ 63.7 $ & $ 84.4 $ & $ 82.0 $ & \underline{$ 76.0 $} & $ 84.8 $ & $ 71.9 $ & $ 96.9 $ & $ 67.4 $ & $ 79.8 $ & $ 76.0 $ & $ \mathbf{92.1} $ & $ 89.2 $\\

\cellcolor{LightCyan}\textbf{RangeFormer~(Ours)} & \cellcolor{LightCyan}$80.1$ & \cellcolor{LightCyan}$\mathbf{85.6}$ & \cellcolor{LightCyan}$47.4$ & \cellcolor{LightCyan}$91.2$ & \cellcolor{LightCyan}$90.9$ & \cellcolor{LightCyan}$70.7$ & \cellcolor{LightCyan}$84.7$ & \cellcolor{LightCyan}$77.1$ & \cellcolor{LightCyan}$74.1$ & \cellcolor{LightCyan}$83.2$ & \cellcolor{LightCyan}$72.6$ & \cellcolor{LightCyan}$97.5$ & \cellcolor{LightCyan}\underline{$70.7$} & \cellcolor{LightCyan}$79.2$ & \cellcolor{LightCyan}$75.4$ & \cellcolor{LightCyan}$91.3$ & \cellcolor{LightCyan}$88.9$
\\

GASN~\cite{2022GASN}~\small{['22]}  & $80.4 $ & \underline{$ 85.5 $} & $ 43.2 $ & $ 90.5 $ & $ \mathbf{92.1} $ & $ 64.7 $ & $ 86.0 $ & \underline{$ 83.0 $} & $ 73.3 $ & $ 83.9 $ & $\mathbf{75.8}$ & $ 97.0 $ & $\mathbf{71.0}$ & $ \mathbf{81.0} $ & $ \mathbf{77.7} $ & \underline{$ 91.6 $} & $ \mathbf{90.2}$\\

2DPASS~\cite{2022_2DPASS}~\small{['22]}  & \underline{$ 80.8 $} & $ 81.7 $ & \underline{$ 55.3 $} & \underline{$ 92.0 $} & \underline{$ 91.8 $} & \underline{$ 73.3 $} & \underline{$ 86.5 $} & $ 78.5 $ & $ 72.5 $ & $ 84.7 $ & \underline{$ 75.5 $} & $ 97.6 $ & $ 69.1 $ & $ 79.9 $ & $ 75.5 $ & $ 90.2 $ & $ 88.0$\\

LidarMultiNet~\cite{lidarmultinet}~\small{['22]}  & $\mathbf{81.4} $ & $ 80.4 $ & $ 48.4 $ & $ \mathbf{94.3} $ & $ 90.0 $ & $ 71.5 $ & $\mathbf{87.2}$ & $ \mathbf{85.2} $ & $ \mathbf{80.4} $ & $ \mathbf{86.9} $ & $ 74.8 $ & $ \mathbf{97.8} $ & $ 67.3 $ & \underline{$ 80.7 $} & \underline{$ 76.5 $} & $ \mathbf{92.1} $ & \underline{$ 89.6$}
\\\bottomrule
\end{tabular}}
\label{table:nuscenes-class-test}
\end{table*}

\begin{table*}[t]
\caption{\textbf{The class-wise scores} of SoTA \textbf{LiDAR panoptic segmentation} approaches on the \textbf{SemanticKITTI}~\cite{SemanticKITTI} leaderboard. All scores are given in percentage (\%). For each class in each metric: \textbf{bold} - best in column; \underline{underline} - second best in column. Symbol $^{\dagger}$: $W_{\text{train}}=384$.}
\vspace{-0.2cm}
\centering\scalebox{0.652}{
\begin{tabular}{r|ccccccccccccccccccc|c}
\toprule
\textbf{Method~\small{(year)}} & \rotatebox{90}{car} & \rotatebox{90}{bicycle} & \rotatebox{90}{motorcycle} & \rotatebox{90}{truck} & \rotatebox{90}{other-vehicle} & \rotatebox{90}{person} & \rotatebox{90}{bicyclist} & \rotatebox{90}{motorcyclist} & \rotatebox{90}{road} & \rotatebox{90}{parking} & \rotatebox{90}{sidewalk} & \rotatebox{90}{other-ground} & \rotatebox{90}{building} & \rotatebox{90}{fence} & \rotatebox{90}{vegetation} & \rotatebox{90}{trunk} & \rotatebox{90}{terrain} & \rotatebox{90}{pole} & \rotatebox{90}{traffic-sign} & \rotatebox{90}{average}
\\\midrule
Metric & \multicolumn{20}{c}{Panoptic Quality ($\text{PQ}$)}
\\\midrule
Panoptic-PHNet~\cite{Panoptic-PHNet}~\small{['22]} & $\mathbf{94.0}$ & $\mathbf{54.6}$ & \underline{$62.4$} & \underline{$45.1$} & $51.2$ & $\mathbf{74.4}$ & $\mathbf{76.3}$ & $52.0$ & $\mathbf{89.9}$ & $49.4$ & $70.6$ & $11.7$ & $\mathbf{87.8}$ & $52.6$ & $79.4$ & $57.2$ & $45.0$ & $54.5$ & $61.2$ & $61.5$
\\
\cellcolor{LightCyan}\textbf{Panoptic-RangeFormer} & \cellcolor{LightCyan}\underline{$87.1$} & \cellcolor{LightCyan}\underline{$43.7$} & \cellcolor{LightCyan}$\mathbf{64.8}$ & \cellcolor{LightCyan}$\mathbf{49.2}$ & \cellcolor{LightCyan}$\mathbf{56.7}$ & \cellcolor{LightCyan}\underline{$65.2$} & \cellcolor{LightCyan}\underline{$75.0$} & \cellcolor{LightCyan}$\mathbf{66.7}$ & \cellcolor{LightCyan}$\mathbf{89.9}$ & \cellcolor{LightCyan}$\mathbf{57.8}$ & \cellcolor{LightCyan}$\mathbf{71.3}$ & \cellcolor{LightCyan}\underline{$20.1$} & \cellcolor{LightCyan}\underline{$87.7$} & \cellcolor{LightCyan}$\mathbf{59.0}$ & \cellcolor{LightCyan}$\mathbf{82.4}$ & \cellcolor{LightCyan}$\mathbf{62.3}$ & \cellcolor{LightCyan}$\mathbf{49.6}$ & \cellcolor{LightCyan}$\mathbf{61.3}$ & \cellcolor{LightCyan}$\mathbf{69.4}$ & \cellcolor{LightCyan}$\mathbf{64.2}$
\\\midrule
\cellcolor{red!4}\textbf{\textit{w/} STR}$^{\dagger}$ & \cellcolor{red!4}$85.6$ & \cellcolor{red!4}$40.1$ & \cellcolor{red!4}$61.1$ & \cellcolor{red!4}\underline{$45.1$} & \cellcolor{red!4}\underline{$52.9$} & \cellcolor{red!4}$61.1$ & \cellcolor{red!4}$72.0$ & \cellcolor{red!4}\underline{$64.3$} & \cellcolor{red!4}\underline{$89.3$} & \cellcolor{red!4}\underline{$57.7$} & \cellcolor{red!4}\underline{$71.1$} & \cellcolor{red!4}$\mathbf{21.5}$ & \cellcolor{red!4}$87.0$ & \cellcolor{red!4}\underline{$56.7$} & \cellcolor{red!4}\underline{$80.6$} & \cellcolor{red!4}\underline{$57.5$} & \cellcolor{red!4}\underline{$49.0$} & \cellcolor{red!4}\underline{$56.1$} & \cellcolor{red!4}\underline{$65.2$} & \cellcolor{red!4}\underline{$61.8$}
\\\midrule\midrule
Metric & \multicolumn{20}{c}{Recognition Quality ($\text{RQ}$)}
\\\midrule
Panoptic-PHNet~\cite{Panoptic-PHNet}~\small{['22]} & $\mathbf{98.6}$ & $\mathbf{71.8}$ & $69.9$ & $47.5$ & $54.9$ & $\mathbf{82.8}$ & \underline{$83.2$} & $54.6$ & \underline{$96.0$} & $63.2$ & $\mathbf{85.3}$ & $15.6$ & \underline{$93.4$} & $68.6$ & $95.0$ & $77.0$ & $59.0$ & $72.5$ & $80.2$ & $72.1$
\\
\cellcolor{LightCyan}\textbf{Panoptic-RangeFormer} & \cellcolor{LightCyan}\underline{$96.2$} & \cellcolor{LightCyan}\underline{$60.1$} & \cellcolor{LightCyan}$\mathbf{74.7}$ & \cellcolor{LightCyan}$\mathbf{54.2}$ & \cellcolor{LightCyan}$\mathbf{62.3}$ & \cellcolor{LightCyan}\underline{$77.4$} & \cellcolor{LightCyan}$\mathbf{84.7}$ & \cellcolor{LightCyan}$\mathbf{74.4}$ & \cellcolor{LightCyan}$\mathbf{96.2}$ & \cellcolor{LightCyan}$\mathbf{71.5}$ & \cellcolor{LightCyan}$\mathbf{85.3}$ & \cellcolor{LightCyan}\underline{$27.2$} & \cellcolor{LightCyan}$\mathbf{94.1}$ & \cellcolor{LightCyan}$\mathbf{75.8}$ & \cellcolor{LightCyan}$\mathbf{96.8}$ & \cellcolor{LightCyan}$\mathbf{81.8}$ & \cellcolor{LightCyan}$\mathbf{64.4}$ & \cellcolor{LightCyan}$\mathbf{80.4}$ & \cellcolor{LightCyan}$\mathbf{85.1}$ & \cellcolor{LightCyan}$\mathbf{75.9}$
\\\midrule
\cellcolor{red!4}\textbf{\textit{w/} STR}$^{\dagger}$ & \cellcolor{red!4}$95.7$ & \cellcolor{red!4}$56.0$ & \cellcolor{red!4}\underline{$70.3$} & \cellcolor{red!4}\underline{$49.0$} & \cellcolor{red!4}\underline{$58.1$} & \cellcolor{red!4}$73.5$ & \cellcolor{red!4}$82.2$ & \cellcolor{red!4}\underline{$72.5$} & \cellcolor{red!4}$95.7$ & \cellcolor{red!4}\underline{$71.2$} & \cellcolor{red!4}\underline{$85.1$} & \cellcolor{red!4}$\mathbf{29.1}$ & \cellcolor{red!4}$\mathbf{94.1}$ & \cellcolor{red!4}\underline{$73.6$} & \cellcolor{red!4}\underline{$95.9$} & \cellcolor{red!4}\underline{$77.9$} & \cellcolor{red!4}\underline{$63.8$} & \cellcolor{red!4}\underline{$76.2$} & \cellcolor{red!4}\underline{$81.9$} & \cellcolor{red!4}\underline{$73.8$}
\\\midrule\midrule
Metric & \multicolumn{20}{c}{Segmentation Quality ($\text{SQ}$)}
\\\midrule
Panoptic-PHNet~\cite{Panoptic-PHNet}~\small{['22]} & $\mathbf{95.4}$ & $\mathbf{76.0}$ & $\mathbf{89.3}$ & $\mathbf{95.0}$ & $\mathbf{93.3}$ & $\mathbf{89.8}$ & $\mathbf{91.7}$ & $\mathbf{95.2}$ & $\mathbf{93.6}$ & $78.1$ & $82.8$ & $\mathbf{75.0}$ & $\mathbf{94.1}$ & $76.6$ & $83.6$ & \underline{$74.3$} & $76.3$ & \underline{$75.2$} & $76.3$ & $\mathbf{84.8}$
\\
\cellcolor{LightCyan}\textbf{Panoptic-RangeFormer} & \cellcolor{LightCyan}\underline{$90.5$} & \cellcolor{LightCyan}\underline{$72.6$} & \cellcolor{LightCyan}$86.8$ & \cellcolor{LightCyan}$90.9$ & \cellcolor{LightCyan}$91.0$ & \cellcolor{LightCyan}\underline{$84.3$} & \cellcolor{LightCyan}\underline{$88.5$} & \cellcolor{LightCyan}\underline{$89.6$} & \cellcolor{LightCyan}\underline{$93.4$} & \cellcolor{LightCyan}\underline{$80.8$} & \cellcolor{LightCyan}$\mathbf{83.6}$ & \cellcolor{LightCyan}\underline{$74.0$} & \cellcolor{LightCyan}\underline{$93.3$} & \cellcolor{LightCyan}$\mathbf{77.8}$ & \cellcolor{LightCyan}$\mathbf{85.1}$ & \cellcolor{LightCyan}$\mathbf{76.2}$ & \cellcolor{LightCyan}$\mathbf{76.9}$ & \cellcolor{LightCyan}$\mathbf{76.3}$ & \cellcolor{LightCyan}$\mathbf{81.5}$ & \cellcolor{LightCyan}\underline{$83.8$}
\\\midrule
\cellcolor{red!4}\textbf{\textit{w/} STR}$^{\dagger}$ & \cellcolor{red!4}$89.5$ & \cellcolor{red!4}$71.6$ & \cellcolor{red!4}\underline{$86.9$} & \cellcolor{red!4}\underline{$92.0$} & \cellcolor{red!4}\underline{$91.1$} & \cellcolor{red!4}$83.2$ & \cellcolor{red!4}$87.7$ & \cellcolor{red!4}$88.7$ & \cellcolor{red!4}\underline{$93.4$} & \cellcolor{red!4}$\mathbf{80.9}$ & \cellcolor{red!4}\underline{$83.5$} & \cellcolor{red!4}$73.9$ & \cellcolor{red!4}$92.5$ & \cellcolor{red!4}\underline{$77.1$} & \cellcolor{red!4}\underline{$84.0$} & \cellcolor{red!4}$73.8$ & \cellcolor{red!4}\underline{$76.8$} & \cellcolor{red!4}$73.6$ & \cellcolor{red!4}\underline{$79.6$} & \cellcolor{red!4}$83.1$
\\\midrule\midrule
Metric & \multicolumn{20}{c}{Intersection-over-Union ($\text{IoU}$)}
\\\midrule
Panoptic-PHNet~\cite{Panoptic-PHNet}~\small{['22]} & $\mathbf{96.3}$ & $59.4$ & $55.5$ & $56.4$ & $48.0$ & $66.2$ & $\mathbf{70.0}$ & $22.9$ & $92.1$ & $67.9$ & $77.5$ & \underline{$33.0$} & $\mathbf{92.8}$ & $68.5$ & $84.9$ & $69.3$ & $69.8$ & $61.2$ & $62.2$ & $66.0$
\\
\cellcolor{LightCyan}\textbf{Panoptic-RangeFormer} & \cellcolor{LightCyan}$\mathbf{96.3}$ & \cellcolor{LightCyan}$\mathbf{64.4}$ & \cellcolor{LightCyan}$\mathbf{70.2}$ & \cellcolor{LightCyan}$\mathbf{60.2}$ & \cellcolor{LightCyan}\underline{$65.8$} & \cellcolor{LightCyan}$\mathbf{72.2}$ & \cellcolor{LightCyan}\underline{$68.9$} & \cellcolor{LightCyan}$\mathbf{57.3}$ & \cellcolor{LightCyan}$\mathbf{92.3}$ & \cellcolor{LightCyan}$\mathbf{72.7}$ & \cellcolor{LightCyan}$\mathbf{78.4}$ & \cellcolor{LightCyan}$\mathbf{42.3}$ & \cellcolor{LightCyan}\underline{$92.2$} & \cellcolor{LightCyan}$\mathbf{69.9}$ & \cellcolor{LightCyan}$\mathbf{86.6}$ & \cellcolor{LightCyan}$\mathbf{73.3}$ & \cellcolor{LightCyan}$\mathbf{72.7}$ & \cellcolor{LightCyan}$\mathbf{66.1}$ & \cellcolor{LightCyan}$\mathbf{66.5}$ & \cellcolor{LightCyan}$\mathbf{72.0}$
\\\midrule
\cellcolor{red!4}\textbf{\textit{w/} STR}$^{\dagger}$ & \cellcolor{red!4}\underline{$96.0$} & \cellcolor{red!4}\underline{$62.3$} & \cellcolor{red!4}\underline{$68.7$} & \cellcolor{red!4}\underline{$59.2$} & \cellcolor{red!4}$\mathbf{66.8}$ & \cellcolor{red!4}\underline{$70.0$} & \cellcolor{red!4}$67.8$ & \cellcolor{red!4}\underline{$56.8$} & \cellcolor{red!4}$92.0$ & \cellcolor{red!4}\underline{$72.3$} & \cellcolor{red!4}\underline{$78.0$} & \cellcolor{red!4}$\mathbf{42.3}$ & \cellcolor{red!4}$91.8$ & \cellcolor{red!4}\underline{$69.5$} & \cellcolor{red!4}\underline{$85.8$} & \cellcolor{red!4}\underline{$70.5$} & \cellcolor{red!4}\underline{$72.3$} & \cellcolor{red!4}\underline{$62.6$} & \cellcolor{red!4}\underline{$64.8$} & \cellcolor{red!4}\underline{$71.0$}
\\\bottomrule
\end{tabular}}
\label{table:panoptic-class}
\end{table*}

\begin{table*}[t]
\caption{\textbf{The class-wise IoU scores} of RangeFormer \textit{with} and \textit{without} (denoted as ${\S}$) test-time augmentation on the \textbf{SemanticKITTI}~\cite{SemanticKITTI} leaderboard. All IoU scores are given in percentage (\%).}
\vspace{-0.1cm}
\centering\scalebox{0.662}{
\begin{tabular}{r|c|ccccccccccccccccccc}
\toprule
\textbf{Method} & \rotatebox{0}{$\text{mIoU}$} & \rotatebox{90}{car} & \rotatebox{90}{bicycle} & \rotatebox{90}{motorcycle} & \rotatebox{90}{truck} & \rotatebox{90}{other-vehicle} & \rotatebox{90}{person} & \rotatebox{90}{bicyclist} & \rotatebox{90}{motorcyclist} & \rotatebox{90}{road} & \rotatebox{90}{parking} & \rotatebox{90}{sidewalk} & \rotatebox{90}{other-ground} & \rotatebox{90}{building} & \rotatebox{90}{fence} & \rotatebox{90}{vegetation} & \rotatebox{90}{trunk} & \rotatebox{90}{terrain} & \rotatebox{90}{pole} & \rotatebox{90}{traffic-sign}
\\\midrule
\textbf{RangeFormer}$^{~\S}$ & $69.5$ & $94.7$ & $60.0$ & $69.7$ & $57.9$ & $64.1$ & $72.3$ & $72.5$ & $54.9$ & $90.3$ & $69.9$ & $74.9$ & $38.9$ & $90.2$ & $66.1$ & $84.1$ & $68.1$ & $70.0$ & $58.9$ & $63.1$
\\\midrule
\cellcolor{LightCyan}\textbf{RangeFormer} & \cellcolor{LightCyan}$73.3$ & \cellcolor{LightCyan}$96.7$ & \cellcolor{LightCyan}$69.4$ & \cellcolor{LightCyan}$73.7$ & \cellcolor{LightCyan}$59.9$ & \cellcolor{LightCyan}$66.2$ & \cellcolor{LightCyan}$78.1$ & \cellcolor{LightCyan}$75.9$ & \cellcolor{LightCyan}$58.1$ & \cellcolor{LightCyan}$92.4$ & \cellcolor{LightCyan}$73.0$ & \cellcolor{LightCyan}$78.8$ & \cellcolor{LightCyan}$42.4$ & \cellcolor{LightCyan}$92.3$ & \cellcolor{LightCyan}$70.1$ & \cellcolor{LightCyan}$86.6$ & \cellcolor{LightCyan}$73.3$ & \cellcolor{LightCyan}$72.8$ & \cellcolor{LightCyan}$66.4$ & \cellcolor{LightCyan}$66.6$
\\\bottomrule
\end{tabular}}
\label{table:semantickitti-class-notta}
\end{table*}

\begin{table*}[t]
\caption{\textbf{The class-wise IoU scores} of RangeFormer \textit{with} and \textit{without} (denoted as ${\S}$) test-time augmentation on the \textbf{nuScenes}~\cite{Panoptic-nuScenes} leaderboard. All IoU scores are given in percentage (\%).}
\vspace{-0.1cm}
\centering\scalebox{0.662}{
\begin{tabular}{r|c|cccccccccccccccc}
\toprule
\textbf{Method} & \rotatebox{0}{mIoU} & \rotatebox{90}{barrier} & \rotatebox{90}{bicycle} & \rotatebox{90}{bus} & \rotatebox{90}{car} & \rotatebox{90}{construction-vehicle} & \rotatebox{90}{motorcycle} & \rotatebox{90}{pedestrian} & \rotatebox{90}{traffic-cone} & \rotatebox{90}{trailer} & \rotatebox{90}{truck} & \rotatebox{90}{driveable-surface} & \rotatebox{90}{other-ground} & \rotatebox{90}{sidewalk} & \rotatebox{90}{terrain} & \rotatebox{90}{manmade} & \rotatebox{90}{vegetation}
\\\midrule
\textbf{RangeFormer}$^{~\S}$ & $78.3$ & $83.9$ & $46.1$ & $89.4$ & $89.2$ & $70.3$ & $83.3$ & $75.4$ & $72.5$ & $81.4$ & $71.1$ & $95.6$ & $68.5$ & $77.3$ & $73.4$ & $89.3$ & $86.9$
\\\midrule
\cellcolor{LightCyan}\textbf{RangeFormer} & \cellcolor{LightCyan}$80.1$ & \cellcolor{LightCyan}$85.6$ & \cellcolor{LightCyan}$47.4$ & \cellcolor{LightCyan}$91.2$ & \cellcolor{LightCyan}$90.9$ & \cellcolor{LightCyan}$70.7$ & \cellcolor{LightCyan}$84.7$ & \cellcolor{LightCyan}$77.1$ & \cellcolor{LightCyan}$74.1$ & \cellcolor{LightCyan}$83.2$ & \cellcolor{LightCyan}$72.6$ & \cellcolor{LightCyan}$97.5$ & \cellcolor{LightCyan}$70.7$ & \cellcolor{LightCyan}$79.2$ & \cellcolor{LightCyan}$75.4$ & \cellcolor{LightCyan}$91.3$ & \cellcolor{LightCyan}$88.9$
\\\bottomrule
\end{tabular}}
\label{table:nuscenes-class-test-notta}
\end{table*}

\begin{table*}[t]
\caption{\textbf{The class-wise IoU scores} of different \textbf{STR Partition Strategies} on the \textit{val} set of \textbf{SemanticKITTI}~\cite{SemanticKITTI}. All IoU scores are given in percentage (\%). For each class in each partition: \textbf{bold} - best in column; \underline{underline} - second best in column. Note that we have applied the proposed \textit{RangeAug} and \textit{RangePost} to FIDNet~\cite{zhao2021fidnet} and CENet~\cite{cheng2022cenet}, so the comparisons are only correlated to the model architecture.}
\vspace{-0.2cm}
\centering\scalebox{0.662}{
\begin{tabular}{r|c|ccccccccccccccccccc}
\toprule
\textbf{Method~\small{(year)}} & \rotatebox{0}{$\text{mIoU}$} & \rotatebox{90}{car} & \rotatebox{90}{bicycle} & \rotatebox{90}{motorcycle} & \rotatebox{90}{truck} & \rotatebox{90}{other-vehicle} & \rotatebox{90}{person} & \rotatebox{90}{bicyclist} & \rotatebox{90}{motorcyclist} & \rotatebox{90}{road} & \rotatebox{90}{parking} & \rotatebox{90}{sidewalk} & \rotatebox{90}{other-ground} & \rotatebox{90}{building} & \rotatebox{90}{fence} & \rotatebox{90}{vegetation} & \rotatebox{90}{trunk} & \rotatebox{90}{terrain} & \rotatebox{90}{pole} & \rotatebox{90}{traffic-sign}
\\\midrule
STR Partition & \multicolumn{20}{c}{$Z=10$ ($W_{\text{train}}=192$)}
\\\midrule
FIDNet~\cite{zhao2021fidnet}~\small{['21]} & $61.1$ & \underline{$92.5$} & \underline{$51.0$} & \underline{$52.1$} & $61.9$ & \underline{$50.8$} & $70.7$ & $79.4$ & $0.0$ & $93.7$ & $42.4$ & $79.8$ & $\mathbf{14.8}$ & $85.6$ & \underline{$54.0$} & \underline{$85.3$} & $62.7$ & \underline{$70.8$} & $\mathbf{64.0}$ & $48.8$
\\
CENet~\cite{cheng2022cenet}~\small{['22]} & \underline{$61.9$} & $90.2$ & $50.8$ & $56.8$ & \underline{$76.9$} & $44.7$ & \underline{$73.6$} & \underline{$82.3$} & $\mathbf{0.6}$ & \underline{$94.2$} & \underline{$42.6$} & \underline{$80.4$} & \underline{$14.7$} & \underline{$86.2$} & $52.8$ & $84.5$ & \underline{$63.4$} & $69.4$ & \underline{$63.4$} & \underline{$49.0$}
\\
\cellcolor{red!4}\textbf{RangeFormer} & \cellcolor{red!4}$\mathbf{64.3}$ & \cellcolor{red!4}$\mathbf{93.5}$ & \cellcolor{red!4}$\mathbf{57.0}$ & \cellcolor{red!4}$\mathbf{62.3}$ & \cellcolor{red!4}$\mathbf{79.9}$ & \cellcolor{red!4}$\mathbf{56.5}$ & \cellcolor{red!4}$\mathbf{74.7}$ & \cellcolor{red!4}$\mathbf{84.3}$ & \cellcolor{red!4}\underline{$0.2$} & \cellcolor{red!4}$\mathbf{94.3}$ & \cellcolor{red!4}$\mathbf{51.5}$ & \cellcolor{red!4}$\mathbf{81.0}$ & \cellcolor{red!4}$9.0$ & \cellcolor{red!4}$\mathbf{88.2}$ & \cellcolor{red!4}$\mathbf{63.4}$ & \cellcolor{red!4}$\mathbf{86.1}$ & \cellcolor{red!4}$\mathbf{66.4}$ & \cellcolor{red!4}$\mathbf{72.4}$ & \cellcolor{red!4}$62.7$ & \cellcolor{red!4}$\mathbf{50.8}$
\\\midrule\midrule
STR Partition & \multicolumn{20}{c}{$Z=8$ ($W_{\text{train}}=240$)}
\\\midrule
FIDNet~\cite{zhao2021fidnet}~\small{['21]} & $61.7$ & \underline{$92.7$} & \underline{$50.9$} & $52.5$ & \underline{$71.8$} & $50.9$ & $71.3$ & $79.3$ & $0.1$ & $93.6$ & $40.6$ & $79.8$ & $\mathbf{18.4}$ & $85.5$ & $54.2$ & \underline{$85.2$} & $62.6$ & \underline{$70.6$} & \underline{$63.6$} & \underline{$48.9$}
\\
CENet~\cite{cheng2022cenet}~\small{['22]} & \underline{$62.2$} & $92.0$ & $50.4$ & \underline{$56.7$} & $70.0$ & \underline{$51.1$} & \underline{$72.6$} & \underline{$80.2$} & \underline{$0.3$} & \underline{$94.2$} & \underline{$43.4$} & \underline{$80.4$} & \underline{$14.9$} & \underline{$86.3$} & \underline{$56.5$} & $\mathbf{85.3}$ & \underline{$64.2$} & $\mathbf{71.3}$ & $\mathbf{64.5}$ & $47.7$
\\
\cellcolor{red!4}\textbf{RangeFormer} & \cellcolor{red!4}$\mathbf{65.5}$ & \cellcolor{red!4}$\mathbf{94.3}$ & \cellcolor{red!4}$\mathbf{56.8}$ & \cellcolor{red!4}$\mathbf{66.2}$ & \cellcolor{red!4}$\mathbf{88.3}$ & \cellcolor{red!4}$\mathbf{59.7}$ & \cellcolor{red!4}$\mathbf{76.6}$ & \cellcolor{red!4}$\mathbf{83.4}$ & \cellcolor{red!4}$\mathbf{1.3}$ & \cellcolor{red!4}$\mathbf{94.6}$ & \cellcolor{red!4}$\mathbf{55.7}$ & \cellcolor{red!4}$\mathbf{81.6}$ & \cellcolor{red!4}$10.0$ & \cellcolor{red!4}$\mathbf{88.2}$ & \cellcolor{red!4}\underline{$55.1$} & \cellcolor{red!4}$84.5$ & \cellcolor{red!4}$\mathbf{65.7}$ & \cellcolor{red!4}$67.4$ & \cellcolor{red!4}$63.2$ & \cellcolor{red!4}$\mathbf{51.4}$ 
\\\midrule\midrule
STR Partition & \multicolumn{20}{c}{$Z=6$ ($W_{\text{train}}=320$)}
\\\midrule
FIDNet~\cite{zhao2021fidnet}~\small{['21]} & $62.2$ & $92.9$ & \underline{$52.4$} & $51.2$ & $70.6$ & $48.4$ & $72.7$ & \underline{$82.7$} & \underline{$0.2$} & $93.9$ & \underline{$43.5$} & $80.3$ & $\mathbf{16.9}$ & $86.1$ & $56.6$ & \underline{$85.6$} & $63.1$ & \underline{$71.6$} & $\mathbf{64.7}$ & $47.5$
\\
CENet~\cite{cheng2022cenet}~\small{['22]} & \underline{$62.7$} & \underline{$93.3$} & $47.9$ & \underline{$52.9$} & \underline{$84.4$} & \underline{$51.9$} & \underline{$74.1$} & $78.5$ & $\mathbf{0.4}$ & \underline{$94.2$} & $42.7$ & \underline{$80.7$} & \underline{$13.0$} & \underline{$86.6$} & \underline{$60.3$} & $84.8$ & \underline{$63.2$} & $69.5$ & \underline{$64.1$} & \underline{$48.7$}
\\
\cellcolor{red!4}\textbf{RangeFormer} & \cellcolor{red!4}$\mathbf{66.5}$ & \cellcolor{red!4}$\mathbf{95.0}$ & \cellcolor{red!4}$\mathbf{58.1}$ & \cellcolor{red!4}$\mathbf{72.1}$ & \cellcolor{red!4}$\mathbf{85.1}$ & \cellcolor{red!4}$\mathbf{59.8}$ & \cellcolor{red!4}$\mathbf{76.9}$ & \cellcolor{red!4}$\mathbf{86.4}$ & \cellcolor{red!4}\underline{$0.2$} & \cellcolor{red!4}$\mathbf{94.8}$ & \cellcolor{red!4}$\mathbf{55.5}$ & \cellcolor{red!4}$\mathbf{81.7}$ & \cellcolor{red!4}\underline{$13.0$} & \cellcolor{red!4}$\mathbf{88.5}$ & \cellcolor{red!4}$\mathbf{64.5}$ & \cellcolor{red!4}$\mathbf{86.5}$ & \cellcolor{red!4}$\mathbf{66.8}$ & \cellcolor{red!4}$\mathbf{73.0}$ & \cellcolor{red!4}$64.0$ & \cellcolor{red!4}$\mathbf{52.0}$
\\\midrule\midrule
STR Partition & \multicolumn{20}{c}{$Z=5$ ($W_{\text{train}}=384$)}
\\\midrule
FIDNet~\cite{zhao2021fidnet}~\small{['21]} & $62.2$ & $93.0$ & \underline{$52.9$} & $50.1$ & $73.7$ & \underline{$52.1$} & $72.3$ & $82.2$ & \underline{$0.3$} & $93.8$ & \underline{$42.7$} & $79.7$ & $\mathbf{13.8}$ & \underline{$86.1$} & $56.2$ & \underline{$85.6$} & $63.6$ & $\mathbf{71.7}$ & \underline{$64.6$} & \underline{$48.0$}
\\
CENet~\cite{cheng2022cenet}~\small{['22]} & \underline{$63.3$} & \underline{$93.2$} & $52.6$ & \underline{$59.9$} & \underline{$80.4$} & $50.6$ & \underline{$74.6$} & \underline{$82.3$} & $\mathbf{1.2}$ & \underline{$94.3$} & $42.1$ & \underline{$80.6$} & $\mathbf{13.8}$ & $86.0$ & \underline{$57.2$} & $85.2$ & \underline{$64.7$} & \underline{$70.5$} & $\mathbf{65.1}$ & $47.7$
\\
\cellcolor{red!4}\textbf{RangeFormer} & \cellcolor{red!4}$\mathbf{67.6}$ & \cellcolor{red!4}$\mathbf{95.3}$ & \cellcolor{red!4}$\mathbf{58.9}$ & \cellcolor{red!4}$\mathbf{73.4}$ & \cellcolor{red!4}$\mathbf{91.3}$ & \cellcolor{red!4}$\mathbf{68.0}$ & \cellcolor{red!4}$\mathbf{78.5}$ & \cellcolor{red!4}$\mathbf{87.5}$ & \cellcolor{red!4}$0.0$ & \cellcolor{red!4}$\mathbf{95.1}$ & \cellcolor{red!4}$\mathbf{49.1}$ & \cellcolor{red!4}$\mathbf{82.1}$ & \cellcolor{red!4}\underline{$10.8$} & \cellcolor{red!4}$\mathbf{89.2}$ & \cellcolor{red!4}$\mathbf{67.9}$ & \cellcolor{red!4}$\mathbf{85.7}$ & \cellcolor{red!4}$\mathbf{67.7}$ & \cellcolor{red!4}$70.4$ & \cellcolor{red!4}$64.4$ & \cellcolor{red!4}$\mathbf{52.0}$
\\\midrule\midrule
STR Partition & \multicolumn{20}{c}{$Z=4$ ($W_{\text{train}}=480$)}
\\\midrule
FIDNet~\cite{zhao2021fidnet}~\small{['21]} & $62.4$ & \underline{$92.6$} & $52.7$ & $56.8$ & $72.2$ & $49.3$ & \underline{$72.7$} & \underline{$82.0$} & \underline{$2.0$} & $93.8$ & $41.6$ & $79.9$ & $\mathbf{17.0}$ & \underline{$86.0$} & \underline{$55.7$} & $85.1$ & $63.0$ & $70.2$ & \underline{$64.8$} & \underline{$49.0$}
\\
CENet~\cite{cheng2022cenet}~\small{['22]} & \underline{$63.9$} & \underline{$92.6$} & \underline{$54.3$} & \underline{$63.6$} & \underline{$88.0$} & \underline{$53.6$} & $72.2$ & $80.1$ & $\mathbf{3.1}$ & \underline{$94.3$} & \underline{$46.2$} & \underline{$80.7$} & \underline{$14.8$} & $85.3$ & $50.7$ & \underline{$85.3$} & \underline{$63.7$} & \underline{$71.2$} & $\mathbf{65.6}$ & \underline{$49.0$}
\\
\cellcolor{red!4}\textbf{RangeFormer} & \cellcolor{red!4}$\mathbf{67.9}$ & \cellcolor{red!4}$\mathbf{95.4}$ & \cellcolor{red!4}$\mathbf{58.5}$ & \cellcolor{red!4}$\mathbf{73.7}$ & \cellcolor{red!4}$\mathbf{91.3}$ & \cellcolor{red!4}$\mathbf{73.1}$ & \cellcolor{red!4}$\mathbf{76.5}$ & \cellcolor{red!4}$\mathbf{88.7}$ & \cellcolor{red!4}$0.0$ & \cellcolor{red!4}$\mathbf{95.0}$ & \cellcolor{red!4}$\mathbf{56.5}$ & \cellcolor{red!4}$\mathbf{82.0}$ & \cellcolor{red!4}$10.0$ & \cellcolor{red!4}$\mathbf{88.7}$ & \cellcolor{red!4}$\mathbf{65.8}$ & \cellcolor{red!4}$\mathbf{86.8}$ & \cellcolor{red!4}$\mathbf{67.2}$ & \cellcolor{red!4}$\mathbf{73.7}$ & \cellcolor{red!4}$64.3$ & \cellcolor{red!4}$\mathbf{52.2}$
\\\bottomrule
\end{tabular}}
\vspace{0.2cm}
\label{table:str-class}
\end{table*}

\clearpage

\clearpage

\clearpage
{\small
\bibliographystyle{ieee_fullname}
\bibliography{egbib}
}

\end{document}